\pdfoutput=1

\documentclass[accepted]{article}

\usepackage{stfloats} 
\usepackage{microtype}
\usepackage{graphicx}
\graphicspath{{images/}}
\usepackage{subcaption}
\usepackage{float}
\usepackage{booktabs} 
\usepackage[table]{xcolor}
\usepackage{multirow}
\usepackage{array}
\usepackage{tabularx}
\usepackage{pgf}
\usepackage{balance}
\usepackage{hyperref}

\usepackage{algorithm}
\usepackage{algorithmic}
\newcommand{\mapcell}[2]{#1\if\relax\detokenize{#2}\relax\else\hspace{0.22em}{\fontsize{19pt}{21pt}\selectfont\textbf{\textcolor{black}{(}\textcolor{red}{+#2}\textcolor{black}{)}}}\fi}
\newcommand{\mapcellA}[2]{\makebox[5.0em][l]{\hspace{0.15em}#1\if\relax\detokenize{#2}\relax\else\hspace{0.22em}{\fontsize{8pt}{9pt}\selectfont\textbf{\textcolor{black}{(}\textcolor{red}{+#2}\textcolor{black}{)}}}\fi}}
\newcommand{\mapcellS}[2]{\fontsize{4pt}{5pt}\selectfont#1\if\relax\detokenize{#2}\relax\else\makebox[1.2em][l]{\hspace{0.12em}\raisebox{-0.15ex}{\fontsize{2.5pt}{3pt}\selectfont\textcolor{black}{(}\fontsize{2.5pt}{3pt}\selectfont\textcolor{red}{+#2}\textcolor{black}{)}}}\fi}
\newcommand{\tidedown}{\raisebox{-0.15ex}{\scalebox{0.46}{\textcolor{black}{(}\textcolor{green}{$\downarrow$}\textcolor{black}{)}}}}
\newcommand{\tideupAB}[2]{\raisebox{-0.15ex}{\fontsize{2pt}{2.5pt}\selectfont\textcolor{black}{(}\textcolor{red}{+#1/+#2}\textcolor{black}{)}}}
\newcommand{\effcell}[3]{\fontsize{4pt}{5pt}\selectfont#2\raisebox{-0.15ex}{\fontsize{2pt}{2.5pt}\selectfont\textcolor{black}{(}\scalebox{0.55}{\textcolor{green}{$\downarrow$}}\fontsize{2.5pt}{3pt}\selectfont\textcolor{green}{#3\%}\fontsize{2pt}{2.5pt}\selectfont\textcolor{black}{)}}}

\definecolor{impBest}{HTML}{FFE0C2}
\definecolor{impSecond}{HTML}{FFEDDA}
\definecolor{impThird}{HTML}{FFF6EC}
\definecolor{impBase}{HTML}{FFA500}
\colorlet{impStep1}{impBase!24!white}
\colorlet{impStep2}{impBase!40!white}
\colorlet{impStep3}{impBase!56!white}
\colorlet{impStep4}{impBase!72!white}
\colorlet{impStep5}{impBase!86!white}
\colorlet{impStep6}{impBase!96!white}
\newcommand{\applyimpcolor}[1]{%
\ifcase#1\relax
\or\cellcolor{impStep1}\or\cellcolor{impStep2}\or\cellcolor{impStep3}\or\cellcolor{impStep4}\or\cellcolor{impStep5}\or\cellcolor{impStep6}%
\else\cellcolor{impStep6}%
\fi
}
\newcommand{\mapcellSI}[2]{%
\if\relax\detokenize{#2}\relax
  \mapcellS{#1}{#2}%
\else
  \ifdim#2pt<0.005pt
    \textbf{\mapcellS{#1}{#2}}%
  \else
    \pgfmathtruncatemacro{\impsteps}{min(6,max(1,floor(#2/0.005)))}%
    \applyimpcolor{\impsteps}\textbf{\mapcellS{#1}{#2}}%
  \fi
\fi
}
\newcommand{\vheader}[1]{\shortstack[c]{\fontsize{4.2pt}{3.2pt}\selectfont\ttfamily\mdseries #1}}
\newcommand{\vh}[1]{{\fontsize{4.2pt}{3.2pt}\selectfont\bfseries #1}}

\newcommand{\costcellm}[3]{\shortstack[c]{\fontsize{5.0pt}{6.0pt}\selectfont #1{\fontsize{3.8pt}{4.2pt}\selectfont\textcolor{black}{(}\textcolor{red}{#3}\textcolor{black}{)}}\\\fontsize{5.0pt}{6.0pt}\selectfont #2}}

\newcolumntype{Y}{>{\centering\arraybackslash}X}
\newcolumntype{y}{>{\hsize=.75\hsize\centering\arraybackslash}X}  
\newcolumntype{m}{>{\hsize=1.35\hsize\centering\arraybackslash}X} 
\newcolumntype{e}{>{\hsize=1.05\hsize\centering\arraybackslash}X} 

\usepackage{icml2026}

\usepackage{amsmath}
\usepackage{amssymb}
\usepackage{mathtools}
\usepackage{amsthm}
\usepackage{etoolbox}

\AtBeginEnvironment{equation}{\small}
\AtBeginEnvironment{align}{\small}
\AtBeginEnvironment{align*}{\small}
\AtBeginEnvironment{gather}{\small}
\AtBeginEnvironment{gather*}{\small}
\usepackage[capitalize,noabbrev]{cleveref}

\theoremstyle{plain}

\theoremstyle{definition}

\theoremstyle{remark}

\usepackage[textsize=tiny]{todonotes}

\icmltitlerunning{Frequency-Guided Representation Learning for Small Object Detection}

\begin{document}
\makeatletter
\@ifundefined{b@tide}{\global\@namedef{b@tide}{{1}{2020}{{Bolya et~al.}}{{Bolya, Foley, Hays, and Hoffman}}}}{}
\@ifundefined{b@cascade-rcnn}{\global\@namedef{b@cascade-rcnn}{{2}{2017}{{Cai \& Vasconcelos}}{{Cai and Vasconcelos}}}}{}
\@ifundefined{b@detr}{\global\@namedef{b@detr}{{3}{2020}{{Carion et~al.}}{{Carion, Massa, Synnaeve, Usunier, Kirillov, and Zagoruyko}}}}{}
\@ifundefined{b@chen2026freq-detr}{\global\@namedef{b@chen2026freq-detr}{{4}{2026}{{Chen et~al.}}{{Chen, Liu, Sun, and Wang}}}}{}
\@ifundefined{b@freqfusion}{\global\@namedef{b@freqfusion}{{5}{2024}{{Chen et~al.}}{{Chen, Fu, Gu, Yan, Harada, and Huang}}}}{}
\@ifundefined{b@fdconv}{\global\@namedef{b@fdconv}{{6}{2025}{{Chen et~al.}}{{Chen, Gu, Li, Yan, and Fu}}}}{}
\@ifundefined{b@dai2021dyhead}{\global\@namedef{b@dai2021dyhead}{{7}{2021}{{Dai et~al.}}{{Dai, Chen, Xiao, Chen, Liu, Yuan, and Zhang}}}}{}
\@ifundefined{b@du2018unmanned}{\global\@namedef{b@du2018unmanned}{{8}{2018}{{Du et~al.}}{{Du, Qi, Yu, Yang, Duan, Li, Zhang, Huang, and Tian}}}}{}
\@ifundefined{b@du2019visdrone}{\global\@namedef{b@du2019visdrone}{{9}{2019}{{Du et~al.}}{{Du, Zhu, Wen, Bian, Lin, Hu, Peng, Zheng, Wang, Zhang, et~al.}}}}{}
\@ifundefined{b@du2024cfpt}{\global\@namedef{b@du2024cfpt}{{10}{2024}{{Du et~al.}}{{Du, Hu, Zhao, Jin, and Ma}}}}{}
\@ifundefined{b@tood}{\global\@namedef{b@tood}{{11}{2021}{{Feng et~al.}}{{Feng, Zhong, Gao, Scott, and Huang}}}}{}
\@ifundefined{b@wtconv}{\global\@namedef{b@wtconv}{{12}{2025}{{Finder et~al.}}{{Finder, Amoyal, Treister, and Freifeld}}}}{}
\@ifundefined{b@yolox}{\global\@namedef{b@yolox}{{13}{2021}{{Ge et~al.}}{{Ge, Liu, Wang, Li, and Sun}}}}{}
\@ifundefined{b@huang2025deimdetrimprovedmatching}{\global\@namedef{b@huang2025deimdetrimprovedmatching}{{14}{2025}{{Huang et~al.}}{{Huang, Lu, Cun, Yu, Zhou, and Shen}}}}{}
\@ifundefined{b@yolov11}{\global\@namedef{b@yolov11}{{15}{2024}{{Khanam \& Hussain}}{{Khanam and Hussain}}}}{}
\@ifundefined{b@li2024rethinking}{\global\@namedef{b@li2024rethinking}{{16}{2024}{{Li}}{{}}}}{}
\@ifundefined{b@gfl}{\global\@namedef{b@gfl}{{17}{2020}{{Li et~al.}}{{Li, Wang, Wu, Chen, Hu, Li, Tang, and Yang}}}}{}
\@ifundefined{b@wavelet-transformer}{\global\@namedef{b@wavelet-transformer}{{18}{2025}{{Li et~al.}}{{Li, Jiao, Liu, Yang, Zhu, Liu, Li, and Ma}}}}{}
\@ifundefined{b@RetinaNet}{\global\@namedef{b@RetinaNet}{{19}{2017{a}}{{Lin et~al.}}{{Lin, Goyal, Girshick, He, and Doll{\'{a}}r}}}}{}
\@ifundefined{b@lin2014microsoft}{\global\@namedef{b@lin2014microsoft}{{20}{2014}{{Lin et~al.}}{{Lin, Maire, Belongie, Bourdev, Girshick, Hays, Perona, Ramanan, Zitnick, and Doll{\'a}r}}}}{}
\@ifundefined{b@fpn}{\global\@namedef{b@fpn}{{21}{2017{b}}{{Lin et~al.}}{{Lin, Doll{\'a}r, Girshick, He, Hariharan, and Belongie}}}}{}
\@ifundefined{b@liu2025wdfs}{\global\@namedef{b@liu2025wdfs}{{22}{2025}{{Liu \& Xie}}{{Liu and Xie}}}}{}
\@ifundefined{b@liu2018pathaggregationnetworkinstance}{\global\@namedef{b@liu2018pathaggregationnetworkinstance}{{23}{2018}{{Liu et~al.}}{{Liu, Qi, Qin, Shi, and Jia}}}}{}
\@ifundefined{b@ssd}{\global\@namedef{b@ssd}{{24}{2016}{{Liu et~al.}}{{Liu, Anguelov, Erhan, Szegedy, Reed, Fu, and Berg}}}}{}
\@ifundefined{b@luan2026wcdb}{\global\@namedef{b@luan2026wcdb}{{25}{2026}{{Luan et~al.}}{{Luan, Dong, Zhou, Li, Xie, Liu, and Zhu}}}}{}
\@ifundefined{b@peng2024dfine}{\global\@namedef{b@peng2024dfine}{{26}{2024}{{Peng et~al.}}{{Peng, Li, Wu, Zhang, Sun, and Wu}}}}{}
\@ifundefined{b@fcanet}{\global\@namedef{b@fcanet}{{27}{2021}{{Qin et~al.}}{{Qin, Zhang, Wu, and Li}}}}{}
\@ifundefined{b@gfnet}{\global\@namedef{b@gfnet}{{28}{2023}{{Rao et~al.}}{{Rao, Zhao, Zhu, Zhou, and Lu}}}}{}
\@ifundefined{b@yolov1}{\global\@namedef{b@yolov1}{{29}{2016}{{Redmon et~al.}}{{Redmon, Divvala, Girshick, and Farhadi}}}}{}
\@ifundefined{b@fasterrcnn}{\global\@namedef{b@fasterrcnn}{{30}{2015}{{Ren et~al.}}{{Ren, He, Girshick, and Sun}}}}{}
\@ifundefined{b@yolo26}{\global\@namedef{b@yolo26}{{31}{2026}{{Sapkota et~al.}}{{Sapkota, Cheppally, Sharda, and Karkee}}}}{}
\@ifundefined{b@hs-fpn}{\global\@namedef{b@hs-fpn}{{32}{2025}{{Shi et~al.}}{{Shi, Hu, Ren, Ye, Yuan, Ouyang, He, Ji, and Guo}}}}{}
\@ifundefined{b@tan2020efficientdetscalableefficientobject}{\global\@namedef{b@tan2020efficientdetscalableefficientobject}{{33}{2020}{{Tan et~al.}}{{Tan, Pang, and Le}}}}{}
\@ifundefined{b@mobileuvit}{\global\@namedef{b@mobileuvit}{{34}{2025}{{Tang et~al.}}{{Tang, Nian, Ding, Ma, Quan, Dong, Yang, Liu, and Zhou}}}}{}
\@ifundefined{b@yolov12}{\global\@namedef{b@yolov12}{{35}{2025}{{Tian et~al.}}{{Tian, Ye, and Doermann}}}}{}
\@ifundefined{b@sod_applications}{\global\@namedef{b@sod_applications}{{36}{2020}{{Tong et~al.}}{{Tong, Wu, and Zhou}}}}{}
\@ifundefined{b@lsnet}{\global\@namedef{b@lsnet}{{37}{2025}{{Wang et~al.}}{{Wang, Chen, Lin, Han, and Ding}}}}{}
\@ifundefined{b@sabl2019}{\global\@namedef{b@sabl2019}{{38}{2020}{{Wang et~al.}}{{Wang, Zhang, Cao, Chen, Pang, Gong, Shi, Loy, and Lin}}}}{}
\@ifundefined{b@xia2018dota}{\global\@namedef{b@xia2018dota}{{39}{2018}{{Xia et~al.}}{{Xia, Bai, Ding, Zhu, Belongie, Luo, Datcu, Pelillo, and Zhang}}}}{}
\@ifundefined{b@fbrt-yolo}{\global\@namedef{b@fbrt-yolo}{{40}{2025}{{Xiao et~al.}}{{Xiao, Xu, Xin, and Li}}}}{}
\@ifundefined{b@haar}{\global\@namedef{b@haar}{{41}{2023}{{Xu et~al.}}{{Xu, Liao, Zhang, Li, He, and Wu}}}}{}
\@ifundefined{b@yu2016multiscalecontextaggregationdilated}{\global\@namedef{b@yu2016multiscalecontextaggregationdilated}{{42}{2016}{{Yu \& Koltun}}{{Yu and Koltun}}}}{}
\@ifundefined{b@yu2020scale}{\global\@namedef{b@yu2020scale}{{43}{2020}{{Yu et~al.}}{{Yu, Gong, Jiang, Ye, and Han}}}}{}
\@ifundefined{b@dino}{\global\@namedef{b@dino}{{44}{2022{a}}{{Zhang et~al.}}{{Zhang, Li, Liu, Zhang, Su, Zhu, Ni, and Shum}}}}{}
\@ifundefined{b@zhang2019shiftinvariant}{\global\@namedef{b@zhang2019shiftinvariant}{{45}{2019}{{Zhang}}{{}}}}{}
\@ifundefined{b@elan}{\global\@namedef{b@elan}{{46}{2022{b}}{{Zhang et~al.}}{{Zhang, Zeng, Guo, and Zhang}}}}{}
\@ifundefined{b@rt-detr}{\global\@namedef{b@rt-detr}{{47}{2024}{{Zhao et~al.}}{{Zhao, Lv, Xu, Wei, Wang, Dang, Liu, and Chen}}}}{}
\makeatother

\makeatletter
\renewcommand{\Notice@String}{Preprint. \today}
\renewcommand{\ICML@appearing}{}
\makeatother
\twocolumn[
  \icmltitle{From Spatial to Spectral: An Efficient, Frequency-Guided Feature Representation Learner for Small Object Detection}



  \icmlsetsymbol{equal}{*}
  \icmlsetsymbol{corr}{\ensuremath{\dagger}}

  \begin{icmlauthorlist}
    \icmlauthor{Yuhan Rui}{equal,sch}
    \icmlauthor{Shihan Qiao}{equal,sch}
    \icmlauthor{Yibin Lou}{equal,sch}
    \icmlauthor{Mingxi Yu}{equal,sch}
    \icmlauthor{Yutong Wan}{sch}
    \icmlauthor{Yanqiao Chen}{sch}
    \icmlauthor{Dongsheng Hou}{sch}
    \icmlauthor{Zhen Cao}{sch}
    \icmlauthor{Athena Zhuoming Zhong}{upenn}
    \icmlauthor{Qi Hao}{corr,sch}
  \end{icmlauthorlist}

  \icmlaffiliation{sch}{Southern University of Science and Technology, Shenzhen, China}
  \icmlaffiliation{upenn}{University of Pennsylvania}

  \icmlcorrespondingauthor{Yuhan Rui}{12310520@mail.sustech.edu.cn}
  \icmlcorrespondingauthor{Qi Hao}{hao.q@sustech.edu.cn}

  \icmlkeywords{Small object detection, Spectral representation, Plug-and-play modules, Decompose-Enhance-Reconstruct, Efficient detectors, Architecture-agnostic}

  \vskip 0.3in
]



\printAffiliationsAndNotice{\icmlEqualContribution \textsuperscript{\ensuremath{\dagger}}Corresponding author. }

\begin{abstract}

Efficient small object detection is bottlenecked by the inherent feature scarcity of tiny targets, which is further aggravated by operations of spatial-domain detectors that indiscriminately discard critical high-frequency details. 
Recovering these fragile cues within the spatial domain is notoriously difficult, as it often requires computationally expensive architectural upscaling that inadvertently amplifies background noise. 
To bridge this gap, we propose a paradigm \textbf{shift from spatial to spectral} feature processing, introducing a holistic solution with the following novelty: 
(1) A versatile \textbf{Frequency-Guided Feature Representation framework} that generalizes across diverse detector architectures (both CNN and Transformer-based), offering a robust alternative to spatial-only feature extraction; (2) The unified \textbf{Decompose--Enhance--Reconstruct (DER)} operator, instantiated via three \textbf{lightweight, plug-and-play} modules---Wavelet-Difference Gate (WDG), Log-Gabor Enhancer (LGE), and Frequency-Driven Head (FDHead)---to systematically inject frequency-aware modulation into the backbone, neck, and head. This mechanism decouples feature modeling from resolution reduction, capturing discriminative high-frequency components to enable accurate localization with significantly reduced parameter redundancy; (3) Extensive validation on multi-domain benchmarks (VisDrone2019, UAVDT, TinyPerson, DOTAv1) demonstrating consistent gains. Notably, our proposed \textbf{DERNet} series outperforms YOLOv11 models under the same scale while requiring \textbf{only 1/6 of the parameters}, backed by rigorous spectral diagnostics and error decomposition analysis.

\end{abstract}

\section{Introduction}

\begin{figure}[h]
\centering
\includegraphics[width=\columnwidth]{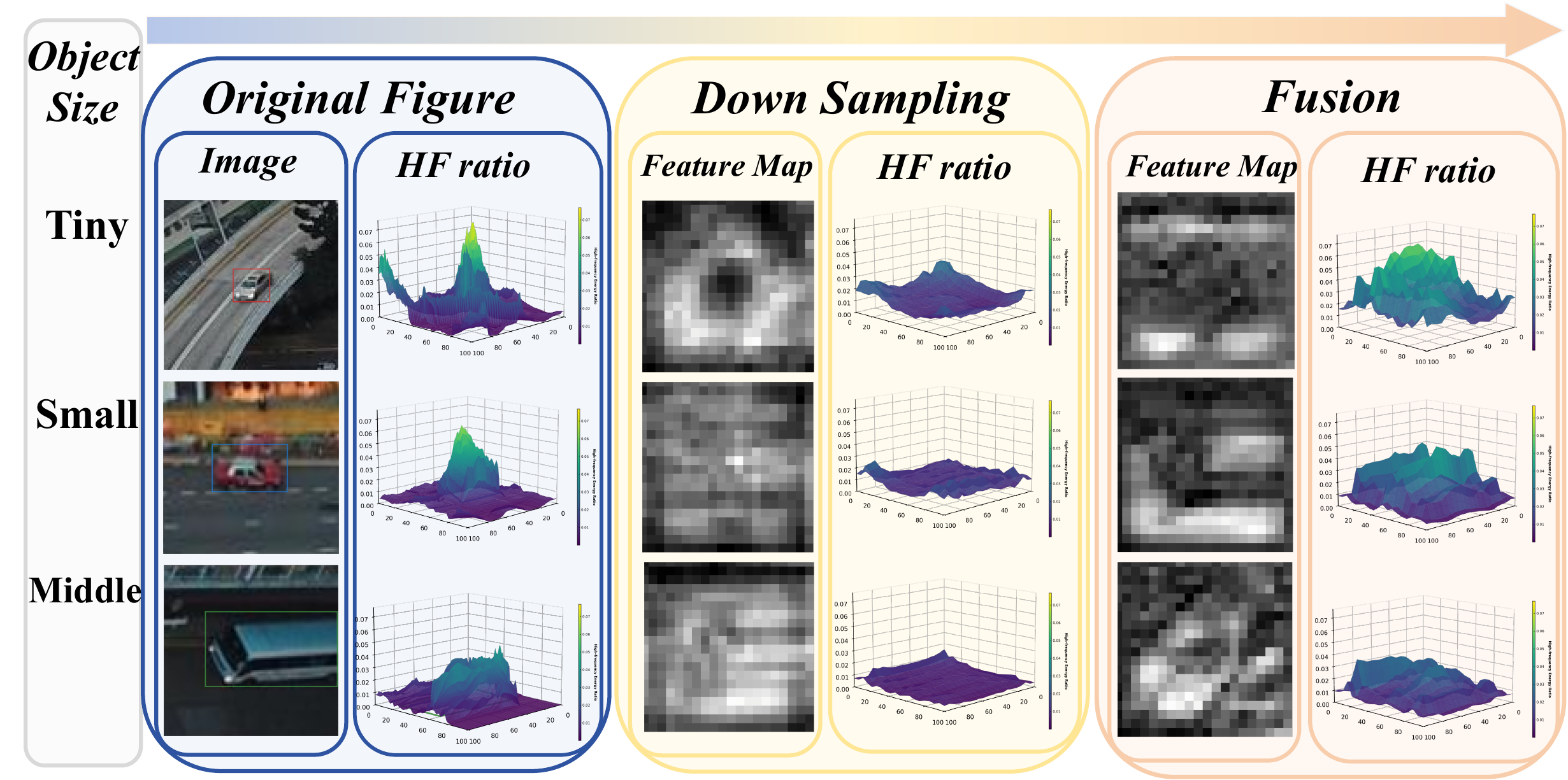}
\caption{\textbf{Motivation of frequency-domain bias in small-object detection.} High-frequency energy ratio (HF/total) across object scales, where HF is defined by 2D FFT components whose radial distance from the spectrum center exceeds 33\% of the maximum radius.}
\label{fig:frequency}
\vspace{-0.5cm}
\end{figure}

Small object detection underpins critical applications ranging from aerial surveillance to autonomous navigation \cite{sod_applications}. Despite advances in general-purpose detectors, performance degrades sharply on extreme scales (e.g., $<32^2$ pixels)\cite{lin2014microsoft}. We attribute this failure not merely to resolution limits, but to \textbf{compounding spectral degradation} across the entire detection pipeline.

As shown in Fig.~\ref{fig:frequency}, high-frequency components dominate tiny objects yet decay rapidly after downsampling and fusion, leading to \textbf{compounding spectral degradation} across the \textbf{backbone}, \textbf{neck}, and \textbf{head} stages of the detection pipeline.

(i) In the \textbf{backbone}, strided downsampling is rarely anti-aliased, which induces spectral aliasing and mixes attenuated high-frequency cues into low-frequency components, causing contaminated features that propagate downstream \cite{zhang2019shiftinvariant};
(ii) in the \textbf{neck}, multi-scale fusion (e.g., summation/concatenation) further favors dominant low-frequency semantics, diluting the already fragile detail residuals;
and (iii) in the \textbf{head}, box regression operates on over-smoothed representations without explicit mechanisms to emphasize boundary evidence, which can lead to unstable localization for tiny targets \cite{sabl2019}.

This motivates a \textbf{key question}: \emph{can we preserve and reinject \textbf{stage-specific spectral cues} \emph{without spatial upscaling}, while remaining \textbf{plug-and-play across detector architectures}?}
Prior remedies largely rely on spatial restoration (e.g., dilation or denser pyramids) \cite{yu2016multiscalecontextaggregationdilated,fpn}, which is often compute-heavy and may amplify background noise \cite{tan2020efficientdetscalableefficientobject}. Meanwhile, frequency-domain learning \cite{gfnet,fcanet} has mostly been explored for classification and lacks a unified mechanism tailored to detection.

\begin{figure}[h]
\centering
\includegraphics[width=\columnwidth]{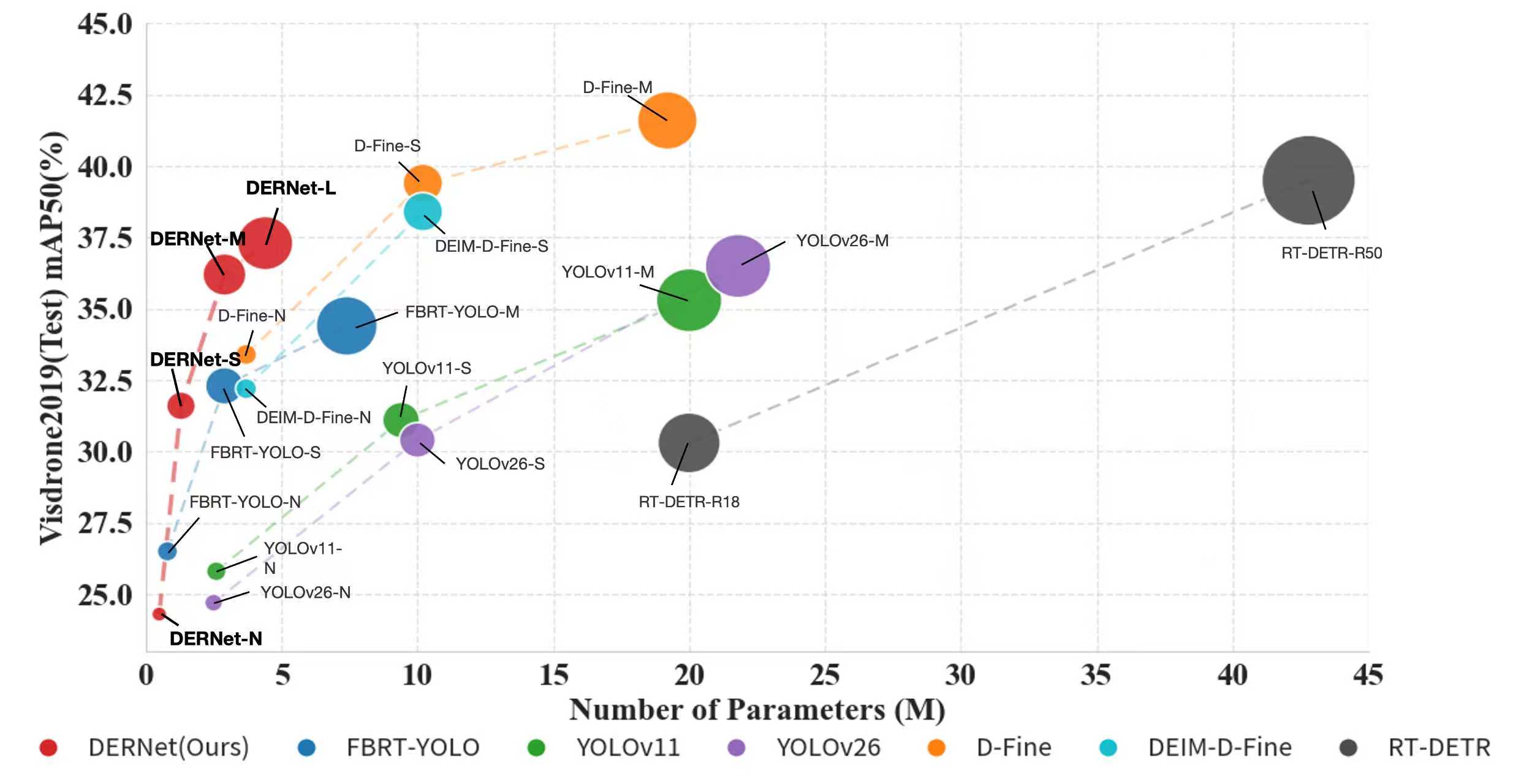}
\caption{Performance comparison of DERNet with state-of-the-art methods on the VisDrone2019 test set.}
\label{fig:comparison}
\vspace{-0.6cm}
\end{figure}

In this paper, we propose a paradigm shift from spatial restoration to \textbf{Frequency-Guided Feature Representation}. Central to our approach is the unified \textbf{Decompose--Enhance--Reconstruct (DER)} operator, which decouples feature modeling from resolution reduction by explicitly separating and enhancing detail-relevant spectral cues. 
We instantiate this philosophy via three \textbf{stage-adaptive} modules: 
(1) a \textbf{Wavelet-Difference Gate (WDG)} in the backbone to refine the low-frequency approximation while using high-frequency subbands as a content-adaptive gate, thereby reducing low-frequency contamination without amplifying high-frequency noise;
(2) a \textbf{Log-Gabor Enhancer (LGE)} in the neck to re-activate directional high-frequency residuals before multi-scale fusion, mitigating the tendency of fusion to bias toward dominant semantics and dilute fine details; 
and (3) a \textbf{Frequency-Driven Head (FDHead)} to inject boundary-sensitive gains (estimated from high-frequency energy) into dense regression, stabilizing localization for tiny targets.
Our contributions are summarized as follows:
\vspace{-0.5cm}
\begin{itemize}
  \setlength{\topsep}{-14pt}
  \setlength{\partopsep}{-9pt}
  \setlength{\itemsep}{1pt}
  \setlength{\parskip}{0pt}
  \setlength{\parsep}{0pt}
  \setlength{\leftmargin}{0pt}
\item \textbf{A unified frequency-guided abstraction.}
We propose the \textbf{Decompose--Enhance--Reconstruct (DER)} operator as a
stage-agnostic interface for frequency-aware feature modeling, enabling explicit
manipulation of high-frequency cues \emph{before} irreversible operations such
as downsampling, fusion, and regression.

\item \textbf{Stage-adaptive instantiation and universality.}
We allow DER to be flexibly instantiated as WDG, LGE, and FDHead to address
specific spectral needs across the backbone, neck, and head. This design enables
\textbf{plug-and-play integration} into diverse CNN- and Transformer-based
detectors via functionally equivalent insertion points without architectural
overhaul.

\item \textbf{Rigorous validation, mechanism verification, and performance.}
Beyond extensive standard benchmarking, we employ rigorous spectral analysis to explicitly prove the restoration of boundary details. 
These mechanisms culminate in \textbf{DERNet}, which delivers competitive accuracy with a fraction of the computational cost (Fig.~\ref{fig:comparison}).

\end{itemize}
\vspace{-0.5cm}

  \section{Related Work}
  We review prior work from three angles that are most relevant to our goal.

  \begin{figure*}[t]
  \centering
  \includegraphics[width=\textwidth]{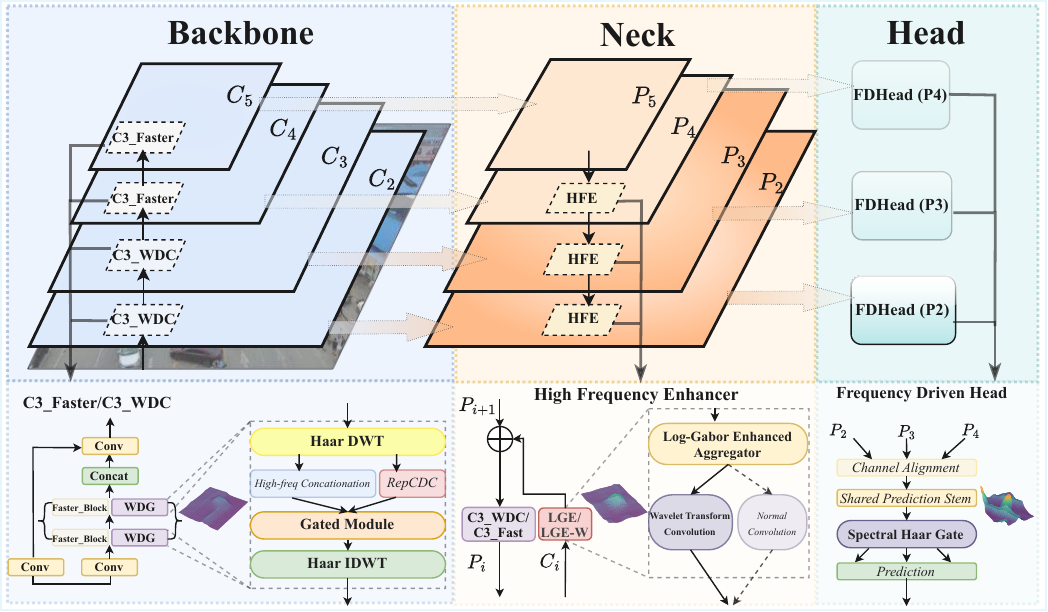}
  \vspace{-0.2cm}
  \caption{\textbf{Overall framework of our Frequency-Guided Feature Representation Learner.} This architecture instantiates the \textbf{Decompose--Enhance--Reconstruct (DER)} operator via WDG, LGE, and FDHead to \textbf{systematically decouple feature modeling from resolution reduction}, ensuring that discriminative high-frequency cues are explicitly preserved and amplified across the entire feature stream.}
  \label{fig:framework}
  \vspace{-0.5cm}
  \end{figure*}

  \subsection{Efficient Detector Architectures}
  
  Real-time detection advances largely through architectural optimizations in backbones and feature pyramids. One-stage YOLO-style detectors \cite{yolov1,yolov11,yolo26} balance accuracy and latency via advanced block designs and multi-scale prediction, while lightweight enhancements often employ multi-kernel perception or large-receptive-field modeling to boost representational diversity in cluttered scenes \cite{fbrt-yolo,elan}. 
  
  Concurrently, Transformer-based detectors \cite{detr} eliminate hand-crafted anchors for end-to-end prediction, with recent efficient variants (e.g., RT-DETR \cite{rt-detr}, DINO \cite{dino}) achieving competitive performance under constrained budgets. Despite these strides, a common tension persists for small objects across both families: enhancing fine-detail sensitivity typically incurs significant computational overhead or requires specific architectural intrusions (e.g., large kernels or U-shaped designs \cite{mobileuvit}), complicating the deployment of a uniformly effective solution.
  
  \subsection{Small Object Detection}
  
  Small objects suffer from inherent information scarcity, occupying minimal pixel area and easily vanishing during downsampling and coarse fusion. Classical detectors like Faster R-CNN and SSD \cite{fasterrcnn,ssd} revealed the difficulty of preserving these weak cues under deep feature hierarchies. To mitigate this, extensive research has focused on strengthening multi-scale fusion (e.g., FPN and its variants \cite{fpn,liu2018pathaggregationnetworkinstance,li2024rethinking}), introducing finer pyramid levels, and designing cross-layer pyramid interaction modules to enhance small-scale feature representation \cite{du2024cfpt}.
  
  Recently, the focus has shifted toward \emph{detail-aware} enrichment. For instance, HS-FPN utilizes high-frequency responses as mask weights to highlight tiny objects \cite{hs-fpn}, while context modeling techniques employ large receptive fields to distinguish targets from background clutter \cite{lsnet,fbrt-yolo}. However, many of these approaches focus on either spatial fusion or receptive-field engineering, while the \emph{mechanism of how fine details are suppressed and should be reconstructed} is often left implicit, and portability across heterogeneous detector designs is not always validated(Appendix~\ref{sec:appendix_l_comparison}).
  
  \subsection{Frequency-Domain Modeling for Dense Prediction}
  
  Frequency-domain analysis provides efficient tools for representation learning. For instance, GFNet leverages FFT for efficient global token mixing with log-linear complexity \cite{gfnet}, while FcaNet reinterprets channel attention as multi-spectral compression \cite{fcanet}. Recently, spectral methods have advanced dense prediction: FDConv enhances dynamic convolutions by constructing frequency-diverse kernels \cite{fdconv}, and FreqFusion utilizes adaptive filtering to sharpen boundaries during upsampling \cite{freqfusion}. Wavelet-based approaches further expand receptive fields or preserve downsampling fidelity via sub-band transforms \cite{wtconv,haar,wavelet-transformer}.
  
  Despite these advances, existing designs remain \emph{task- or component-specific}—often restricted to classification backbones or fusion layers. Critically, they typically overlook the stage-dependent nature of frequency perception (i.e., the varying spectral needs across the architecture) and lack mechanisms to explicitly decouple and differentially process low- versus high-frequency components, even in recent frequency-aware detection frameworks \cite{chen2026freq-detr}. We bridge this gap with a \textbf{stage-aware} frequency-guided operator that is compatible with heterogeneous detectors.

\section{Method}

\begin{figure*}[t]
  \centering
  \includegraphics[width=\textwidth]{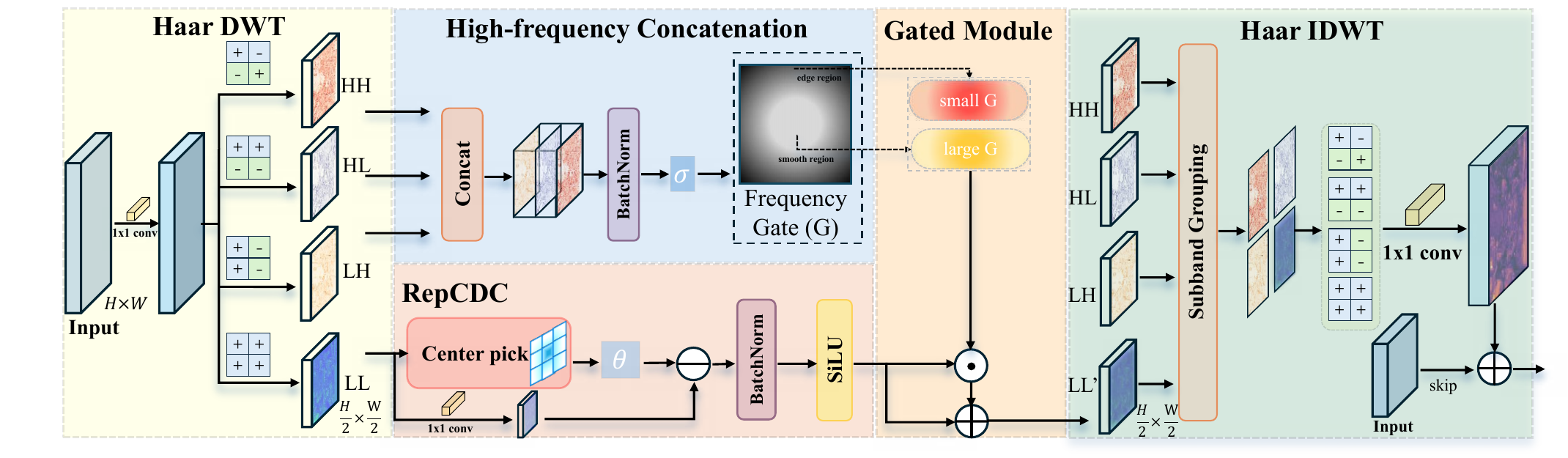}
  \caption{\textbf{Architecture of the Wavelet-Difference Gate (WDG)}. WDG decomposes features via Haar DWT, refines the low-frequency subband with RepCDC, predicts a content-adaptive gate from high-frequency subbands, and reconstructs via IDWT with a skip connection.}
  \label{fig:wdg}
  \vspace{-0.5cm}
  \end{figure*}
  \begin{algorithm}[ht]
    \caption{DER insertion across the architecture.}
    \label{alg:der_insertion}
    \scriptsize
    \begingroup
    \setlength{\fboxsep}{1.2pt}
    \newcommand{\algbox}[2]{\colorbox{#1}{\parbox[t]{\dimexpr\linewidth-2\fboxsep\relax}{#2}}}
    \begin{algorithmic}
      \STATE \algbox{black!4}{\textbf{Input:} image $\mathbf{I}$; detector $(\mathcal{B},\mathcal{N},\mathcal{H})$}
      \STATE \algbox{black!4}{\textbf{Output:} enhanced prediction $\widehat{\mathbf{Y}}$}
      \STATE \algbox{gray!10}{\textbf{Definitions:}}
      \STATE \algbox{gray!10}{\hspace{1.0em}$\mathcal{B}$: backbone,\; $\mathcal{N}$: neck,\; $\mathcal{H}$: detection head}
      \STATE \algbox{gray!10}{\hspace{1.0em}$\{\mathbf{C}_\ell\}$: backbone feature maps,\; $\{\mathbf{P}_\ell\}$: neck feature maps}
      \STATE \algbox{gray!10}{\hspace{1.0em}$(\mathbf{X}_{L},\mathbf{X}_{H})$: low/high-frequency components of a generic feature $\mathbf{X}$}
      \STATE \algbox{yellow!10}{\textbf{Goals:}}
      \STATE \algbox{yellow!10}{\hspace{1.0em}backbone: anti-aliasing and low-frequency cleaning before stride}
      \STATE \algbox{yellow!10}{\hspace{1.0em}neck: directional high-frequency residual recovery before fusion smoothing}
      \STATE \algbox{yellow!10}{\hspace{1.0em}head: boundary-aligned regression gain at the finest level}
      \STATE \algbox{cyan!8}{\textbf{DER interface:}}
      \STATE \algbox{cyan!8}{\hspace{1.0em}$(\mathbf{X}_{L},\mathbf{X}_{H})\leftarrow\mathcal{D}(\mathbf{X})$ \COMMENT{decompose into low/high-frequency}}
      \STATE \algbox{cyan!8}{\hspace{1.0em}$\mathbf{X}_{L}^{+}\leftarrow\mathcal{E}_{L}(\mathbf{X}_{L})$, $\mathbf{X}_{H}^{+}\leftarrow\mathcal{E}_{H}(\mathbf{X}_{H})$ \COMMENT{stage-specific enhancement}}
      \STATE \algbox{cyan!8}{\hspace{1.0em}$\mathbf{X}^{+}\leftarrow\mathcal{R}(\mathbf{X}_{L}^{+},\mathbf{X}_{H}^{+})$ \COMMENT{reconstruct enhanced feature}}
      \STATE \algbox{blue!6}{\textbf{Backbone module (WDG).}}
      \STATE \algbox{blue!6}{$\{\mathbf{C}_\ell\} \leftarrow \mathcal{B}(\mathbf{I})$ \COMMENT{compute backbone features}}
      \FOR{each high-resolution bottleneck $\mathbf{C}_\ell$}
        \STATE \algbox{blue!6}{$\mathbf{C}_\ell \leftarrow \mathrm{DER}_{\mathrm{WDG}}(\mathbf{C}_\ell)$ \COMMENT{apply WDG before stride}}
      \ENDFOR
      \STATE \algbox{violet!7}{\textbf{Neck module (LGE).}}
      \STATE \algbox{violet!7}{$\{\mathbf{P}_\ell\} \leftarrow \mathcal{N}(\{\mathbf{C}_\ell\})$ \COMMENT{compute neck features}}
      \FOR{each designated fusion output $\mathbf{P}_\ell$}
        \STATE \algbox{violet!7}{$\mathbf{P}_\ell \leftarrow \mathrm{DER}_{\mathrm{LGE}}(\mathbf{P}_\ell)$ \COMMENT{apply LGE before fusion smoothing}}
      \ENDFOR
    \STATE \algbox{red!10}{\textbf{Head module (FDHead).}}
    \STATE \algbox{red!10}{$\widehat{\mathbf{Y}} \leftarrow \mathrm{DER}_{\mathrm{FDHead}}(\mathcal{H}(\{\mathbf{P}_\ell\}))$ \COMMENT{apply FDHead at finest-level regression}}
    \end{algorithmic}
    \endgroup
    \end{algorithm}
    \vspace{-0.1cm}
\subsection{Overall Framework}

Rather than proposing isolated frequency modules, we formulate a single operator interface (DER) whose instantiations are tailored to the functional role of each stage.
We propose a \textbf{Frequency-Guided Feature Representation} framework that injects spectral inductive bias into detectors \emph{without changing their macro-architecture}. The framework is architecture-agnostic because it targets \emph{functionally equivalent insertion points} across detector families (Appendix~\ref{sec:appendix_c_implementation_details}, Table~\ref{tab:unified_insertion_points}): high-resolution backbone bottlenecks before stride, designated fusion outputs in the neck, and the finest-level regression pathway in the head (Fig.~\ref{fig:framework}).

Our key idea is \emph{stage-wise spectral objectives}: backbone anti-aliasing / low-frequency contamination suppression, neck directional residual recovery, and head boundary-aligned regression gain. These objectives are realized by \textbf{DER} and its stage-specific instantiations.


Given an input feature tensor $\mathbf{X} \in \mathbb{R}^{C \times H \times W}$, DER is defined as a unified interface:
\vspace{0cm}
\begin{equation}
\label{eq:der_operator}
\begin{gathered}
\mathrm{DER}(\mathbf{X}) = \mathcal{R}\bigl(\mathcal{E}_{L}(\mathbf{X}_{L}),\,\mathcal{E}_{H}(\mathbf{X}_{H})\bigr)\\
(\mathbf{X}_{L},\mathbf{X}_{H}) = \mathcal{D}(\mathbf{X})
\end{gathered}
\vspace{0cm}
\end{equation}
Here $\mathcal{D}$ decomposes $\mathbf{X}$, $\mathcal{E}_L/\mathcal{E}_H$ perform stage-specific enhancement, and $\mathcal{R}$ reinjects the enhanced components through a stable base path.
WDG/LGE/FDHead should be read as stage-specific realizations of this DER interface, rather than independent add-on modules.

\noindent\textbf{DER Instantiations Across Backbone, Neck, and Head.}  
We instantiate DER under two verifiable principles (Appendix~\ref{sec:appendix_d_frequency_choice}): \emph{(i) directional spectral selectivity}, i.e., the decomposition isolates boundary-aligned responses instead of amplifying global textures; and \emph{(ii) low-cost reinjection compatibility}, i.e., the enhanced signals can be injected through invertible reconstruction or re-parameterizable blocks with negligible inference overhead. It is therefore reasonable that $\mathcal{D}$ differs across stages: wavelets (invertible, local, multi-resolution) support reconstruction and energy-based confidence for repeated use in the backbone/head, while Log-Gabor filters (zero-DC, orientation-selective, fixed kernels) better match the neck’s role of restoring anisotropic residuals that fusion tends to average out. Alg.~\ref{alg:der_insertion} summarizes the insertion procedure.

\noindent\textbf{Design intuition (Operational Decoupling).}  
DER structurally decouples feature modeling from resolution reduction by intervening \emph{prior to} stride, fusion, and regression. This allows spectral cues to be explicitly shaped while high-resolution details remain intact, preventing discriminative information from being passively determined or discarded by subsequent irreversible operations.

\noindent\textbf{Principle (Stage-compatible decomposition).}  
The decomposition operator $\mathcal{D}$ adheres to a \textbf{stage-dependent yet interface-consistent} design. By selecting from a family of operators that are local, directionally selective, and reconstruction-compatible, we render the heterogeneous use of Wavelets and Log-Gabor filters principled rather than ad hoc. This formulation \textbf{reconciles} unified architectural insertion points (Appendix~\ref{sec:appendix_c_implementation_details}) with stage-specific operator tailoring (Appendix~\ref{sec:appendix_d_frequency_choice}).

\subsection{Wavelet-Difference Gate (WDG)}

\noindent\textbf{Design rationale.}  
Backbone bottlenecks are where small object boundaries are most vulnerable: subsequent strided convolutions attenuate high-frequency cues and contaminate low-frequency features with aliased residuals. WDG intervenes at this juncture by \emph{refining the low-frequency subband with an edge-aware operator while using high-frequency subbands as a self-derived gating prior}, thereby reinforcing boundary-relevant regions \emph{before} irreversible resolution reduction (Fig.~\ref{fig:wdg}).

\noindent\textbf{Wavelet decomposition.}  
WDG projects the input $\mathbf{x}\in\mathbb{R}^{C\times H\times W}$ to a hidden space and applies 2D Haar DWT:
\vspace{0cm}
\begin{equation}
\label{eq:wdg_proj}
(\mathbf{x}_{LL},\mathbf{x}_{LH},\mathbf{x}_{HL},\mathbf{x}_{HH}) = \mathrm{DWT}\bigl(f_{1\times 1}(\mathbf{x})\bigr),
\vspace{0cm}
\end{equation}
yielding a low-freq approximation $\mathbf{x}_{LL}$ and three directional high-freq subbands at half resolution (see Appendix~\ref{sec:appendix_wdg_details}).

\noindent\textbf{Edge-aware low-frequency refinement via RepCDC.}  
The LL subband inherits smoothing bias that suppresses fine gradients critical for small object boundaries. We compensate with a \emph{Re-parameterized Central-Difference Convolution} (RepCDC). Let $\mathbf{W}\in\mathbb{R}^{C_{\mathrm{out}}\times C_{\mathrm{in}}\times 3\times 3}$ be the base kernel and $\boldsymbol{\theta}\in\mathbb{R}^{C_{\mathrm{out}}\times C_{\mathrm{in}}}$ a learnable center-difference parameter; the output is:
\vspace{0cm}
\begin{equation}
\label{eq:wdg_repcdc}
\mathbf{y}^{(o)}_{p,q} = \sum_{c=1}^{C_{\mathrm{in}}}\sum_{i=-1}^{1}\sum_{j=-1}^{1} \mathbf{W}^{(o,c)}_{i,j}\,\mathbf{z}^{(c)}_{p+i,q+j} \;-\; \sum_{c=1}^{C_{\mathrm{in}}} \boldsymbol{\theta}^{(o,c)}\,\mathbf{z}^{(c)}_{p,q},
\vspace{0cm}
\end{equation}
where $\mathbf{z}=\mathbf{x}_{LL}$. By subtracting the center response, RepCDC embeds a \emph{learned edge detector} into the convolution---unlike standard or depthwise kernels that weight all positions uniformly. This synergizes with the wavelet decomposition: the LL subband provides a denoised canvas free of high-frequency noise, while RepCDC re-injects the gradient information that was attenuated during averaging. At inference, the modified kernel fuses into a single $3{\times}3$ convolution with \textbf{zero additional latency}.

\noindent\textbf{High-frequency gated modulation.}  
Rather than treating high-frequency subbands as auxiliary features to be fused, WDG repurposes them as a \emph{self-derived attention prior}. We concatenate $\{\mathbf{x}_{LH},\mathbf{x}_{HL},\mathbf{x}_{HH}\}$ and project to a scalar gate per location:
\vspace{0cm}
\begin{equation}
\label{eq:wdg_gate}
\mathbf{g} = \sigma\!\left(\mathrm{BN}\bigl(f_{1\times 1}([\mathbf{x}_{LH};\mathbf{x}_{HL};\mathbf{x}_{HH}])\bigr)\right) \in (0,1)^{\frac{H}{2}\times\frac{W}{2}}.
\vspace{0cm}
\end{equation}
Crucially, $\mathbf{g}$ is computed from the \emph{same decomposition} used for refinement, forming a \textbf{closed-loop} where boundary evidence directly steers low-frequency enhancement:
\vspace{0cm}
\begin{equation}
\label{eq:wdg_modulate}
\widetilde{\mathbf{x}}_{LL} = \mathbf{y}_{LL} \odot (1 + \mathbf{g}).
\vspace{0cm}
\end{equation}
The additive form $(1+\mathbf{g})$ preserves baseline magnitude while amplifying boundary-critical regions---no external attention module or additional feature stream is required. Positioned at the backbone bottleneck, this gating mechanism ensures that \emph{before} any resolution-reducing stride, the network has already allocated extra representational capacity to locations with strong edge evidence, directly counteracting the information loss that would otherwise disproportionately harm small objects.

\noindent\textbf{Reconstruction and residual output.}  
We keep the original HF subbands unchanged and reconstruct via inverse Haar transform:
\vspace{0cm}
\begin{equation}
\label{eq:wdg_recon}
\mathbf{y} = f_{1\times 1}^{\mathrm{out}}\!\left(\mathrm{IDWT}(\widetilde{\mathbf{x}}_{LL},\mathbf{x}_{LH},\mathbf{x}_{HL},\mathbf{x}_{HH})\right).
\vspace{0cm}
\end{equation}
A residual connection $\mathbf{y}\leftarrow\mathbf{x}+\mathbf{y}$ is applied when channels match. All refinement operates at $\frac{H}{2}{\times}\frac{W}{2}$, adding $<$5\% FLOPs; WDG thus serves as a drop-in bottleneck at high-resolution stages (C2/C3) where small objects occupy the largest relative area.

\begin{figure}[t]
  \centering
  \includegraphics[width=\columnwidth]{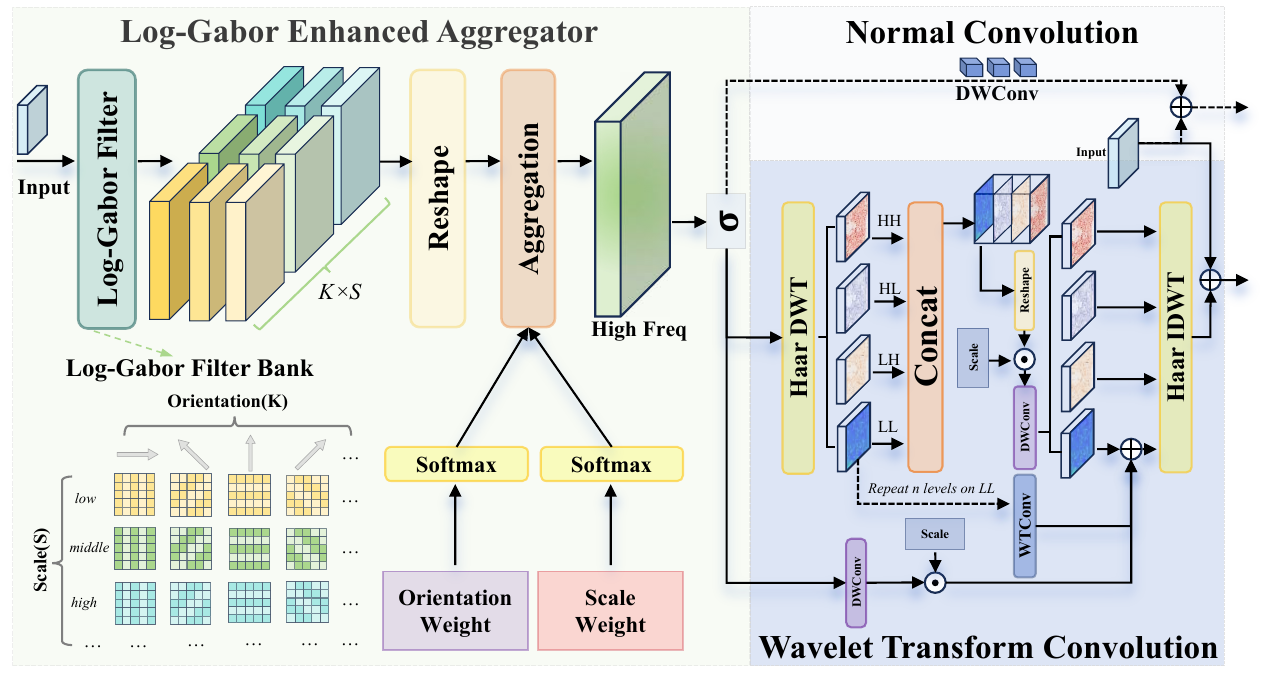}
  \caption{\textbf{Architecture of Log-Gabor Enhancer (LGE) and its WTConv variant (LGE-W).} LGE captures directional high-frequency residuals via log-gabor filters and learnable aggregation, injecting them through a skip pathway to prevent feature dilution.}
  \label{fig:lge}
  \vspace{-0.5cm}
  \end{figure}
\subsection{Log-Gabor Enhancer (LGE) and WTConv Variant (LGE-W)}
\begin{figure*}[t]
  \centering
  \includegraphics[width=\textwidth]{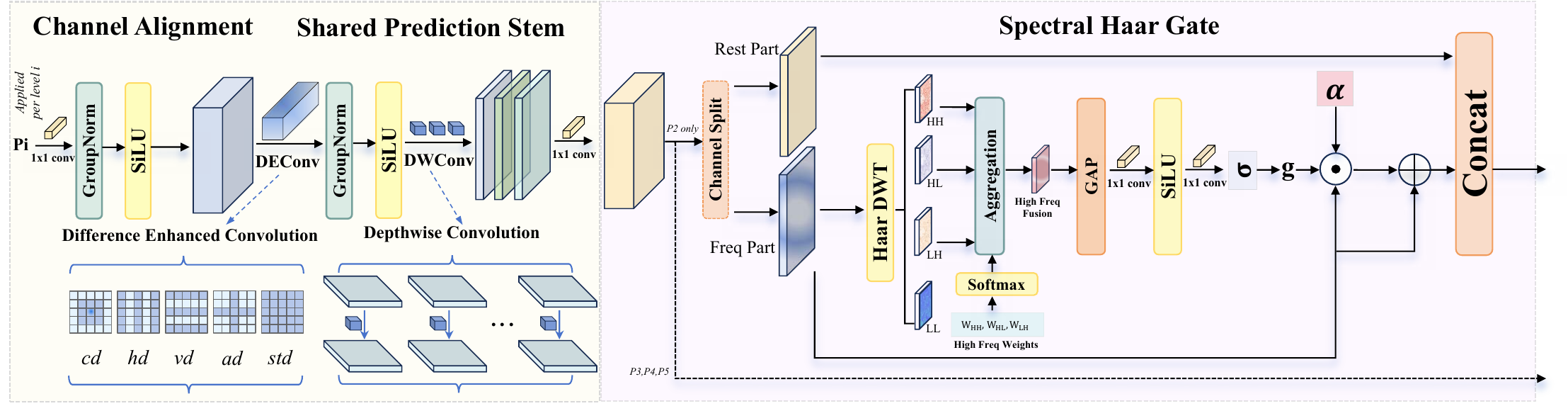}
  \caption{Architecture of the Frequency-Driven Head (FDHead).}
  \label{fig:fdhead}
  \vspace{-0.5cm}
  \end{figure*}

\noindent\textbf{Design rationale.}  
Top-down fusion averages out orientation-specific high-frequency components, diluting the directional edge cues that small objects rely on for discrimination. LGE restores this lost anisotropy by decomposing features with Log-Gabor filter bank and re-weighting orientations via learnable importance, amplifying boundary-aligned responses before downstream processing (Fig.~\ref{fig:lge}).

\noindent\textbf{Log-Gabor filter bank.}
Given $\mathbf{x}\in\mathbb{R}^{C\times H\times W}$, we apply a fixed Log-Gabor filter bank via depthwise convolution:
\vspace{0cm}
\begin{equation}
\label{eq:lgn_lgf}
\mathbf{h}^{(c)}_{s,k} = \mathbf{x}^{(c)} * \mathbf{g}_{s,k},
\vspace{0cm}
 \end{equation}
where $\mathbf{g}_{s,k}$ is a non-learnable kernel indexed by scale $s$ and orientation $k$ (see Appendix~\ref{sec:appendix_lge_details}). Log-Gabor kernels exhibit \emph{zero DC response} and \emph{optimal bandwidth} in the log-frequency domain, making them more robust to scale variation than single-order edge filters (Sobel, Laplacian). Fixed filters make LGE add no learnable filter parameters, preventing memorization of dataset-specific textures.

\noindent\textbf{Learnable aggregation.}
LGE aggregates the $K{\times}S$ subbands with learnable orientation/scale importance:
\vspace{0cm}
\begin{equation}
\label{eq:lgn_agg}
\mathbf{h}^{(c)} = \sum_{s=1}^{S}\sum_{k=1}^{K} \operatorname{softmax}(\boldsymbol{\alpha})_{s}\,\operatorname{softmax}(\boldsymbol{\beta})_{k}\,\mathbf{h}^{(c)}_{s,k},
\vspace{0cm}
\end{equation}
where $\boldsymbol{\alpha}\in\mathbb{R}^{S}$ and $\boldsymbol{\beta}\in\mathbb{R}^{K}$ are learnable logits. A sigmoid-gated global scale $\gamma$ and a local mixing operator $f_{\mathrm{mix}}$ (depthwise $3{\times}3$ conv) produce the final residual:
\vspace{0cm}
\begin{equation}
\label{eq:lgn_out}
\mathbf{y} = \mathbf{x}_{\mathrm{skip}} + f_{\mathrm{mix}}\bigl(\sigma(\gamma)\,\mathbf{h}\bigr).
\vspace{0cm}
\end{equation}
We adopt small $K{=}2$ orientations and $S{=}1$ scale as a lightweight default; ablations (Appendix~\ref{sec:appendix_e_topk}) confirm that gains saturate quickly while cost grows super-linearly on high-resolution maps, making this configuration an effective efficiency--accuracy trade-off for a neck-stage plug-in.

\noindent\textbf{Wavelet variant (LGE-W).}
At the highest-resolution neck layer (P3$\to$P2), feature maps are $4{\times}$ larger and small objects occupy the largest area---yet a standard $3{\times}3$ mixing conv sees only a tiny fraction of each object's context. We therefore replace $f_{\mathrm{mix}}$ with a \emph{Wavelet-Transform Convolution} (WTConv):
\vspace{0cm}
\begin{equation}
\label{eq:wtconv_main}
\mathrm{WTConv}(\mathbf{z})=\mathcal{S}_0\mathcal{D}_0(\mathbf{z})+\textstyle\sum_{i=1}^{L}\mathcal{S}_i\,\mathrm{IDWT}_i(\mathcal{D}_{4,i}(\mathrm{DWT}_i(\mathbf{z})))\,.
\end{equation}
where $\mathcal{D}_0$ is a spatial depthwise conv and $\{\mathcal{S}_i\}$ are learnable scales (see Appendix~\ref{sec:appendix_lge_details}). By operating in the wavelet domain, WTConv aggregates multi-scale context with a larger effective receptive field while keeping parameters compact---critical for capturing the sparse but informative boundary pixels of small objects. We apply LGE-W only at this single layer to balance overhead and benefit.
 
 \subsection{Frequency-Driven Head (FDHead)}

\noindent\textbf{Design rationale.}  
FDHead extends the frequency-guided paradigm to the detection head by using wavelet subband energy as a \emph{boundary-confidence prior}: high-frequency magnitude indicates where edges are sharp, and we amplify box predictions at those locations while leaving classification untouched. This injects a boundary-sensitive bias into regression without adding frequency transforms to the dense prediction path (Fig.~\ref{fig:fdhead}).

\noindent\textbf{Shared prediction stem.}
Each level $\mathbf{x}_i$ ($P2$--$P4$) is aligned to $C_h$ channels and refined by a shared stack:
\vspace{0cm}
\begin{equation}
\mathbf{f}_i = \mathrm{DW{\text -}PW}\bigl(\mathrm{DEConv}(\mathrm{Conv}_{1{\times}1}(\mathbf{x}_i))\bigr),
\vspace{0cm}
\end{equation}
where DEConv is a re-parameterizable $3{\times}3$ conv fusing five directional-difference kernels into one at inference, embedding edge-detection priors with no extra latency (Appendix~\ref{sec:appendix_fdhead_details}).

\noindent\textbf{Spectral Haar Gate (SHG, P2 box branch only).}
At $P2$, we split $\mathbf{f}_i$ into a frequency subset $\mathbf{f}_a$ and the remainder $\mathbf{f}_b$. A fixed Haar DWT extracts high-frequency subbands, whose energy is aggregated into a channel-wise gate that amplifies boundary-sensitive features for more precise localization:
\begin{equation}
\label{eq:fdhead_hfgate}
\begin{gathered}
\mathbf{h} = \textstyle\sum_{k\in\{LH,HL,HH\}} \operatorname{softmax}(\boldsymbol{\omega})_k\,|\mathbf{f}_k|,\\
\mathbf{s} = \operatorname{GAP}(\mathbf{h}),\quad \mathbf{g} = \mathrm{Gate}(\mathbf{s}),\quad
\widetilde{\mathbf{f}}_a = \mathbf{f}_a\odot(1+\alpha\,\mathbf{g}),
\end{gathered}
\end{equation}
where $\operatorname{GAP}(\cdot)$ is global average pooling and $\mathrm{Gate}(\cdot)$ is a lightweight channel gate implemented as $1\times 1$ conv $\rightarrow$ SiLU $\rightarrow$ $1\times 1$ conv $\rightarrow$ sigmoid. The gated $[\widetilde{\mathbf{f}}_a,\mathbf{f}_b]$ is fed \emph{only to the box branch}; classification uses $\mathbf{f}_i$ directly.

\noindent$\cdot$ \emph{Box-only modulation.} We keep classification on the ungated feature to preserve stable low-frequency semantics, while injecting SHG only into the regression stream where boundary sharpness is the dominant cue for precise offsets.

\noindent$\cdot$ \emph{P2-only application.} SHG is applied only at the finest level where high-frequency energy is still informative and spatially aligned to tiny boundaries; on coarser levels, such signals are largely attenuated and gating becomes redundant.
 
 \noindent\textbf{Box/class prediction and decoding.}
 FDHead predicts per-location class logits and distributional box offsets (DFL) as
 \vspace{0cm}
 \begin{equation}
 \label{eq:fdhead_pred}
 \mathbf{b}_i = \mathrm{Scale}_i\bigl(\mathcal{H}_{\mathrm{box}}(\widetilde{\mathbf{f}}_i)\bigr),\qquad
 \mathbf{p}_i = \mathcal{H}_{\mathrm{cls}}(\mathbf{f}_i),
 \vspace{0cm}
 \end{equation}
 and decodes boxes by $\widehat{\mathbf{B}}=\mathrm{dist2bbox}(\mathrm{DFL}(\mathbf{b}),\mathbf{A})\cdot\mathbf{s}$ with anchors $\mathbf{A}$ and strides $\mathbf{s}$. This design targets small objects by frequency-gating only the finest level while keeping the remaining head computation shared and lightweight.

 \section{Experiment}

 \begin{table*}[t]
  \caption{\textbf{Across-architecture study on VisDrone2019, UAVDT, TinyPerson, and DOTAv1.} Superscript \textsuperscript{\tiny +} denotes modified versions. Progressively deeper colors highlight improvements ($\Delta\geq0.005$), indicating larger gains over the unenhanced architecture.}
  \label{tab:across_arch}
  \centering
  \setlength\tabcolsep{0.1pt}
    \renewcommand{\arraystretch}{0.82}
    \fontsize{5pt}{6pt}\selectfont
    \resizebox{\textwidth}{!}{
    \begin{tabular}{@{}>{\raggedright\arraybackslash}p{8.5em} @{}*{7}{p{0.75cm}}p{0.7cm}p{0.7cm}p{0.7cm}p{0.35cm}p{0.7cm}@{}}
    \toprule
    \textbf{Architecture}
    & \multicolumn{7}{c}{\tiny\textbf{Performance/mAP50}}
    & \multicolumn{5}{c}{\centering\tiny\textbf{Efficiency}} \\[-2.2pt]
    \cmidrule[0.5pt](l{-0.5em}r{1em}){2-8}\cmidrule[0.5pt](l{-0.5em}r{1em}){10-13}
    & \multicolumn{2}{l}{\textbf{VisDrone2019}}
    & \multicolumn{2}{l}{\textbf{UAVDT}}
    & \multicolumn{2}{l}{\textbf{TinyPerson}}
    & \multicolumn{1}{l}{\textbf{DOTAv1}}
    & & & & & \\[-2.2pt]
    \cmidrule[0.5pt](l{-0.5em}r{1em}){2-3}\cmidrule[0.5pt](l{-0.5em}r{1em}){4-5}\cmidrule[0.5pt](l{-0.5em}r{1em}){6-7}\cmidrule[0.5pt](l{-0.5em}r{1em}){8-8}
    & \cellcolor{blue!5}{\fontsize{4pt}{5pt}\selectfont\textbf{Val}}
    & \cellcolor{blue!5}{\fontsize{4pt}{5pt}\selectfont\textbf{Test}}
    & \cellcolor{blue!5}{\fontsize{4pt}{5pt}\selectfont\textbf{Val}}
    & \cellcolor{blue!5}{\fontsize{4pt}{5pt}\selectfont\textbf{Test}}
    & \cellcolor{blue!5}{\fontsize{4pt}{5pt}\selectfont\textbf{Val}}
    & \cellcolor{blue!5}{\fontsize{4pt}{5pt}\selectfont\textbf{Test}}
    & \cellcolor{blue!5}{\fontsize{4pt}{5pt}\selectfont\textbf{Val}}
    & \cellcolor{blue!5}{\fontsize{4pt}{5pt}\selectfont\textbf{Input}}
    & \cellcolor{blue!5}{\fontsize{4pt}{5pt}\selectfont\textbf{Params/M}}
    & \cellcolor{blue!5}{\fontsize{4pt}{5pt}\selectfont\textbf{GFLOPs}}
    & \cellcolor{blue!5}{\fontsize{4pt}{5pt}\selectfont\textbf{Size/MB}}\\[-2.2pt]
    \midrule
    
    \textbf{YOLOv11-S}
    & \mapcellSI{0.384}{} & \mapcellSI{0.311}{} & \mapcellSI{0.875}{} & \mapcellSI{0.910}{} & \mapcellSI{0.264}{} & \mapcellSI{0.222}{} & \mapcellSI{0.447}{}
    & \fontsize{4pt}{5pt}\selectfont 640 & \fontsize{4pt}{5pt}\selectfont 9.4 & \fontsize{4pt}{5pt}\selectfont 21.6 & \fontsize{4pt}{5pt}\selectfont 18.3 \\
    \textbf{DERNet-S}
    & \mapcellSI{0.398}{0.014} & \mapcellSI{0.316}{0.005} & \mapcellSI{0.844}{} & \mapcellSI{0.909}{} & \mapcellSI{0.298}{0.034} & \mapcellSI{0.252}{0.030} & \mapcellSI{0.420}{}
    & \fontsize{4pt}{5pt}\selectfont 640 & \effcell{9.4}{1.3}{86.2} & \effcell{21.6}{13.3}{38.4} & \effcell{18.3}{3.4}{81.4} \\
    \textbf{YOLOv11-M}
    & \mapcellSI{0.442}{} & \mapcellSI{0.353}{} & \mapcellSI{0.889}{} & \mapcellSI{0.926}{} & \mapcellSI{0.283}{} & \mapcellSI{0.239}{} & \mapcellSI{0.470}{}
    & \fontsize{4pt}{5pt}\selectfont 640 & \fontsize{4pt}{5pt}\selectfont 20.0 & \fontsize{4pt}{5pt}\selectfont 67.7 & \fontsize{4pt}{5pt}\selectfont 38.7 \\
    \textbf{DERNet-M}
    & \mapcellSI{0.447}{0.005} & \mapcellSI{0.362}{0.009} & \mapcellSI{0.875}{} & \mapcellSI{0.903}{} & \mapcellSI{0.324}{0.041} & \mapcellSI{0.274}{0.035} & \mapcellSI{0.452}{}
    & \fontsize{4pt}{5pt}\selectfont 640 & \effcell{20.0}{2.9}{85.5} & \effcell{67.7}{29.4}{56.6} & \effcell{38.7}{7.1}{81.7} \\
    
    \specialrule{0.45pt}{0pt}{0pt}
    \specialrule{0.15pt}{0.20ex}{0.4ex}
    
    \textbf{RTMDet-R2-T}
    & \mapcellSI{0.375}{} & \mapcellSI{0.309}{} & \mapcellSI{0.817}{} & \mapcellSI{0.859}{} & \mapcellSI{0.275}{} & \mapcellSI{0.240}{} & \mapcellSI{0.497}{}
    & \fontsize{4pt}{5pt}\selectfont 1024 & \fontsize{4pt}{5pt}\selectfont 6.2 & \fontsize{4pt}{5pt}\selectfont 27.7 & \fontsize{4pt}{5pt}\selectfont 37.1 \\
    \textbf{RTMDet-R2-T}\textsuperscript{\tiny +}
    & \mapcellSI{0.448}{0.073} & \mapcellSI{0.358}{0.049} & \mapcellSI{0.828}{0.011} & \mapcellSI{0.866}{0.007} & \mapcellSI{0.292}{0.017} & \mapcellSI{0.304}{0.064} & \mapcellSI{0.496}{}
    & \fontsize{4pt}{5pt}\selectfont 1024 & \effcell{6.2}{4.5}{27.4} & \effcell{27.7}{20.8}{24.9} & \effcell{37.1}{29.0}{21.8} \\
    \textbf{RTMDet-R2-S}
    & \mapcellSI{0.391}{} & \mapcellSI{0.325}{} & \mapcellSI{0.853}{} & \mapcellSI{0.884}{} & \mapcellSI{0.290}{} & \mapcellSI{0.244}{} & \mapcellSI{0.527}{}
    & \fontsize{4pt}{5pt}\selectfont 1024 & \fontsize{4pt}{5pt}\selectfont 11.2 & \fontsize{4pt}{5pt}\selectfont 50.5 & \fontsize{4pt}{5pt}\selectfont 59.9 \\
    \textbf{RTMDet-R2-S}\textsuperscript{\tiny +}
    & \mapcellSI{0.485}{0.094} & \mapcellSI{0.394}{0.069} & \mapcellSI{0.846}{} & \mapcellSI{0.881}{} & \mapcellSI{0.333}{0.043} & \mapcellSI{0.320}{0.076} & \mapcellSI{0.567}{0.040}
    & \fontsize{4pt}{5pt}\selectfont 1024 & \effcell{11.2}{7.3}{34.8} & \effcell{50.5}{38.3}{24.2} & \effcell{59.9}{40.8}{31.9} \\
    
    \specialrule{0.45pt}{0pt}{0pt}
    \specialrule{0.15pt}{0.20ex}{0.4ex}
    
    \textbf{PP-PicoDet-S}
    & \mapcellSI{0.282}{} & \mapcellSI{0.245}{} & \mapcellSI{0.591}{} & \mapcellSI{0.611}{} & \mapcellSI{0.151}{} & \mapcellSI{0.132}{} & \mapcellSI{0.169}{}
    & \fontsize{4pt}{5pt}\selectfont 416 & \fontsize{4pt}{5pt}\selectfont 3.5 & \fontsize{4pt}{5pt}\selectfont 2.2 & \fontsize{4pt}{5pt}\selectfont 13.4 \\
    \textbf{PP-PicoDet-S}\textsuperscript{\tiny +}
    & \mapcellSI{0.274}{} & \mapcellSI{0.251}{0.006} & \mapcellSI{0.583}{} & \mapcellSI{0.605}{} & \mapcellSI{0.175}{0.024} & \mapcellSI{0.149}{0.017} & \mapcellSI{0.183}{0.014}
    & \fontsize{4pt}{5pt}\selectfont 416 & \effcell{3.5}{2.7}{22.9} & \effcell{2.2}{1.6}{27.3} & \effcell{13.4}{11.1}{17.2} \\
    \textbf{PP-PicoDet-L}
    & \mapcellSI{0.357}{} & \mapcellSI{0.334}{} & \mapcellSI{0.631}{} & \mapcellSI{0.614}{} & \mapcellSI{0.213}{} & \mapcellSI{0.189}{} & \mapcellSI{0.205}{}
    & \fontsize{4pt}{5pt}\selectfont 640 & \fontsize{4pt}{5pt}\selectfont 5.9 & \fontsize{4pt}{5pt}\selectfont 8.4 & \fontsize{4pt}{5pt}\selectfont 22.3 \\
    \textbf{PP-PicoDet-L}\textsuperscript{\tiny +}
    & \mapcellSI{0.361}{0.004} & \mapcellSI{0.339}{0.005} & \mapcellSI{0.628}{} & \mapcellSI{0.617}{0.003} & \mapcellSI{0.237}{0.024} & \mapcellSI{0.202}{0.013} & \mapcellSI{0.212}{0.007}
    & \fontsize{4pt}{5pt}\selectfont 640 & \effcell{5.9}{4.5}{23.7} & \effcell{8.4}{6.6}{21.4} & \effcell{22.3}{17.3}{22.4} \\
    
    \specialrule{0.45pt}{0pt}{0pt}
    \specialrule{0.15pt}{0.20ex}{0.4ex}
    
    \textbf{RT-DETR-R18}
    & \mapcellSI{0.362}{} & \mapcellSI{0.301}{} & \mapcellSI{0.717}{} & \mapcellSI{0.691}{} & \mapcellSI{0.127}{} & \mapcellSI{0.117}{} & \mapcellSI{0.204}{}
    & \fontsize{4pt}{5pt}\selectfont 640 & \fontsize{4pt}{5pt}\selectfont 20.1 & \fontsize{4pt}{5pt}\selectfont 61.2 & \fontsize{4pt}{5pt}\selectfont 307.8 \\
    \textbf{RT-DETR-R18}\textsuperscript{\tiny +}
    & \mapcellSI{0.358}{} & \mapcellSI{0.299}{} & \mapcellSI{0.792}{0.075} & \mapcellSI{0.749}{0.058} & \mapcellSI{0.147}{0.020} & \mapcellSI{0.162}{0.045} & \mapcellSI{0.213}{0.009}
    & \fontsize{4pt}{5pt}\selectfont 640 & \effcell{20.1}{13.8}{31.3} & \effcell{61.2}{45.4}{25.8} & \effcell{307.8}{213.0}{30.8} \\
    \textbf{RT-DETR-R50}
    & \mapcellSI{0.467}{} & \mapcellSI{0.395}{} & \mapcellSI{0.837}{} & \mapcellSI{0.843}{} & \mapcellSI{0.214}{} & \mapcellSI{0.181}{} & \mapcellSI{0.209}{}
    & \fontsize{4pt}{5pt}\selectfont 640 & \fontsize{4pt}{5pt}\selectfont 42.8 & \fontsize{4pt}{5pt}\selectfont 137.5 & \fontsize{4pt}{5pt}\selectfont 654.9 \\
    \textbf{RT-DETR-R50}\textsuperscript{\tiny +}
    & \mapcellSI{0.465}{} & \mapcellSI{0.392}{} & \mapcellSI{0.884}{0.047} & \mapcellSI{0.882}{0.039} & \mapcellSI{0.243}{0.029} & \mapcellSI{0.227}{0.046} & \mapcellSI{0.220}{0.011}
    & \fontsize{4pt}{5pt}\selectfont 640 & \effcell{42.8}{32.0}{25.2} & \effcell{137.5}{89.1}{35.2} & \effcell{654.9}{459.4}{29.8} \\
    
    \bottomrule
    \end{tabular}}
    \vspace{-0.2cm}
    \end{table*}
 \subsection{Datasets and Metrics}
We evaluate our framework on four benchmarks to demonstrate its robustness and cross-domain generalization: VisDrone2019 \cite{du2019visdrone}, TinyPerson \cite{yu2020scale}, UAVDT \cite{du2018unmanned}, and DOTAv1 \cite{xia2018dota}. \textbf{VisDrone2019} is our primary benchmark and is particularly challenging due to dense small objects and severe scale variation, where most targets are smaller than $50 \times 50$ pixels.

We report both accuracy and efficiency, including precision, recall, mAP50, APs, as well as the number of parameters, GFLOPs, model size, and FPS in the following tables.
 \subsection{Configuration}
Training and primary accuracy evaluation are conducted on an NVIDIA A100 GPU (40 GB VRAM), while full training hyperparameters and hardware configurations for all architectures, including FPS evaluation on NVIDIA A100 and Jetson Nano (batch size 1, post-warm-up), are provided in Appendix~\ref{sec:appendix_h_training_configs}, Tables~\ref{tab:config} and \ref{tab:training_configs}.

\section{Main Results}

\subsection{Ablation Study on YOLO-style architectures}

\begin{table*}[ht]
  \centering
  \caption{Ablation study on YOLO-style architectures.}
  \label{tab:ablation}
  \setlength{\tabcolsep}{3.5pt}  
  \renewcommand{\arraystretch}{1.20}
  \fontsize{10pt}{12pt}\selectfont
  \resizebox{\textwidth}{!}{  
    \begin{tabular}{@{}*{4}{>{\centering\arraybackslash\large}p{1.2em}}
      >{\centering\arraybackslash\large}p{3.6em} >{\centering\arraybackslash\large}p{3.6em} @{\hspace{3pt}} >{\centering\arraybackslash\large}p{6.6em}
      >{\centering\arraybackslash\large}p{3.6em} >{\centering\arraybackslash\large}p{3.6em} @{\hspace{3pt}} >{\centering\arraybackslash\large}p{6.6em}
      >{\large}c >{\large}c >{\large}c @{}}
      \toprule[1.5pt]
      \textbf{P} & \textbf{W} & \textbf{L} & \textbf{F}
      & \multicolumn{6}{c}{{\fontsize{14pt}{16pt}\selectfont\textbf{Performance}}} & \multicolumn{3}{c}{{\fontsize{14pt}{16pt}\selectfont\textbf{Efficiency}}} \\[-2.5pt]
      \cmidrule[1.25pt](lr){5-10} \cmidrule[1.25pt](lr){11-13}
      \textbf{2} & \textbf{D} & \textbf{G} & \textbf{D}
      & \multicolumn{3}{c}{{\fontsize{14pt}{16pt}\selectfont\textbf{Val}}} & \multicolumn{3}{c}{{\fontsize{14pt}{16pt}\selectfont\textbf{Test}}} & & & \\[-2.5pt]
      \cmidrule[1.15pt](lr){5-7} \cmidrule[1.15pt](lr){8-10}
      \textbf{H} & \textbf{G} & \textbf{E} & \textbf{H}
      & \cellcolor{blue!5}{\fontsize{12pt}{14pt}\selectfont\textbf{P}} & \cellcolor{blue!5}{\fontsize{12pt}{14pt}\selectfont\textbf{R}} & \cellcolor{blue!5}{\fontsize{12pt}{14pt}\selectfont\textbf{mAP50}}
      & \cellcolor{blue!5}{\fontsize{12pt}{14pt}\selectfont\textbf{P}} & \cellcolor{blue!5}{\fontsize{12pt}{14pt}\selectfont\textbf{R}} & \cellcolor{blue!5}{\fontsize{12pt}{14pt}\selectfont\textbf{mAP50}}
      & \cellcolor{blue!5}{\fontsize{12pt}{14pt}\selectfont\textbf{Params/M}} & \cellcolor{blue!5}{\fontsize{12pt}{14pt}\selectfont\textbf{GFLOPs}} & \cellcolor{blue!5}{\fontsize{12pt}{14pt}\selectfont\textbf{Model size/MB}} \\[-2.5pt]
      \midrule[1.20pt]
      $\times$ & $\times$ & $\times$ & $\times$
      & 0.496 & 0.377 & \mapcellA{0.384}{} & 0.421 & 0.337 & \mapcellA{0.311}{} & 9.4 & 21.6 & 18.3 \\
      \checkmark & $\times$ & $\times$ & $\times$
      & 0.516 & 0.408 & \mapcellA{0.423}{} & 0.453 & 0.354 & \mapcellA{0.337}{} & 6.4 & 24.7 & 12.5 \\
      \checkmark & \checkmark & $\times$ & $\times$
      & 0.543 & 0.420 & \mapcellA{0.438}{0.015} & 0.455 & 0.371 & \mapcellA{0.354}{0.017} & 6.6 & 23.5 & 13.2 \\
      \checkmark & $\times$ & \checkmark & $\times$
      & 0.547 & 0.412 & \mapcellA{0.436}{0.013} & 0.465 & 0.373 & \mapcellA{0.355}{0.018} & 6.5 & 25.4 & 12.8 \\
      \checkmark & $\times$ & $\times$ & \checkmark
      & 0.551 & 0.435 & \mapcellA{0.454}{0.031} & 0.478 & 0.375 & \mapcellA{0.364}{0.027} & 3.8 & 30.5 & 8.4 \\
      \checkmark & \checkmark & \checkmark & $\times$
      & 0.547 & 0.419 & \mapcellA{0.445}{0.022} & 0.466 & 0.376 & \mapcellA{0.360}{0.023} & 3.9 & 25.5 & 8.2 \\
      \checkmark & \checkmark & $\times$ & \checkmark
      & 0.559 & 0.435 & \mapcellA{0.464}{0.041} & 0.487 & 0.383 & \mapcellA{0.370}{0.033} & 3.9 & 28.5 & 8.6 \\
      \checkmark & $\times$ & \checkmark & \checkmark
      & 0.549 & 0.437 & \mapcellA{0.457}{0.034} & 0.470 & 0.378 & \mapcellA{0.365}{0.028} & 3.8 & 28.3 & 8.4 \\
      \checkmark & \checkmark & \checkmark & \checkmark
      & 0.552 & 0.429 & \mapcellA{0.458}{0.035} & 0.484 & 0.381 & \mapcellA{0.370}{0.033} & 3.8 & 26.3 & 8.6 \\
      \multicolumn{4}{c}{\textbf{DERNet-S}}
      & \cellcolor{blue!5}0.503 & \cellcolor{blue!5}0.382 & \cellcolor{blue!5}\mapcellA{0.398}{0.014} & \cellcolor{blue!5}0.420 & \cellcolor{blue!5}0.344 & \cellcolor{blue!5}\mapcellA{0.316}{0.005}       & \cellcolor{blue!5}1.3 & \cellcolor{blue!5}13.3 & \cellcolor{blue!5}3.4 \\
      \bottomrule[1.5pt]
    \end{tabular}
  }
  \vspace{-0.3cm}
\end{table*}

\noindent Table~\ref{tab:ablation} details the stepwise integration of our proposed modules into YOLOv11-S baseline ($0.384/0.311$ mAP$_{50}$ on val/test). Each component contributes consistent gains ($>0.013$), with FDHead yielding the most significant boost ($>0.03$). The full integration culminates in a peak performance of 0.458/0.370 mAP$_{50}$, validating the cumulative efficacy of our design. Notably, the DERNet-S variant, optimized with C3\_Faster and channel reduction, still outperforms the baseline while slashing parameters to just 1.3M.
\subsection{Across-architecture Study}

\noindent As shown in Table~\ref{tab:across_arch}, for CNN-based models, YOLO-style detectors achieve comparable or superior performance to baseline while reducing parameters by over 85\%; RTMDet-R2 variants exceed original versions on most datasets, with RTMDet-R2-S\textsuperscript{\tiny +} achieving a notable 0.076 mAP$_{50}$ gain on TinyPerson test set and 0.04 on both TinyPerson and DOTAv1 val set. Transformer-based detectors also show significant improvements, with RT-DETR-R18\textsuperscript{\tiny +} reducing parameters by 31.3\% and achieving a 0.075 accuracy gain on UAVDT val set. These results indicate that our frequency-guided design offers consistent accuracy gains under tight parameter and GFLOPs budgets across both CNN and Transformer architectures.

\begin{table*}[t]
  \caption{\textbf{TIDE error decomposition.} Values are reported as A/B for paired settings in each row.}
  \label{tab:tide}
  \centering
  \setlength\tabcolsep{1.8pt}
  \renewcommand{\arraystretch}{1.18}
  \fontsize{3pt}{4pt}\selectfont
  \resizebox{\textwidth}{!}{
      \begin{tabular}{@{}>{\raggedright\arraybackslash}p{1.2cm}*{8}{>{\raggedright\arraybackslash}p{0.5cm}}>{\raggedright\arraybackslash}p{0.95cm}@{}}
      \toprule
      {\fontsize{4pt}{5pt}\selectfont\textbf{Architecture}}
      & \multicolumn{6}{c}{\fontsize{4pt}{5pt}\selectfont\textbf{Main Errors}}
  & \multicolumn{2}{c}{\fontsize{4pt}{5pt}\selectfont\textbf{Special Errors}}
  & {\fontsize{4pt}{5pt}\selectfont\textbf{Small bbox}}\\[-2.2pt]
  \cmidrule(lr){2-7}\cmidrule(lr){8-9}\cmidrule{10-10}
  & \cellcolor{blue!5}{\fontsize{3.5pt}{4.5pt}\selectfont\textbf{Cls}}
  & \cellcolor{blue!5}{\fontsize{3.5pt}{4.5pt}\selectfont\textbf{Loc}}
  & \cellcolor{blue!5}{\fontsize{3.5pt}{4.5pt}\selectfont\textbf{Both}}
  & \cellcolor{blue!5}{\fontsize{3.5pt}{4.5pt}\selectfont\textbf{Dupe}}
  & \cellcolor{blue!5}{\fontsize{3.5pt}{4.5pt}\selectfont\textbf{Bkg}}
  & \cellcolor{blue!5}{\fontsize{3.5pt}{4.5pt}\selectfont\textbf{Miss}}
  & \cellcolor{blue!5}{\fontsize{3.5pt}{4.5pt}\selectfont\textbf{FalsePos}}
  & \cellcolor{blue!5}{\fontsize{3.5pt}{4.5pt}\selectfont\textbf{FalseNeg}}
  & \cellcolor{blue!5}{\fontsize{3.5pt}{4.5pt}\selectfont\textbf{APs}}\\[-2.2pt]
      \midrule
      \noalign{\vspace{-1.5ex}}
      {\fontsize{3pt}{4pt}\selectfont\textbf{YOLOv11-S/M}} & 22.03/22.04 & 3.97/4.51 & 0.27/0.32 & 0.09/0.08 & 3.07/3.65 & 7.06/8.08 & 13.47/13.77 & 29.13/32.96 & 12.57/16.13 \\
      {\fontsize{3pt}{4pt}\selectfont\textbf{DERNet-S/M}} & 21.84/21.95\tidedown & 3.53/3.88\tidedown & 0.36/0.43 & 0.28/0.25 & 2.46/2.88\tidedown & 6.72/6.90\tidedown & 17.95/18.23 & 25.10/26.34\tidedown & 15.72/18.40\tideupAB{3.15}{2.27} \\\specialrule{0.45pt}{0pt}{0pt}
      \specialrule{0.15pt}{0.20ex}{0pt}
      {\fontsize{3pt}{4pt}\selectfont\textbf{RT-DETR-R18/R50}} & 16.54/15.87 & 3.39/3.98 & 0.67/0.75 & 0.50/0.45 & 2.68/3.16 & 15.02/14.94 & 15.85/16.79 & 30.24/30.01 & 23.85/30.04 \\
      {\fontsize{3pt}{4pt}\selectfont\textbf{RT-DETR-R18}\textsuperscript{\fontsize{2pt}{2.5pt}\selectfont +}\textbf{/R50}\textsuperscript{\fontsize{2pt}{2.5pt}\selectfont +}}  & 16.23/15.26\tidedown & 3.26/3.54\tidedown & 0.70/0.77 & 0.77/0.73 & 2.45/2.87\tidedown & 13.90/13.62\tidedown & 16.22/16.91 & 29.55/29.34\tidedown & 25.60/32.37\tideupAB{1.75}{2.33} \\
      \specialrule{0.45pt}{0pt}{0pt}
      \specialrule{0.15pt}{0.20ex}{0.4ex}
      {\fontsize{3pt}{4pt}\selectfont\textbf{RTMDet-R2-T/S}} & 8.56/10.72 & 6.66/6.69 & 0.55/0.59 & 0.00/0.00 & 3.99/4.27 & 6.82/6.50 & 27.60/27.11 & 19.47/19.98 & 16.06/17.45 \\
      {\fontsize{3pt}{4pt}\selectfont\textbf{RTMDet-R2-T}\textsuperscript{\fontsize{2pt}{2.5pt}\selectfont +}\textbf{/S}\textsuperscript{\fontsize{2pt}{2.5pt}\selectfont +}} & 10.30/10.91 & 6.60/6.60\tidedown & 0.53/0.57\tidedown & 0.00/0.00 & 4.28/4.55 & 6.71/6.44\tidedown & 28.31/27.86 & 18.22/19.18\tidedown & 20.10/23.61\tideupAB{4.04}{6.16} \\
      \specialrule{0.45pt}{0pt}{0pt}
      \specialrule{0.15pt}{0.20ex}{0.4ex}
      {\fontsize{3pt}{4pt}\selectfont\textbf{PP-PicoDet-S/L}} & 13.04/8.34 & 3.42/2.86 & 0.19/0.13 & 0.02/0.01 & 1.82/1.62 & 11.84/13.79 & 14.33/14.99 & 29.99/30.56 & 8.67/9.13 \\
      {\fontsize{3pt}{4pt}\selectfont\textbf{PP-PicoDet-S}\textsuperscript{\fontsize{2pt}{2.5pt}\selectfont +}\textbf{/L}\textsuperscript{\fontsize{2pt}{2.5pt}\selectfont +}} & 10.01/7.29\tidedown & 3.28/2.28\tidedown & 0.28/0.15 & 0.02/0.01 & 1.60/1.57\tidedown & 10.92/12.32\tidedown & 11.10/10.48\tidedown & 27.32/28.71\tidedown & 10.91/11.75\tideupAB{2.24}{2.62} \\
      \noalign{\vspace{-1.5ex}}
      \bottomrule
      \end{tabular}}
      \vspace{-0.2cm}
  \end{table*}

\subsection{Comparison with State-of-the-art}

\noindent We benchmark DERNet against state-of-the-art detectors in Table~\ref{tab:sota_comparison}. DERNet establishes a superior efficiency-accuracy trade-off: it matches YOLOv11 performance with $\sim$\textbf{85\% fewer parameters} and, notably, \textbf{outperforms the recent FBRT-YOLO (AAAI 2025) across all key metrics}~\cite{fbrt-yolo} (mAP$_{50}$, AP$_S$, and GFLOPs), validating the efficacy of our design.

\begin{table}[h]
  \centering
  \caption{Comparison with state-of-the-art methods on VisDrone2019.}
  \label{tab:sota_comparison}
  \setlength{\tabcolsep}{3pt}
  \renewcommand{\arraystretch}{1.5}
  \fontsize{7pt}{6pt}\selectfont
  \begin{tabular}{@{}l *{4}{c} @{}}
    \toprule
    \cellcolor{blue!5}{\textbf{Models}} & \cellcolor{blue!5}{\fontsize{6pt}{7pt}\selectfont\textbf{mAP50}} & \cellcolor{blue!5}{\fontsize{6pt}{7pt}\selectfont\textbf{APs}} & \cellcolor{blue!5}{\fontsize{6pt}{7pt}\selectfont\textbf{Params/M}} & \cellcolor{blue!5}{\fontsize{6pt}{7pt}\selectfont\textbf{GFLOPs}} \\
    \midrule
    \textbf{ATSS-R50-FPN-DyHead}~{\fontsize{4.5pt}{5pt}\selectfont\cite{dai2021dyhead}} & 0.338 & 16.0 & 38.91 & 110 \\
    \textbf{GFL}~{\fontsize{4.5pt}{5pt}\selectfont\cite{gfl}} & 0.321 & 16.4 & 32.28 & 206 \\
    \textbf{TOOD-R50}~{\fontsize{4.5pt}{5pt}\selectfont\cite{tood}} & 0.338 & 16.2 & 32.04 & 199 \\
    \textbf{Cascade-RCNN-R50-FPN}~{\fontsize{4.5pt}{5pt}\selectfont\cite{cascade-rcnn}} & 0.326 & 14.9 & 69.29 & 236 \\
    \textbf{Faster-RCNN-R50-FPN-CIOU} & 0.329 & 15.5 & 41.39 & 208 \\
    \textbf{RetinaNet-R50-FPN}~{\fontsize{4.5pt}{5pt}\selectfont\cite{RetinaNet}} & 0.276 & 12.0 & 36.52 & 210 \\
    \textbf{RT-DETR-R18} & 0.301 & 14.85 & 20.1 & 61.2 \\

    \textbf{YOLOX-Tiny}~{\fontsize{4.5pt}{5pt}\selectfont\cite{yolox}} & 0.278 & 11.6 & 5.035 & 7.578 \\
    \textbf{YOLOv11-N} & 0.258 & 13.8 & 2.59 & 6.3 \\
    \textbf{YOLOv11-S} & 0.311 & 14.57 & 9.4 & 21.6 \\
    \textbf{YOLOv11-M} & 0.353 & 16.13 & 20.0 & 67.7 \\
    \textbf{YOLOv12-N}~{\fontsize{4.5pt}{5pt}\selectfont\cite{yolov12}} & 0.259 & 13.5 & 2.56 & 6.3 \\
    \textbf{YOLOv12-S} & 0.312 & 15.8 & 9.23 & 21.2 \\
    \textbf{YOLOv12-M} & 0.336 & 17.4 & 9.23 & 21.2 \\
    \textbf{YOLOv26-N} & 0.247 & 11.8 & 2.5 & 5.2 \\
    \textbf{YOLOv26-S} & 0.304 & 14.57 & 10.0 & 20.5 \\
    \textbf{YOLOv26-M} & 0.365 & 19.15 & 21.8 & 67.9 \\

    \textbf{D-Fine-N}~{\fontsize{4.5pt}{5pt}\selectfont\cite{peng2024dfine}}& 0.334 & 16.3 & 3.73 & 7.1238 \\
    \textbf{DEIM-D-Fine-N}~{\fontsize{4.5pt}{5pt}\selectfont\cite{huang2025deimdetrimprovedmatching}} & 0.322 & 16.0 & 3.73 & 7.1238 \\
    \textbf{DEIM-D-Fine-S} & 0.384 & 23.2 & 10.18 & 24.8595 \\
    \textbf{FBRT-YOLO-N} & 0.265 & 12.2 & 0.8 & 6.7 \\
    \textbf{FBRT-YOLO-S} & 0.323 & 16.2 & 2.9 & 22.9 \\
    \textbf{FBRT-YOLO-M} & 0.344 & 17.5 & 7.36 & 58.7 \\

    \midrule
    \midrule
    \cellcolor{green!5}{\textbf{DERNet-S}} & \cellcolor{green!5}0.316 & \cellcolor{green!5}15.72 & \cellcolor{green!5}1.3 & \cellcolor{green!5}13.3 \\
    \cellcolor{green!5}{\textbf{DERNet-M}} & \cellcolor{green!5}0.362 & \cellcolor{green!5}18.40 & \cellcolor{green!5}2.9 & \cellcolor{green!5}29.4 \\
    \bottomrule
  \end{tabular}
\end{table}

\subsection{Real-world inference efficiency}

\vspace{-0.15cm}
\begin{table}[H]
  \centering
  \caption{Real-world efficiency and accuracy comparison on VisDrone2019 validation set. FPS is evaluated on both NVIDIA A100 GPU and NVIDIA Jetson Nano.}
  \label{tab:fps_efficiency}
  \vspace{-0.45em}
  \setlength{\tabcolsep}{2.2pt}
  \renewcommand{\arraystretch}{1.02}
  \fontsize{5.8pt}{6.8pt}\selectfont
  \resizebox{\columnwidth}{!}{%
  \begin{tabular}{@{}lccccc@{}}
  \toprule[1.1pt]
  \cellcolor{blue!5}\textbf{Model}
  & \cellcolor{blue!5}\textbf{Params/M}
  & \cellcolor{blue!5}\textbf{GFLOPs}
  & \cellcolor{blue!5}\textbf{mAP$_{50}$}
  & \cellcolor{blue!5}\textbf{A100 FPS}
  & \cellcolor{blue!5}\textbf{Nano FPS} \\
  \midrule[0.55pt]
  WDFS-DETR~{\scriptsize\cite{liu2025wdfs}} & 19.9 & 53.7 & 47.5 & 117 & 10 \\
  WCDB-YOLO~{\scriptsize\cite{luan2026wcdb}} & 19.8 & 60.1 & 45.5 & 114 & 9 \\
  FBRT-YOLO-S~{\scriptsize\cite{fbrt-yolo}} & 2.9 & 22.9 & 42.4 & 143 & 22 \\
  D-FINE-S & 10.18 & 24.86 & 45.4 & 140 & 20 \\
  \midrule[0.55pt]
  \rowcolor{green!5}
  \textbf{DERNet-S} & \textbf{1.3} & \textbf{13.3} & \textbf{39.8} & \textbf{162} & \textbf{22} \\
  \rowcolor{green!5}
  \textbf{DERNet-M} & \textbf{2.9} & \textbf{29.4} & \textbf{44.7} & \textbf{134} & \textbf{16} \\
  \bottomrule[1.1pt]
  \end{tabular}}
  \vspace{-0.3cm}
\end{table}

\noindent As shown in Table~\ref{tab:fps_efficiency} and Appendix~\ref{sec:appendix_h_training_configs}, DERNet offers a favorable accuracy-efficiency trade-off for real-world deployment, particularly compared with prior frequency-domain methods that often improve representation quality at the cost of additional computation. DERNet-S achieves 162/22 FPS on A100/Jetson Nano, while DERNet-M reaches 44.7 val mAP$_{50}$ with 134/16 FPS, showing that our stage-aware frequency-guided design translates spectral enhancement into consistent accuracy gains without compromising practical inference efficiency, even on resource-constrained edge hardware.

\section{Analyses and Discussion}
\subsection{Error Decomposition with TIDE: Disentangling Miss, Localization, and False Positives}

We employ TIDE~\cite{tide} to decompose detection errors (Table~\ref{tab:tide}). Beyond substantial AP$_S$ gains (e.g., \textbf{+4.04\%/6.16\%} for RTMDet-R2 variants), the error distribution reveals a clear causal link: \textbf{improved high-frequency preservation correlates directly with reduced Miss and Localization errors}. Recovering faint spectral signatures helps distinguish tiny targets from background clutter (lowering \textbf{Miss/FalseNeg}), while sharpening boundary-aligned energy explicitly refines box regression (lowering \textbf{Loc}). This supports a causal interpretation: spectral recovery primarily reduces detection errors by restoring faint signatures and sharpening boundary evidence.

\addtocounter{figure}{1}
\begin{figure*}[t]
  \centering
  \setlength{\tabcolsep}{0pt}%
  \renewcommand{\arraystretch}{0}%
  \fontsize{5pt}{6pt}\selectfont
  \begin{tabular}{@{}c@{\hspace{0pt}}c@{\hspace{0pt}}c@{\hspace{0pt}}c@{\hspace{0pt}}c@{\hspace{0pt}}c@{\hspace{0pt}}c@{\hspace{0pt}}c@{\hspace{0pt}}c@{}}
  \includegraphics[width=0.11\textwidth,height=0.10\textwidth]{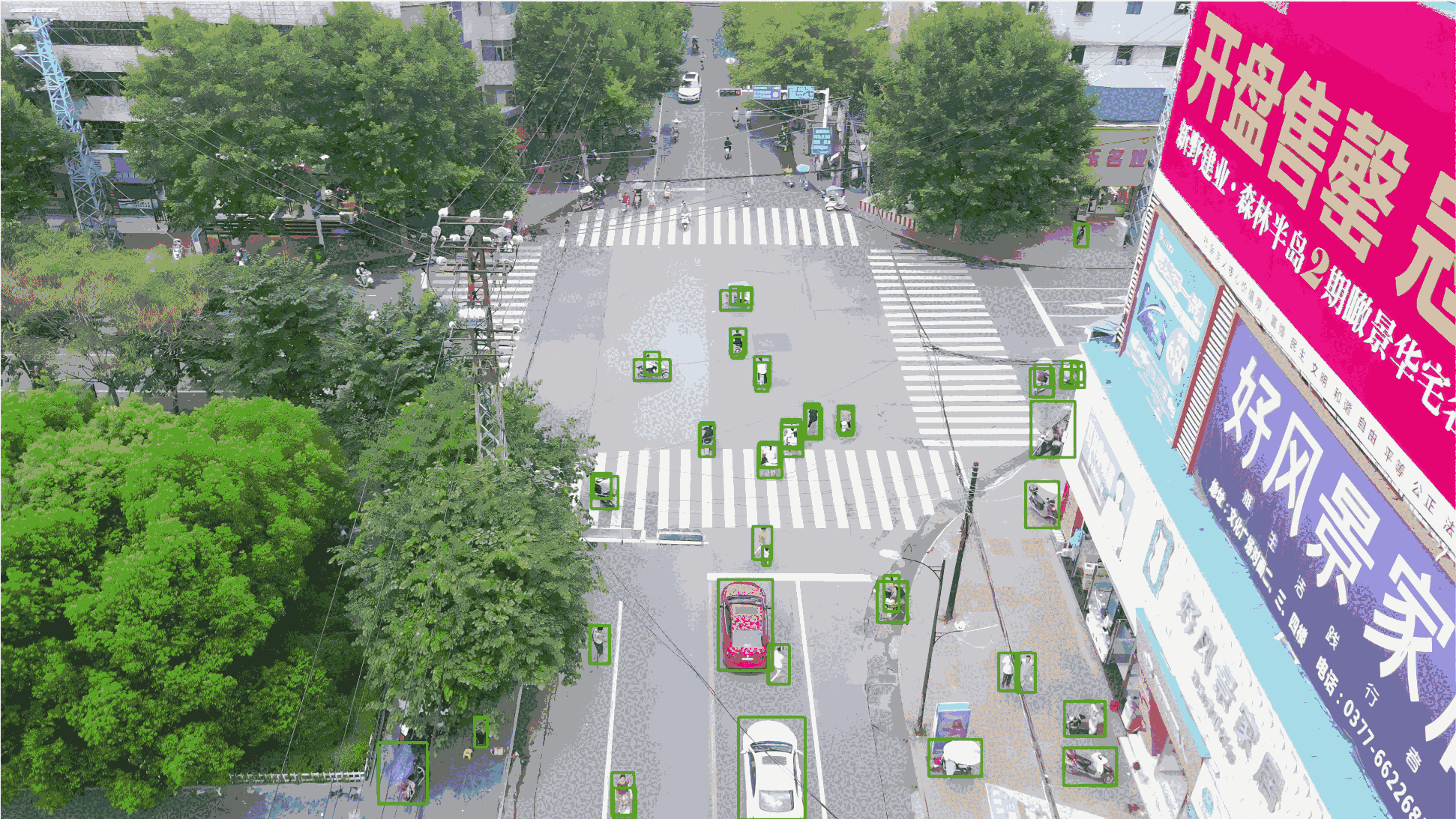} & 
  \includegraphics[width=0.11\textwidth,height=0.10\textwidth]{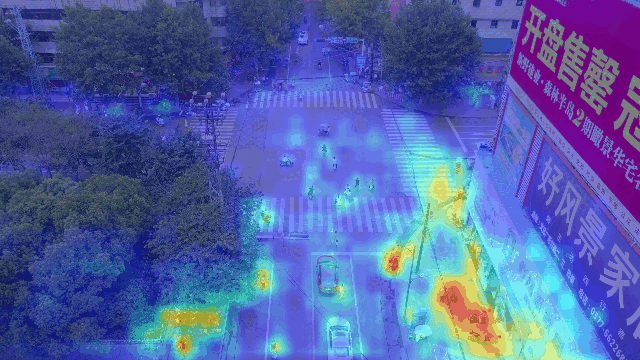} & 
  \includegraphics[width=0.11\textwidth,height=0.10\textwidth]{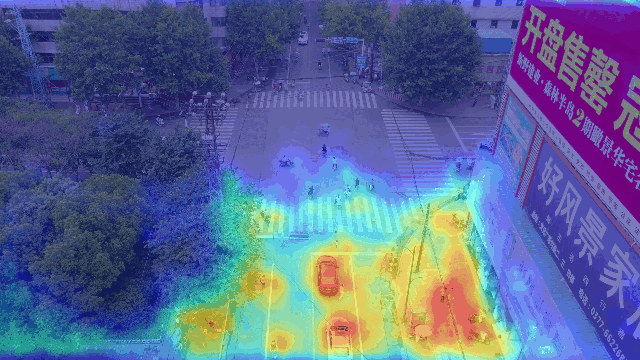} & 
  \includegraphics[width=0.11\textwidth,height=0.10\textwidth]{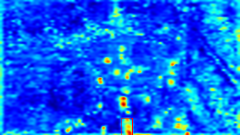} & 
  \includegraphics[width=0.11\textwidth,height=0.10\textwidth]{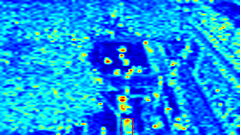} & 
  \includegraphics[width=0.11\textwidth,height=0.10\textwidth]{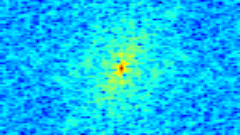} & 
  \includegraphics[width=0.11\textwidth,height=0.10\textwidth]{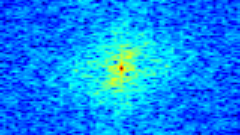} & 
  \includegraphics[width=0.11\textwidth,height=0.10\textwidth]{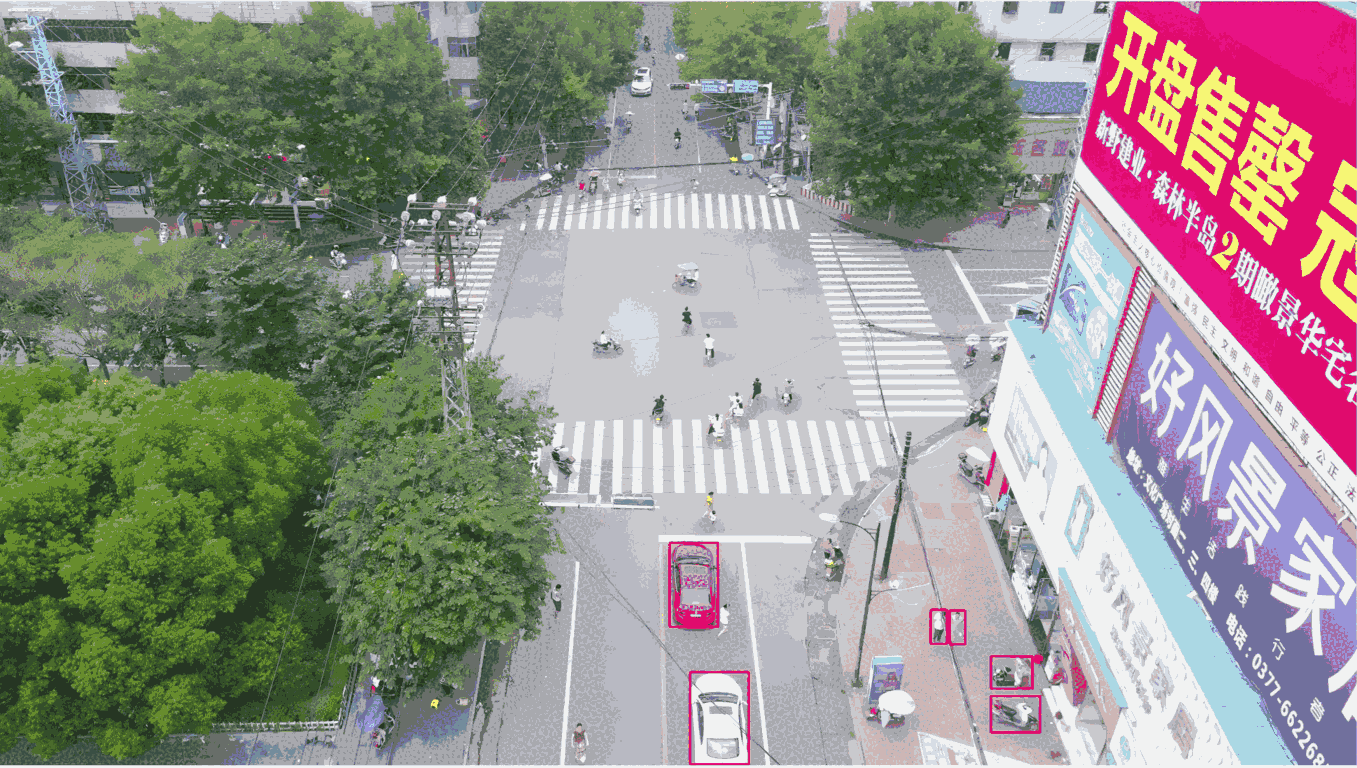} & 
  \includegraphics[width=0.11\textwidth,height=0.10\textwidth]{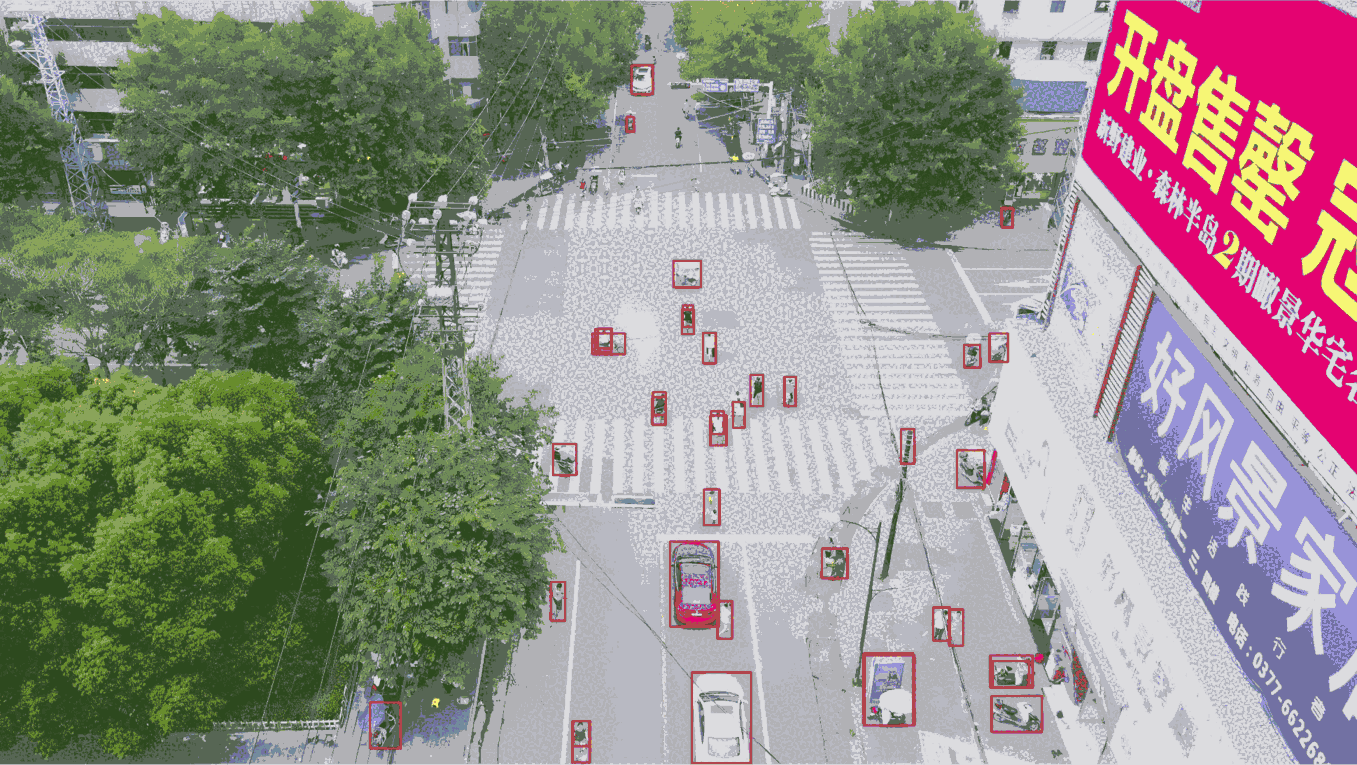} \\
  \includegraphics[width=0.11\textwidth,height=0.10\textwidth]{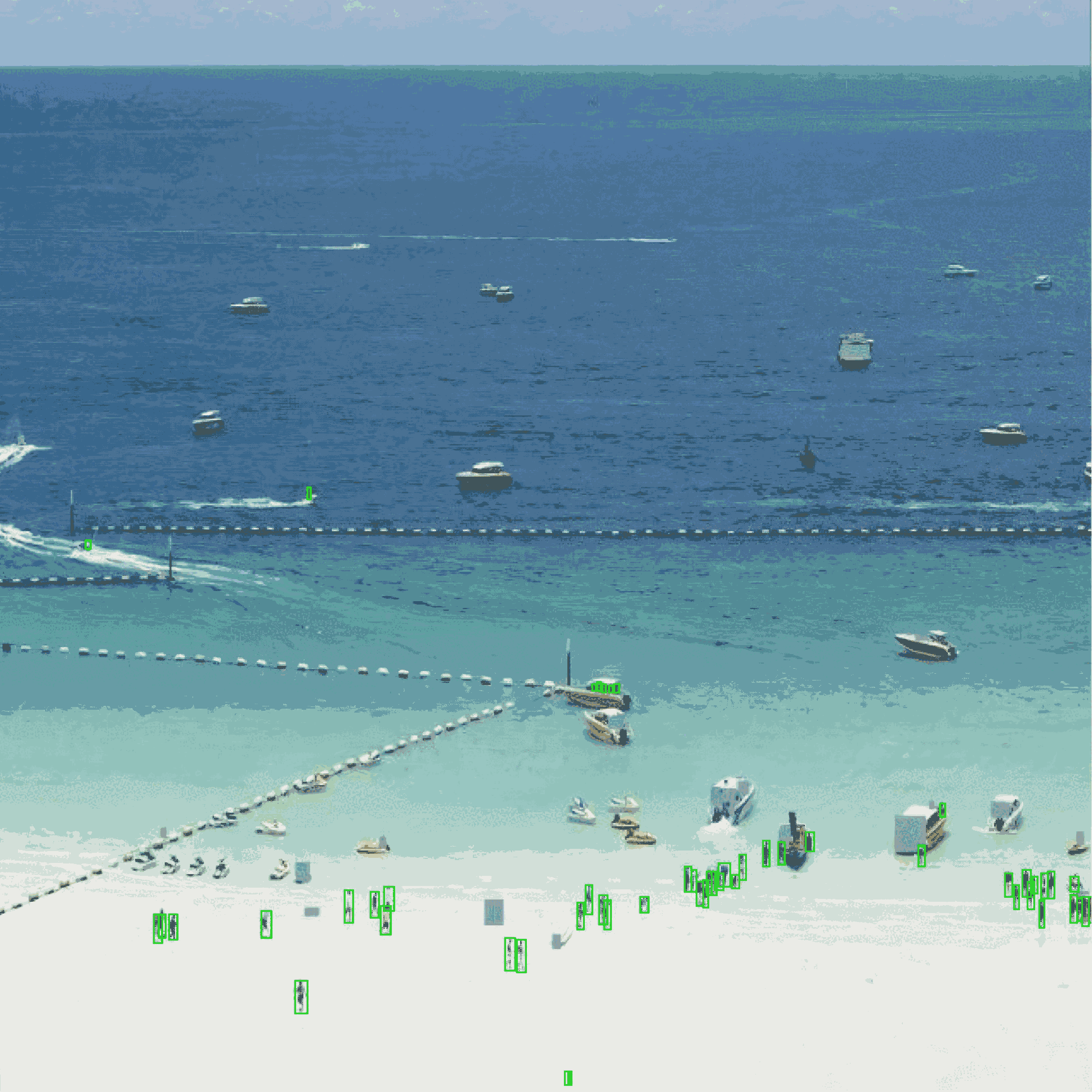} & 
  \includegraphics[width=0.11\textwidth,height=0.10\textwidth]{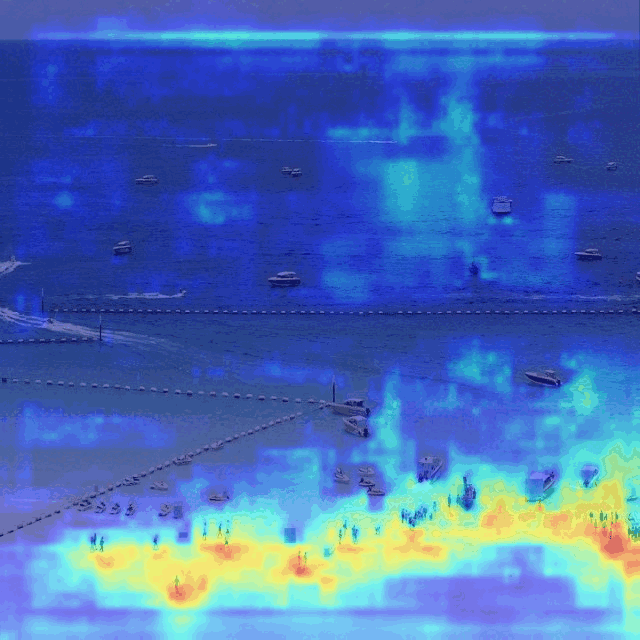} & 
  \includegraphics[width=0.11\textwidth,height=0.10\textwidth]{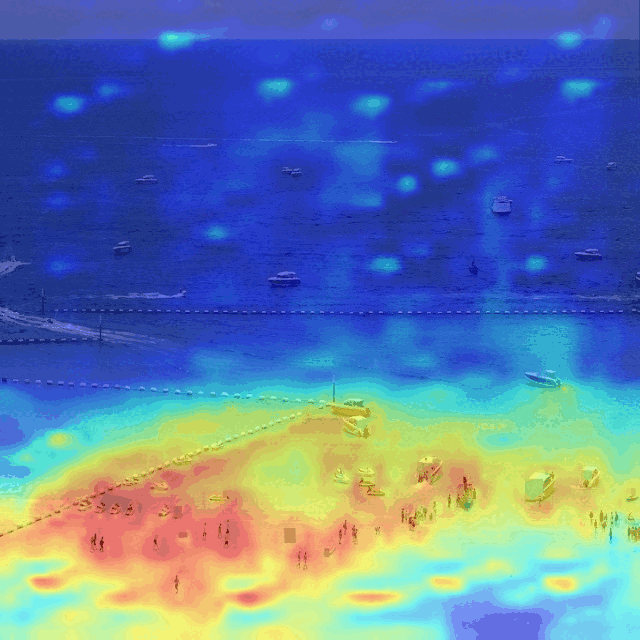} & 
  \includegraphics[width=0.11\textwidth,height=0.10\textwidth]{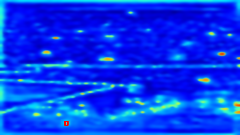} & 
  \includegraphics[width=0.11\textwidth,height=0.10\textwidth]{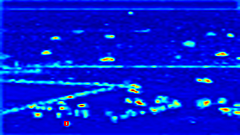} & 
  \includegraphics[width=0.11\textwidth,height=0.10\textwidth]{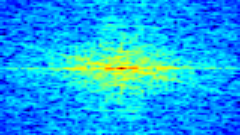} & 
  \includegraphics[width=0.11\textwidth,height=0.10\textwidth]{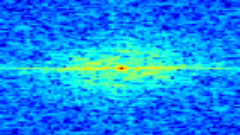} & 
  \includegraphics[width=0.11\textwidth,height=0.10\textwidth]{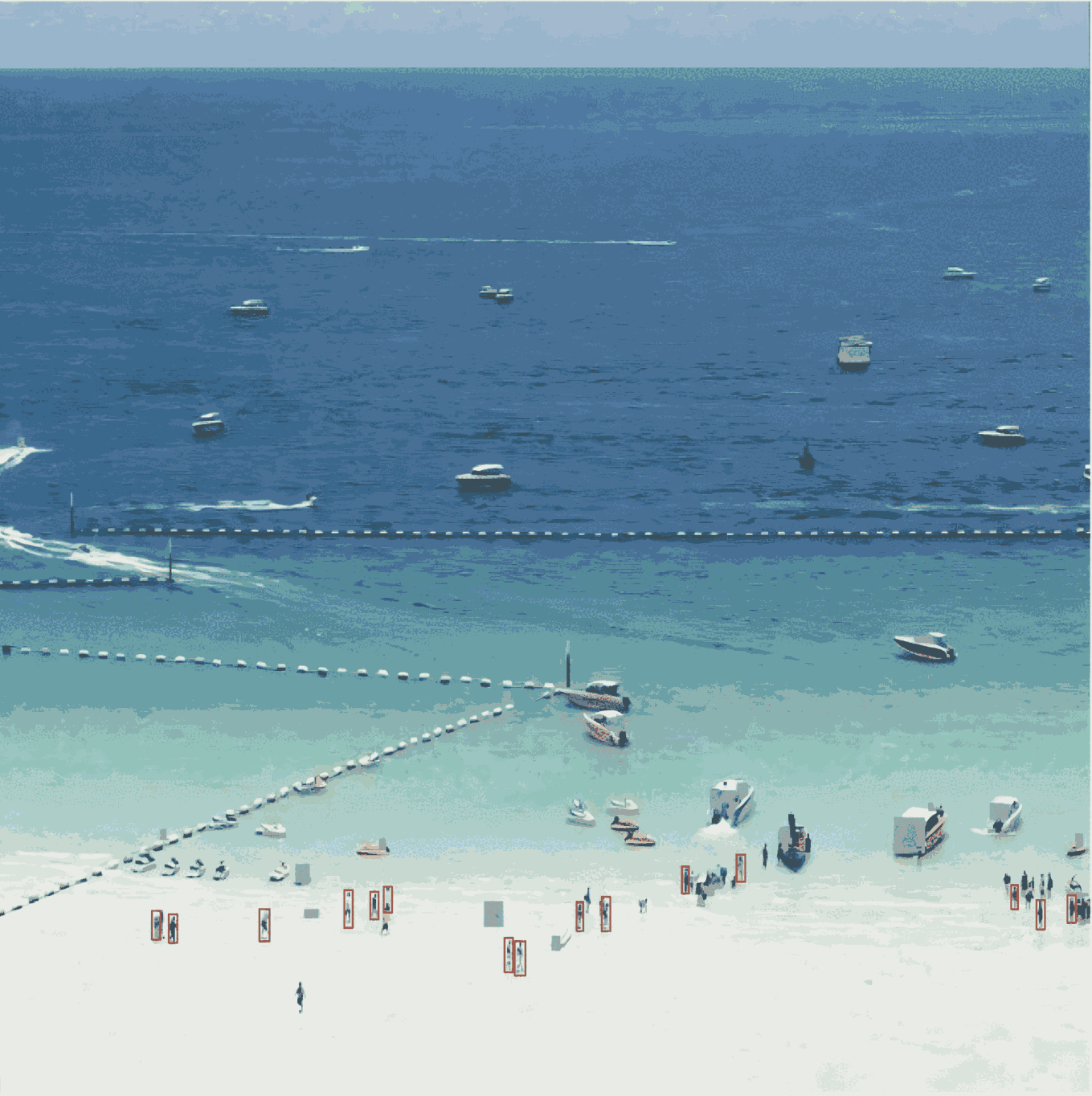} & 
  \includegraphics[width=0.11\textwidth,height=0.10\textwidth]{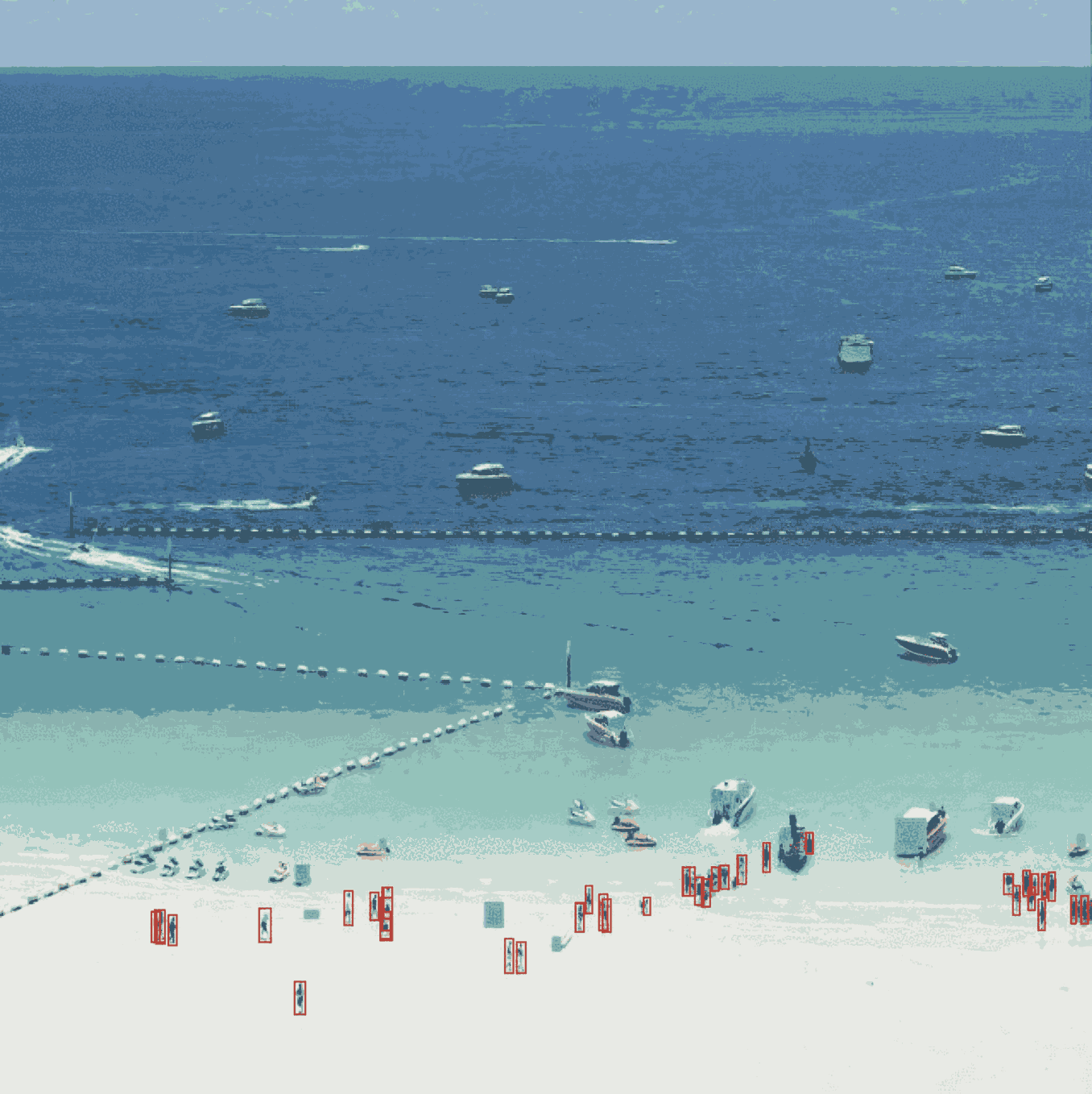} \\[0.7ex]
  \textbf{GT} & \textbf{Heatmaps (Baseline)} & \textbf{Heatmaps (DERNet-S)} & \textbf{Feature (Baseline)} & \textbf{Feature (DERNet-S)} & \textbf{Spectrum (Baseline)} & \textbf{Spectrum (DERNet-S)} & \textbf{Results (Baseline)} & \textbf{Results (DERNet-S)} \\
  \end{tabular}
  \caption{Visualization comparison between baseline and improved models on VisDrone2019 (top row) and TinyPerson (bottom row).}
  \label{fig:visualization}
\end{figure*}

\subsection{Spectral Diagnostics: Layer-wise High-Frequency Preservation and Reconstruction}
We analyze high-frequency preservation across stages on VisDrone2019 and TinyPerson via 2D FFT, averaging spectral magnitude, and partitioning frequency bands at \textbf{1/6} and \textbf{1/3} of the maximum frequency radius to compute the high-frequency energy ratio. Fig.~\ref{fig:frequency_ratio} shows that DER operator demonstrates a progressive high-frequency enrichment from Backbone-C3 to the Head, culminating in an energy ratio increase of over \textbf{0.05} at the final prediction stage, suggesting that DER prevents the spectral collapse phenomenon typically observed in deep CNNs.\vspace{-0.2cm}

\balance

\section{Conclusion}

We present DER, a unified frequency-guided operator interface for small-object detection, and show that its stage-specific instantiations (WDG/LGE/FDHead) improve accuracy-efficiency trade-offs across heterogeneous detectors. By decoupling feature modeling from resolution reduction in the backbone, neck, and head, DER provides a plug-and-play abstraction that consistently boosts small-object AP across four benchmarks without expensive architectural overhauls. DERNet delivers cleaner predictions in dense scenes (Fig.~\ref{fig:visualization}) while using substantially fewer parameters.

Severely occluded, tightly clustered, and extremely small targets remain challenging (Appendix~\ref{sec:appendix_g_error_analysis}, Fig.~\ref{fig:qualitative_analysis}).

Future work will pursue system-aware implementations (e.g., operator fusion, re-parameterization, and kernel-level optimization) to better align DER-style modules with modern accelerators, and extend frequency-guided representation learning to broader \textbf{small-object-dense} vision tasks such as instance segmentation, tracking, and counting. We also view DER as a step toward a more principled spectral theory of detection, where architectural choices can be compared and optimized directly in terms of how they preserve, re-distribute, or destroy high-frequency evidence.
\addtocounter{figure}{-2}
\begin{figure}[h]
  \centering
  \includegraphics[width=\columnwidth,trim={22pt 18pt 22pt 18pt},clip]{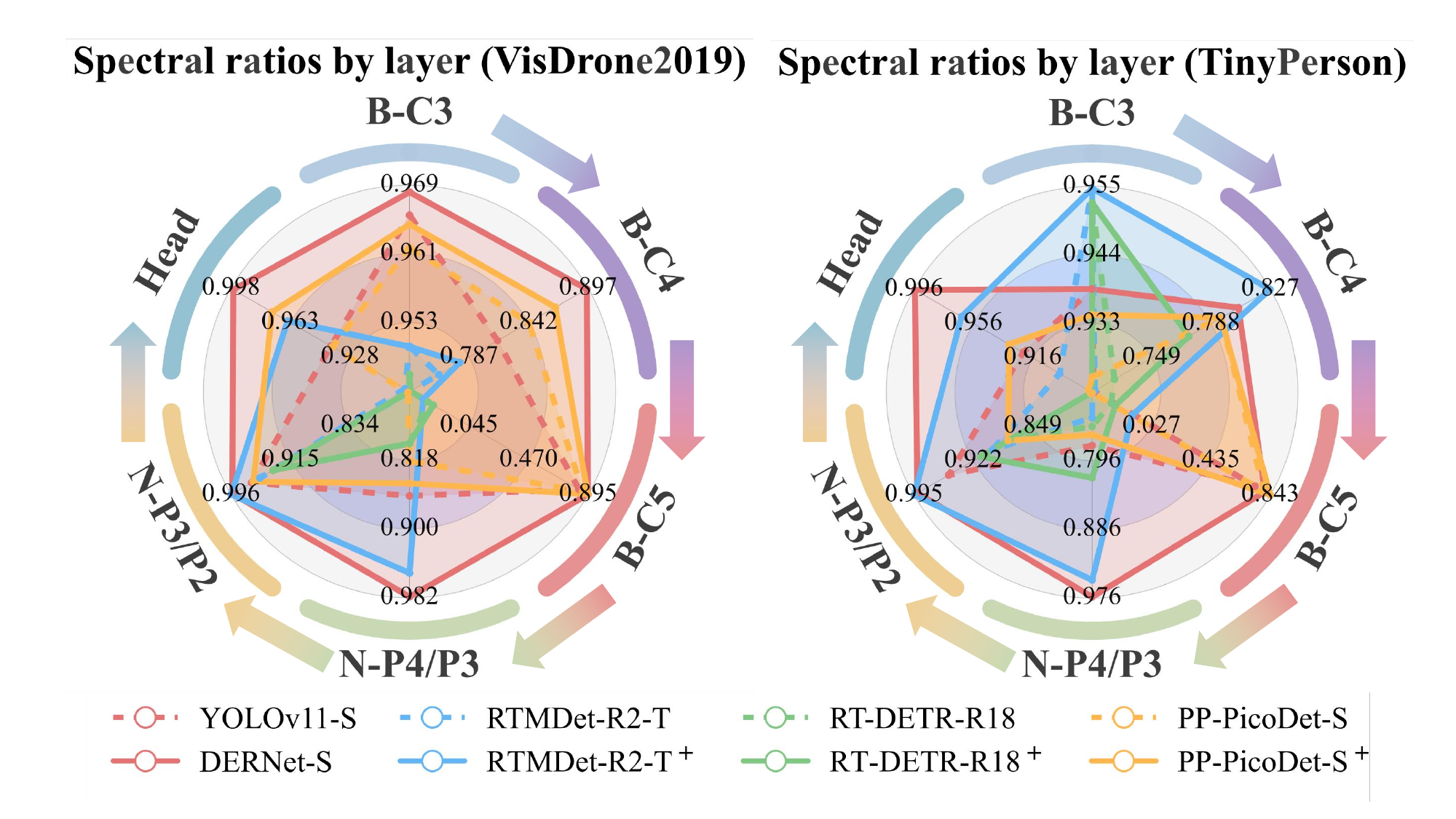}
  \vspace{-0.5em}
  \caption{High-Frequency Ratio Analysis on VisDrone2019 validation set (Left) and TinyPerson validation set (Right).}
  \label{fig:frequency_ratio}
  \vspace{-0.55cm}
\end{figure}
\addtocounter{figure}{1}

\clearpage
\twocolumn
\newpage

\newpage
\appendix
\onecolumn

\section{Appendix A: Stability Analysis}
\label{sec:appendix_a_stability}

We evaluate the training stability and reproducibility of DER-enhanced models by monitoring loss convergence and gradient statistics across multiple training runs with different random seeds. To assess reproducibility, we conduct three independent training runs for YOLOv11-S/M and DERNet-S/M models on VisDrone2019 validation set, reporting the mean and standard deviation of key metrics including precision (P), recall (R), small object AP (AP$_S$), and AP at IoU=0.5 (AP$_{50}$). Our experiments demonstrate that the frequency-guided representation learner maintains stable training dynamics with consistent convergence patterns.

\begin{table}[h]
\centering
\caption{Reproducibility analysis of YOLOv11 and DERNet models with different random seeds on VisDrone2019 validation set. Results are reported as mean${\pm}$std over three runs.}
\label{tab:stability}
\vspace{0.1cm}
\resizebox{0.95\columnwidth}{!}{%
\begin{tabular}{@{}lccccccc@{}}
\toprule
\cellcolor{blue!5}\textbf{Model} & \cellcolor{blue!5}\textbf{Val P} & \cellcolor{blue!5}\textbf{Test P} & \cellcolor{blue!5}\textbf{Val R} & \cellcolor{blue!5}\textbf{Test R} & \cellcolor{blue!5}\textbf{Val AP$_{50}$} & \cellcolor{blue!5}\textbf{Test AP$_{50}$} & \cellcolor{blue!5}\textbf{AP$_S$} \\
\midrule
\textbf{YOLOv11-S} & $0.485{\pm}0.011$ & $0.410{\pm}0.007$ & $0.368{\pm}0.008$ & $0.330{\pm}0.006$ & $0.375{\pm}0.007$ & $0.303{\pm}0.005$ & $12.0{\pm}1.1$ \\
\textbf{DERNet-S} & $0.502{\pm}0.011$ & $0.419{\pm}0.007$ & $0.381{\pm}0.008$ & $0.341{\pm}0.005$ & $0.393{\pm}0.007$ & $0.314{\pm}0.004$ & $15.12{\pm}1.4$ \\
\textbf{YOLOv11-M} & $0.515{\pm}0.012$ & $0.415{\pm}0.008$ & $0.452{\pm}0.010$ & $0.363{\pm}0.006$ & $0.432{\pm}0.008$ & $0.345{\pm}0.005$ & $15.5{\pm}1.4$ \\
\textbf{DERNet-M} & $0.548{\pm}0.014$ & $0.426{\pm}0.008$ & $0.466{\pm}0.012$ & $0.375{\pm}0.006$ & $0.444{\pm}0.008$ & $0.359{\pm}0.005$ & $17.99{\pm}1.6$ \\
\bottomrule
\end{tabular}}
\end{table}

The low standard deviations across all metrics validate the reliability of our experimental results and demonstrate the robustness of our frequency-guided approach for small object detection.

\section{Appendix B: Computational Cost Breakdown}
\label{sec:appendix_b_computational_cost}

We provide a detailed breakdown of the parameter counts and GFLOPs for the key modules in the baseline YOLOv11 models and the substituted/inserted modules in our DERNet variants. The comparison highlights the efficiency of our frequency-guided modules.

\subsection{Small-Scale Models (S-Series)}

Tables~\ref{tab:costSBaseline} and~\ref{tab:costSImproved} present the layer-wise cost breakdown for YOLOv11-S and DERNet-S, where DERNet-S reduces the overall parameter count and computational cost by shrinking channel widths in non-critical modules and using C3\_Faster in backbone, while the proposed frequency-guided enhancement blocks compensate for the resulting capacity reduction and preserve detection accuracy (from 9.4M / 21.6 GFLOPs to 1.3M / 13.3 GFLOPs).

\begin{table}[h!]
\centering
\caption{Cost breakdown for key modules in YOLOv11-S (Baseline).}
\label{tab:costSBaseline}
\vspace{0.1cm}
\resizebox{0.75\columnwidth}{!}{%
\fontsize{9pt}{10pt}\selectfont
\begin{tabular}{@{}llrrp{6cm}@{}}
\toprule
\cellcolor{blue!5}\fontsize{9pt}{10pt}\selectfont\textbf{Layer} & \cellcolor{blue!5}\fontsize{9pt}{10pt}\selectfont\textbf{Type} & \cellcolor{blue!5}\fontsize{9pt}{10pt}\selectfont\textbf{Params} & \cellcolor{blue!5}\fontsize{9pt}{10pt}\selectfont\textbf{GFLOPs} & \cellcolor{blue!5}\fontsize{9pt}{10pt}\selectfont\textbf{Input Shape} \\
\midrule
2 & C3k2 & 26,080 & 1.3599 & (1, 64, 160, 160) \\
4 & C3k2 & 103,360 & 1.3353 & (1, 128, 80, 80) \\
6 & C3k2 & 346,112 & 1.1141 & (1, 256, 40, 40) \\
8 & C3k2 & 1,380,352 & 1.1076 & (1, 512, 20, 20) \\
13 & C3k2 & 443,776 & 1.4246 & (1, 768, 40, 40) \\
16 & C3k2 & 127,680 & 1.6433 & (1, 512, 80, 80) \\
19 & C3k2 & 345,472 & 1.1100 & (1, 384, 40, 40) \\
22 & C3k2 & 1,511,424 & 1.2124 & (1, 768, 20, 20) \\
23 & Detect\_head & 850,368 & 3.2460 & (1,144,80,80), (1,144,40,40), (1,144,20,20) \\
\bottomrule
\end{tabular}}
\end{table}

\begin{table}[h!]
\centering
\caption{Cost breakdown for substituted/inserted modules in DERNet-S.}
\label{tab:costSImproved}
\vspace{0.1cm}
\resizebox{0.75\columnwidth}{!}{%
\fontsize{9pt}{10pt}\selectfont
\begin{tabular}{@{}llrrp{6cm}@{}}
\toprule
\cellcolor{blue!5}\fontsize{9pt}{10pt}\selectfont\textbf{Layer} & \cellcolor{blue!5}\fontsize{9pt}{10pt}\selectfont\textbf{Type} & \cellcolor{blue!5}\fontsize{9pt}{10pt}\selectfont\textbf{Params} & \cellcolor{blue!5}\fontsize{9pt}{10pt}\selectfont\textbf{GFLOPs} & \cellcolor{blue!5}\fontsize{9pt}{10pt}\selectfont\textbf{Input Shape} \\
\midrule
2 & C3\_WDC & 18,336 & 0.6611 & (1, 48, 160, 160) \\
4 & C3\_WDC & 51,680 & 0.4639 & (1, 96, 80, 80) \\
12 & LGE & 3,843 & 0.0135 & (1, 192, 40, 40) \\
16 & LGE & 3,203 & 0.0451 & (1, 160, 80, 80) \\
20 & LGE-W & 8,835 & 0.1327 & (1, 96, 160, 160) \\
22 & C3\_WDC & 18,880 & 0.7762 & (1, 192, 160, 160) \\
26 & FDHead & 346,705 & 7.4027 & (1, 144, 160, 160), (1, 144, 80, 80) \\
\bottomrule
\end{tabular}}
\end{table}

\subsection{Medium-Scale Models (M-Series)}

Tables~\ref{tab:costMBaseline} and~\ref{tab:costMImproved} present the layer-wise cost breakdown for YOLOv11-M and DERNet-M, where DERNet-M reduces the overall parameter count and computational cost by shrinking channel widths in non-critical modules and using C3\_Faster in backbone, while the proposed frequency-guided enhancement blocks compensate for the resulting capacity reduction and preserve detection accuracy (from 20.0M / 67.7 GFLOPs to 2.9M / 29.4 GFLOPs).

\begin{table}[h!]
\centering
\caption{Cost breakdown for key modules in YOLOv11-M (Baseline).}
\label{tab:costMBaseline}
\vspace{0.1cm}
\resizebox{0.75\columnwidth}{!}{%
\fontsize{9pt}{10pt}\selectfont
\begin{tabular}{@{}llrrp{6cm}@{}}
\toprule
\cellcolor{blue!5}\fontsize{9pt}{10pt}\selectfont\textbf{Layer} & \cellcolor{blue!5}\fontsize{9pt}{10pt}\selectfont\textbf{Type} & \cellcolor{blue!5}\fontsize{9pt}{10pt}\selectfont\textbf{Params} & \cellcolor{blue!5}\fontsize{9pt}{10pt}\selectfont\textbf{GFLOPs} & \cellcolor{blue!5}\fontsize{9pt}{10pt}\selectfont\textbf{Input Shape} \\
\midrule
2 & C3k2 & 111,872 & 5.7934 & (1, 128, 160, 160) \\
4 & C3k2 & 444,928 & 5.7278 & (1, 256, 80, 80) \\
6 & C3k2 & 1,380,352 & 4.4302 & (1, 512, 40, 40) \\
8 & C3k2 & 1,380,352 & 1.1076 & (1, 512, 20, 20) \\
13 & C3k2 & 1,642,496 & 5.2691 & (1, 1024, 40, 40) \\
16 & C3k2 & 542,720 & 6.9730 & (1, 1024, 80, 80) \\
19 & C3k2 & 1,511,424 & 4.8497 & (1, 768, 40, 40) \\
22 & C3k2 & 1,642,496 & 1.3173 & (1, 1024, 20, 20) \\
23 & Detect\_head & 1,472,704 & 6.7326 & (1,144,80,80), (1,144,40,40), (1,144,20,20) \\
\bottomrule
\end{tabular}}
\end{table}

\begin{table}[h!]
\centering
\caption{Cost breakdown for substituted/inserted modules in DERNet-M.}
\label{tab:costMImproved}
\vspace{0.1cm}
\resizebox{0.75\columnwidth}{!}{%
\fontsize{9pt}{10pt}\selectfont
\begin{tabular}{@{}llrrp{6cm}@{}}
\toprule
\cellcolor{blue!5}\fontsize{9pt}{10pt}\selectfont\textbf{Layer} & \cellcolor{blue!5}\fontsize{9pt}{10pt}\selectfont\textbf{Type} & \cellcolor{blue!5}\fontsize{9pt}{10pt}\selectfont\textbf{Params} & \cellcolor{blue!5}\fontsize{9pt}{10pt}\selectfont\textbf{GFLOPs} & \cellcolor{blue!5}\fontsize{9pt}{10pt}\selectfont\textbf{Input Shape} \\
\midrule
2 & C3\_WDC & 40,896 & 1.4561 & (1, 72, 160, 160) \\
4 & C3\_WDC & 115,680 & 1.0307 & (1, 144, 80, 80) \\
12 & LGE & 5,763 & 0.0203 & (1, 288, 40, 40) \\
16 & LGE & 4,803 & 0.0676 & (1, 240, 80, 80) \\
20 & LGE-W & 13,251 & 0.1991 & (1, 144, 160, 160) \\
22 & C3\_WDC & 42,180 & 1.7203 & (1, 288, 160, 160) \\
26 & FDHead & 767,690 & 15.9091 & (1, 144, 160, 160), (1, 144, 80, 80) \\
\bottomrule
\end{tabular}}
\end{table}

\section{Appendix C: Implementation Details}
\label{sec:appendix_c_implementation_details}

Across all four detector families (YOLOv11/DERNet, RT-DETR, RTMDet-R2, PP-PicoDet), our frequency-guided modules are implemented as plug-and-play, configuration-controlled refinements of baseline operators, enabling reproducible and isolated ablations without source-code changes. We adopt a unified spectral design: in the backbone, \textbf{Wavelet-Difference Gate (WDG)} modules are placed at downsampling bottlenecks to decompose features into low- and high-frequency components and use the latter to gate the former, suppressing aliasing of small-object details into low-frequency channels; in the neck, \textbf{Log-Gabor Enhancer (LGE/LGE-W)} modules are attached to multi-scale fusion stages to restore and amplify directional high-frequency components that would otherwise be smoothed by top-down aggregation; and in the detection head, \textbf{Frequency-Driven Head (FDHead)} modules gate regression features with high-frequency responses so that localization is biased toward boundary-consistent regions. The concrete insertion points and module names for each detector family are summarized in Table~\ref{tab:unified_insertion_points}.

\begin{table}[h]
\centering
\caption{Unified insertion points of WDG-, LGE-, and FDHead-style modules across different detector families.}
\label{tab:unified_insertion_points}
\vspace{0.1cm}
\fontsize{7pt}{8pt}\selectfont
\setlength{\tabcolsep}{3pt}
\renewcommand{\arraystretch}{1.05}
\begin{tabular}{@{}p{0.18\textwidth}p{0.28\textwidth}p{0.27\textwidth}p{0.23\textwidth}@{}}
\toprule
\cellcolor{blue!5}\textbf{Model family} & \cellcolor{blue!5}\textbf{Backbone (WDG)} & \cellcolor{blue!5}\textbf{Neck (LGE)} & \cellcolor{blue!5}\textbf{Head (FDHead)} \\
\midrule
\textbf{YOLOv11 / DERNet-S/M} & WDG replaces bottlenecks @ C2/C3 & LGE filtering in FPN neck & FDHead gating @ P2 head \\
\textbf{RTMDet-R2-T/S} & WDG replaces bottlenecks @ C4/C5 & LGE modules @ upsampling paths & FDHead gating @ P2 level \\
\textbf{PP-PicoDet-S/L} & WDG replaces bottlenecks @ C2/C3 (LCNet) & LGE filters @ FPN outputs & FDHead gating @ P2 reg branch \\
\textbf{RT-DETR-R18/R50} & WDG replaces bottlenecks in CNN backbone & LGE filtering @ encoder input proj & HF gating @ decoder query selection \\
\bottomrule
\end{tabular}
\end{table}

\textbf{YOLOv11/DERNet-S/M}, \textbf{RTMDet-R2-T/S}, and \textbf{PP-PicoDet-S/L} share a consistent three-point pattern: WDG-style blocks replace high-resolution backbone bottlenecks (C2/C3 or C4/C5) to preserve small-object boundaries and reduce aliasing, LGE-style modules are inserted at neck fusion paths to restore directional high-frequency cues, and FDHead-style heads apply P2-level high-frequency gating so that regression gradients concentrate on boundary-supported locations. For \textbf{RT-DETR-R18/R50}, the same spectral roles are realized at functionally equivalent positions: WDG in the CNN backbone before tokens enter the encoder, LGE at the encoder input projection to shape the spectrum mixed by self-attention, and high-frequency gating before decoder query selection to focus queries on boundary-sharp regions, demonstrating that our design seamlessly transfers from CNNs to Transformer-based detectors.

\section{Appendix D: Justification of Frequency-Domain Method Choices}
\label{sec:appendix_d_frequency_choice}

\begin{figure}[H]
  \centering
  \setlength{\tabcolsep}{2pt}
  \renewcommand{\arraystretch}{0}
  \begin{tabular}{@{}c@{}c@{}c@{}c@{}}
      \includegraphics[width=0.21\textwidth]{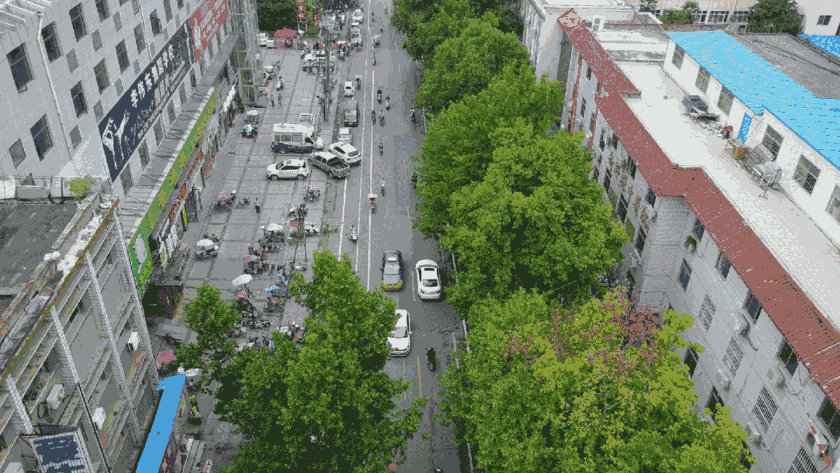} &
      \includegraphics[width=0.24\textwidth]{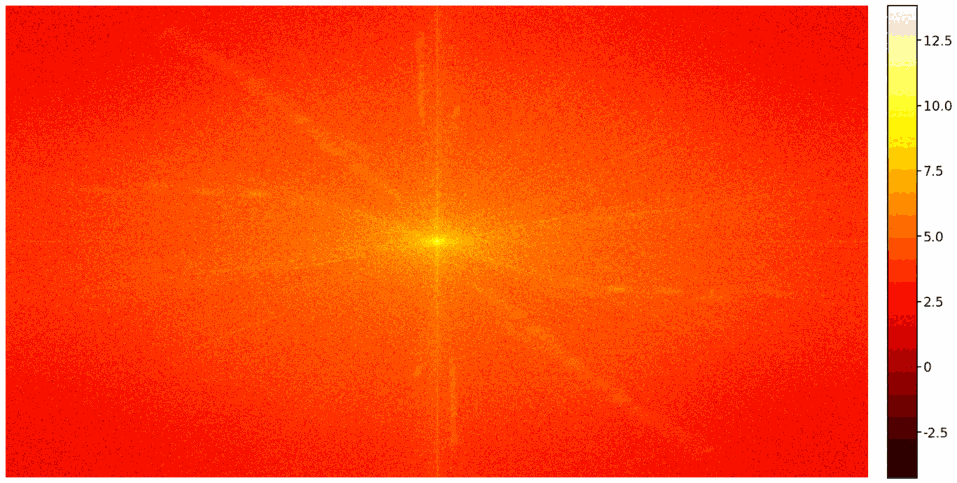} &
      \includegraphics[width=0.24\textwidth]{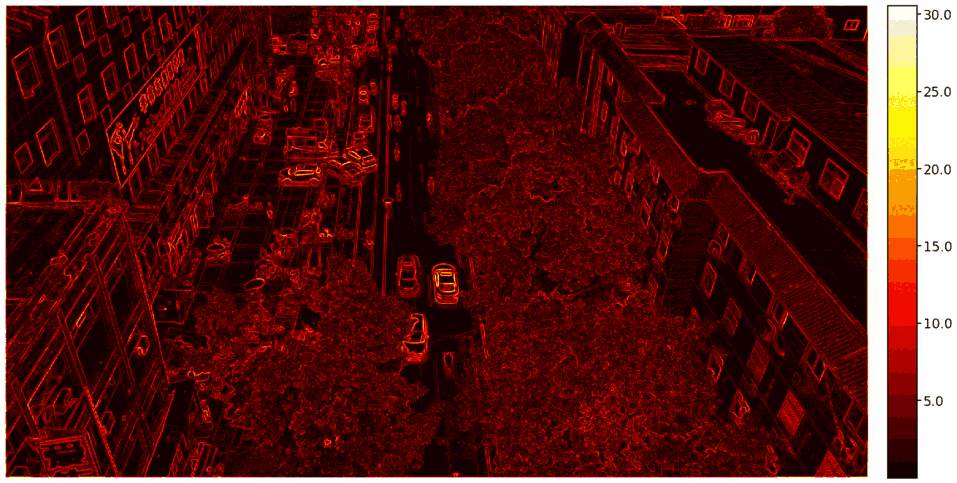} &
      \includegraphics[width=0.24\textwidth]{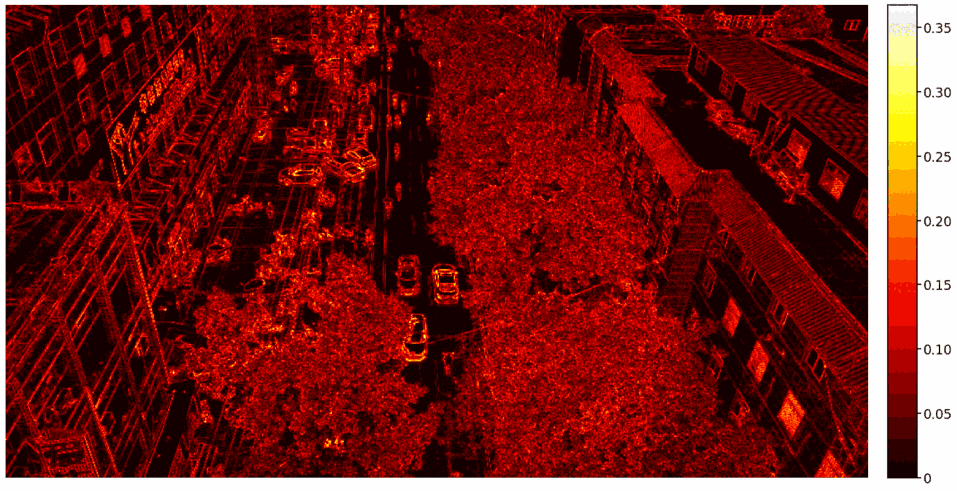} \\[0.7ex]
      \tiny \textbf{(a) Original} & \tiny \textbf{(b) FFT} & \tiny \textbf{(c) Log-Gabor} & \tiny \textbf{(d) Wavelet}
  \end{tabular}
  \caption{\textbf{Qualitative comparison of frequency-domain methods on a VisDrone2019 image.} (a) Original image; (b) FFT magnitude spectrum (global, no spatial localization); (c) Log-Gabor final enhanced response (spatially localized, emphasizes edges and textures); (d) Wavelet high-frequency energy map (spatially localized, highlights structured regions).}
  \label{fig:freq_qualitative}
\end{figure}

\begin{figure}[H]
  \centering
  \setlength{\tabcolsep}{2pt}
  \renewcommand{\arraystretch}{0}
  \begin{tabular}{@{}c@{}c@{}c@{}c@{}}
      \includegraphics[width=0.24\textwidth]{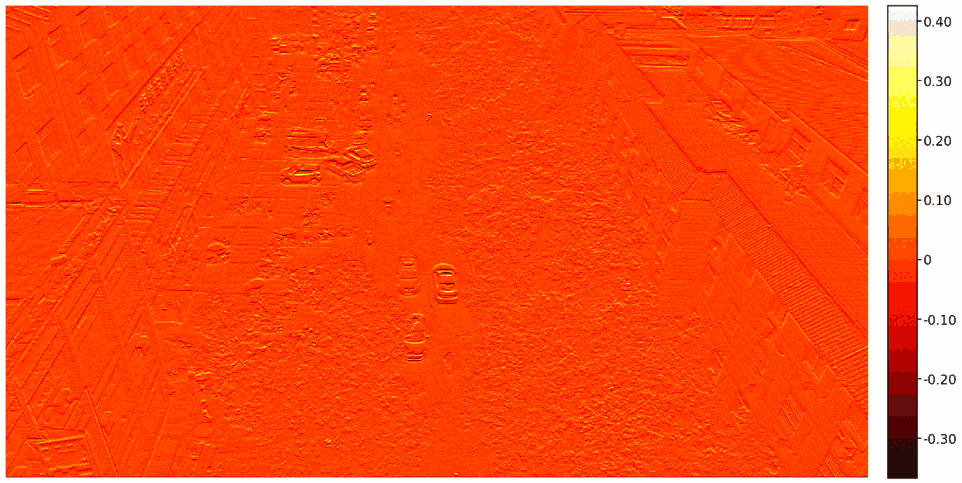} &
      \includegraphics[width=0.24\textwidth]{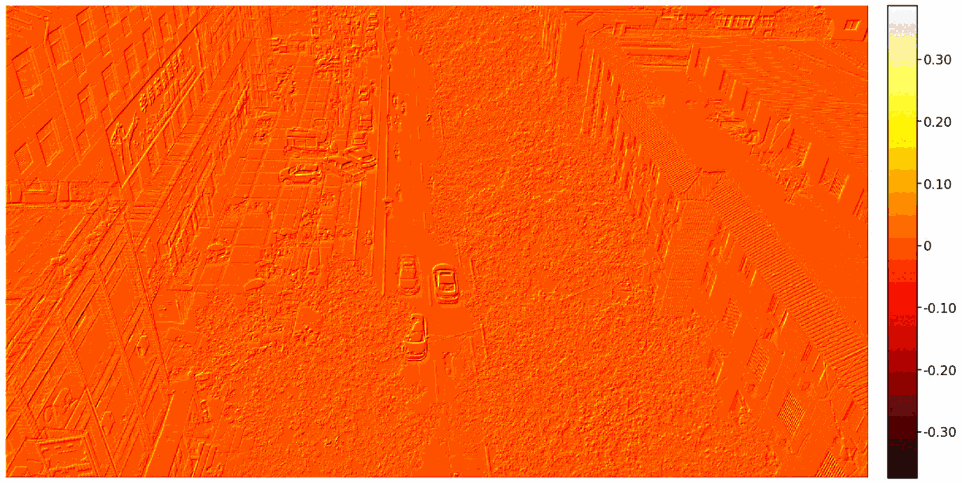} &
      \includegraphics[width=0.24\textwidth]{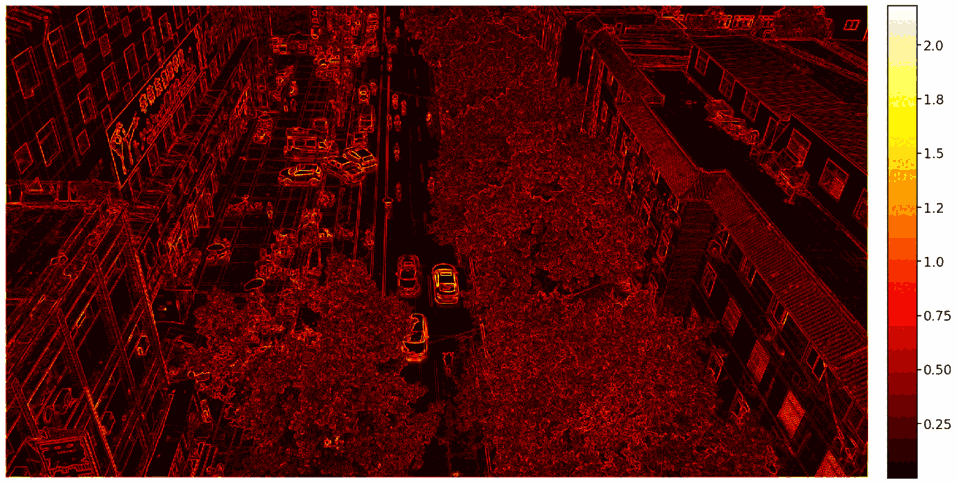} &
      \includegraphics[width=0.24\textwidth]{images/0000001_02999_d_0000005_wavelet_high_freq_energy.png} \\[0.7ex]
      \tiny \textbf{(a) 0° orientation} & \tiny \textbf{(b) 45° orientation} & \tiny \textbf{(c) Log-Gabor aggregated} & \tiny \textbf{(d) Wavelet high-freq}
  \end{tabular}
  \caption{\textbf{Log-Gabor directional selectivity demonstration.} (a-b) Responses at two orientations illustrate directional edge detection; (c) Aggregated Log-Gabor response across orientations; (d) Wavelet high-frequency energy map for comparison.}
  \label{fig:loggabor_orientations}
\end{figure}

We justify using \textbf{wavelet} in the backbone (WDG) and head (FDHead) and \textbf{Log-Gabor} in the neck (LGE/LGE-W) via qualitative visualizations and quantitative analyses on VisDrone2019 images. 

\subsection{Qualitative Visualization Results}

Fig.~\ref{fig:freq_qualitative} shows that FFT operates with globally coupled frequency responses and therefore lacks spatial localization, whereas both wavelet and Log-Gabor maintain spatial correspondence that is crucial for boundary alignment. Within this localized regime, wavelet yields region-level high-frequency energy that highlights structured areas and provides compact, stable edge-aware cues suited to backbone and head stages under tight computational budgets, while Log-Gabor produces smoother band-pass responses that more selectively concentrate on edges and textures (Fig.~\ref{fig:loggabor_orientations}), making it better aligned with the neck’s role in multi-scale feature aggregation and small-object detail enhancement.

\subsection{Quantitative Analysis Results}

Table~\ref{tab:freq_quantitative} shows that wavelet has the lowest computational cost and latency, substantially lower than both FFT and especially Log-Gabor. This efficiency gap makes wavelet preferable in the backbone and head, where frequency-domain operators are invoked repeatedly across many layers, while the higher cost of Log-Gabor restricts its use to a few neck stages rather than pervasive deployment throughout the network.

\begin{table}[h]
\centering
\caption{Quantitative comparison of frequency-domain methods on multiple VisDrone2019 images (1080×1920). Time (ms) is per single forward pass.}
\label{tab:freq_quantitative}
\vspace{0.1cm}
\small
\begin{tabular}{@{}lcc@{}}
\toprule
\cellcolor{blue!5}\textbf{Method} & \cellcolor{blue!5}\textbf{GFLOPs} & \cellcolor{blue!5}\textbf{Time (ms)} \\
\midrule
FFT & $0.218$ & $0.050 \pm 0.020$ \\
Wavelet & $0.015$ & $0.007 \pm 0.002$ \\
Log-Gabor & $0.933$ & $0.940 \pm 0.090$ \\
\bottomrule
\end{tabular}
\end{table}

Table~\ref{tab:freq_properties} summarizes the key properties of each method, justifying our design choices: \textbf{Wavelet} is used in the backbone (WDG) and head (FDHead) because, in these deep and frequently reused stages, it provides fast, spatially localized, region-level high-frequency cues that can be applied repeatedly with low overhead to indicate “where structure is present,” while \textbf{Log-Gabor} is reserved for the neck (LGE/LGE-W) because this shallower, multi-scale fusion part of the network benefits more from its more selective, edge- and texture-focused high-frequency modeling, allowing a few stages to trade higher per-layer cost for sharper boundaries and richer small-object details.

\begin{table}[h]
\centering
\caption{Qualitative comparison of key properties of frequency-domain methods.}
\label{tab:freq_properties}
\vspace{0.1cm}
\small
\begin{tabular}{@{}lccc@{}}
\toprule
\cellcolor{blue!5}\textbf{Property} & \cellcolor{blue!5}\textbf{FFT} & \cellcolor{blue!5}\textbf{Wavelet} & \cellcolor{blue!5}\textbf{Log-Gabor} \\
\midrule
Spatial Localization & \textcolor{red}{No} & \textcolor{green}{Yes} & \textcolor{green}{Yes} \\
Directional Selectivity & \textcolor{red}{None} & \textcolor{orange}{Limited} & \textcolor{green}{Strong} \\
Multi-Scale Response & \textcolor{red}{None} & \textcolor{orange}{Limited} & \textcolor{green}{Strong} \\
Computational Efficiency & \textcolor{orange}{Moderate} & \textcolor{green}{High} & \textcolor{orange}{Moderate} \\
Invertibility & \textcolor{green}{Yes} & \textcolor{green}{Yes} & \textcolor{red}{No} \\
\bottomrule
\end{tabular}
\end{table}

\section{Appendix E: Top-K Frequency Band/Orientation Ablation}
\label{sec:appendix_e_topk}

We further analyze the computational cost of varying the \textbf{number of orientations} (K) and \textbf{number of frequency bands / scales} (S) in the Log-Gabor based neck enhancer. We report the \textbf{module-only} parameter count and GFLOPs when inserting a single module at each target neck layer.
Table~\ref{tab:topk_ablation_cost} summarizes the results.

\begin{table}[h]
\centering
\caption{\textbf{Top-K frequency band/orientation ablation (single-module setting).} K is the number of orientations and S is the number of scales. Each cell reports Params and GFLOPs, with +$\Delta$mAP50 shown in parentheses.}
\label{tab:topk_ablation_cost}
\vspace{0.1cm}
\setlength\tabcolsep{0.1pt}
\renewcommand{\arraystretch}{1.1}
\fontsize{5pt}{6pt}\selectfont
\resizebox{\textwidth}{!}{
\begin{tabular}{@{}l*{9}{c}@{}}
\toprule
\textbf{Target Layer}
& \multicolumn{3}{c}{\vh{\textbf{\fontsize{6pt}{7pt}\selectfont\textnormal{K=1}}}} & \multicolumn{3}{c}{\vh{\textbf{\fontsize{6pt}{7pt}\selectfont\textnormal{K=2}}}} & \multicolumn{3}{c}{\vh{\textbf{\fontsize{6pt}{7pt}\selectfont\textnormal{K=4}}}} \\
\cmidrule(lr){2-4}\cmidrule(lr){5-7}\cmidrule(lr){8-10}
& \cellcolor{blue!5}\vheader{\textbf{\fontsize{6pt}{7pt}\selectfont\textnormal{S=1}}} & \cellcolor{blue!5}\vheader{\textbf{\fontsize{6pt}{7pt}\selectfont\textnormal{S=2}}} & \cellcolor{blue!5}\vheader{\textbf{\fontsize{6pt}{7pt}\selectfont\textnormal{S=3}}}
& \cellcolor{blue!5}\vheader{\textbf{\fontsize{6pt}{7pt}\selectfont\textnormal{S=1}}} & \cellcolor{blue!5}\vheader{\textbf{\fontsize{6pt}{7pt}\selectfont\textnormal{S=2}}} & \cellcolor{blue!5}\vheader{\textbf{\fontsize{6pt}{7pt}\selectfont\textnormal{S=3}}}
& \cellcolor{blue!5}\vheader{\textbf{\fontsize{6pt}{7pt}\selectfont\textnormal{S=1}}} & \cellcolor{blue!5}\vheader{\textbf{\fontsize{6pt}{7pt}\selectfont\textnormal{S=2}}} & \cellcolor{blue!5}\vheader{\textbf{\fontsize{6pt}{7pt}\selectfont\textnormal{S=3}}} \\
\midrule
\raisebox{0.8ex}{\textbf{Layer12 (P5$\rightarrow$P4)}}
& \costcellm{3,843}{0.0135}{+0.001} & \costcellm{5,572}{0.0191}{+0.002} & \costcellm{7,301}{0.0246}{+0.002}
& \cellcolor{impStep1}\costcellm{5,572}{0.0191}{+0.004} & \costcellm{9,029}{0.0301}{+0.005} & \costcellm{12,486}{0.0412}{+0.005}
& \costcellm{9,030}{0.0301}{+0.005} & \costcellm{15,943}{0.0522}{+0.005} & \costcellm{22,856}{0.0743}{+0.007} \\
\specialrule{0.45pt}{0pt}{0pt}
\specialrule{0.15pt}{0.20ex}{0.6ex}

\raisebox{0.8ex}{\textbf{Layer16 (P4$\rightarrow$P3)}}
& \costcellm{3,203}{0.0451}{+0.002} & \costcellm{4,644}{0.0635}{+0.002} & \costcellm{6,085}{0.0819}{+0.003}
& \cellcolor{impStep1}\costcellm{4,644}{0.0635}{+0.006} & \costcellm{7,525}{0.1004}{+0.006} & \costcellm{10,406}{0.1372}{+0.007}
& \costcellm{7,526}{0.1004}{+0.007} & \costcellm{13,287}{0.1741}{+0.008} & \costcellm{19,048}{0.2478}{+0.008} \\
\specialrule{0.45pt}{0pt}{0pt}
\specialrule{0.15pt}{0.20ex}{0.6ex}

\raisebox{0.8ex}{\textbf{Layer20 (P3$\rightarrow$P2)}}
& \costcellm{8,835}{0.1327}{+0.002} & \costcellm{9,700}{0.1770}{+0.003} & \costcellm{10,565}{0.2212}{+0.005}
& \cellcolor{impStep1}\costcellm{9,700}{0.1770}{+0.006} & \costcellm{11,429}{0.2654}{+0.009} & \costcellm{13,158}{0.3539}{+0.008}
& \costcellm{11,430}{0.2654}{+0.007} & \costcellm{14,887}{0.4424}{+0.009} & \costcellm{18,344}{0.6193}{+0.009} \\
\bottomrule
\end{tabular}}
\end{table}

Overall, increasing either K or S consistently increases compute and parameters, with the steepest GFLOPs growth observed at the highest-resolution neck layer (P3$\rightarrow$P2). In our main configuration, we adopt small K and S (e.g., S=1 with limited orientations) to balance efficiency and detail enhancement.
\begin{figure*}[t]
  \centering
  \setlength{\tabcolsep}{0pt}
  \renewcommand{\arraystretch}{0}
  \begin{tabular}{@{}c@{}c@{}c@{}c@{}c@{}}
      \includegraphics[width=0.19\textwidth]{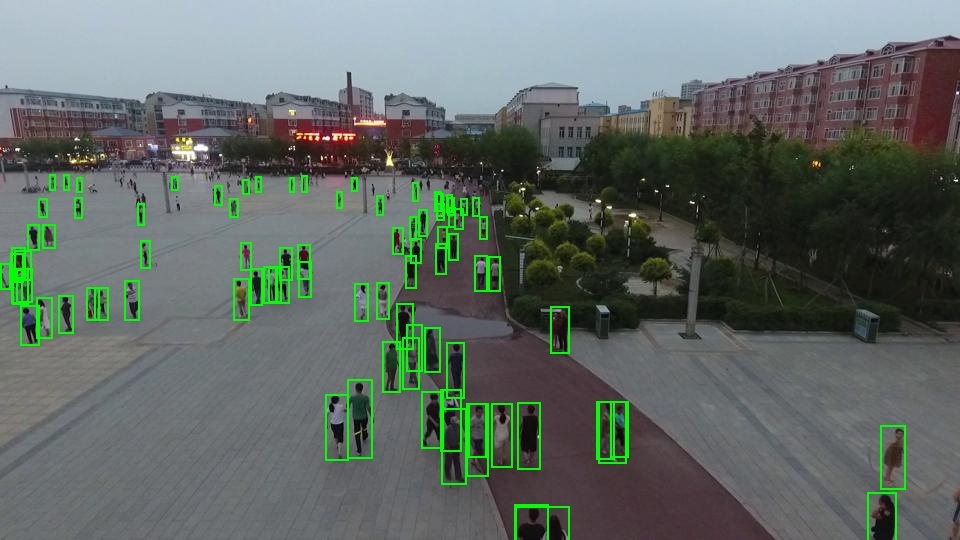} &
      \includegraphics[width=0.19\textwidth]{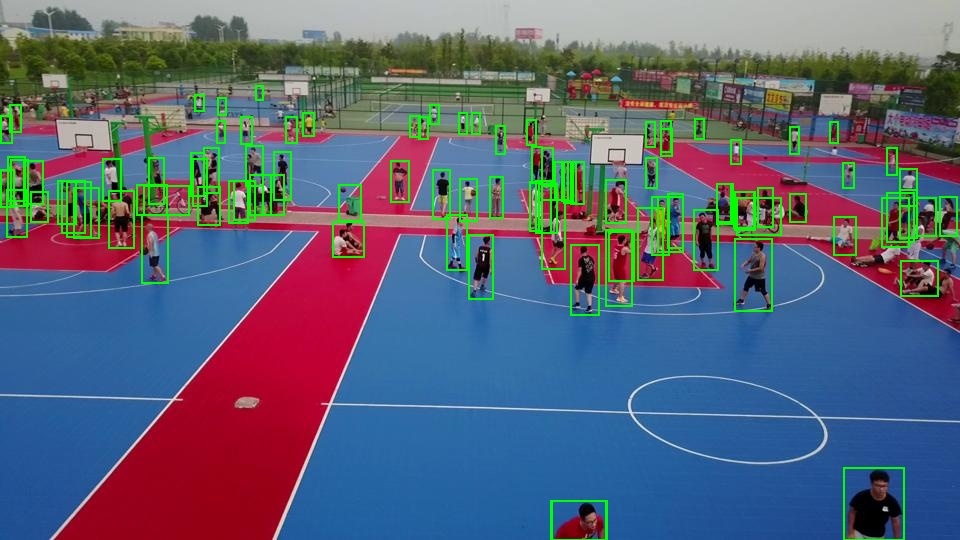} &
      \includegraphics[width=0.19\textwidth]{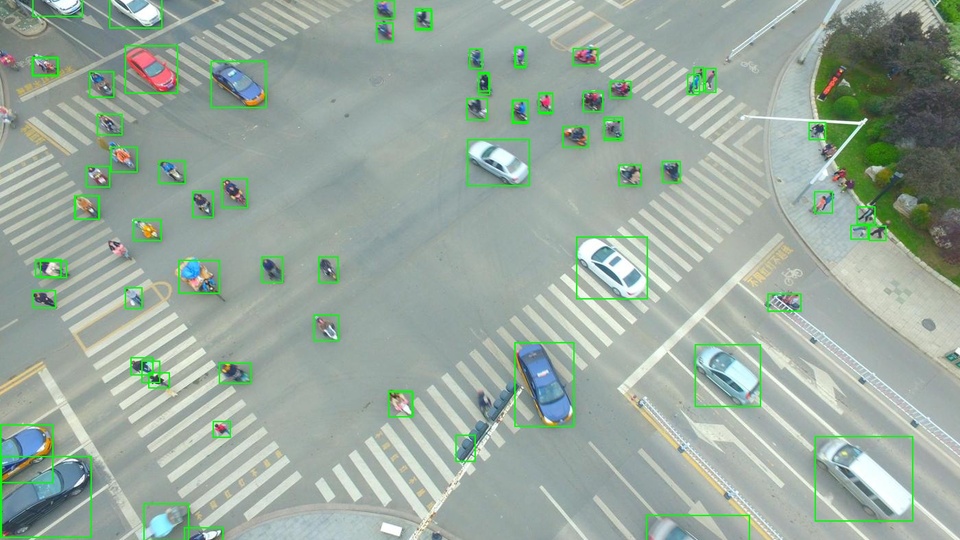} &
      \includegraphics[width=0.19\textwidth]{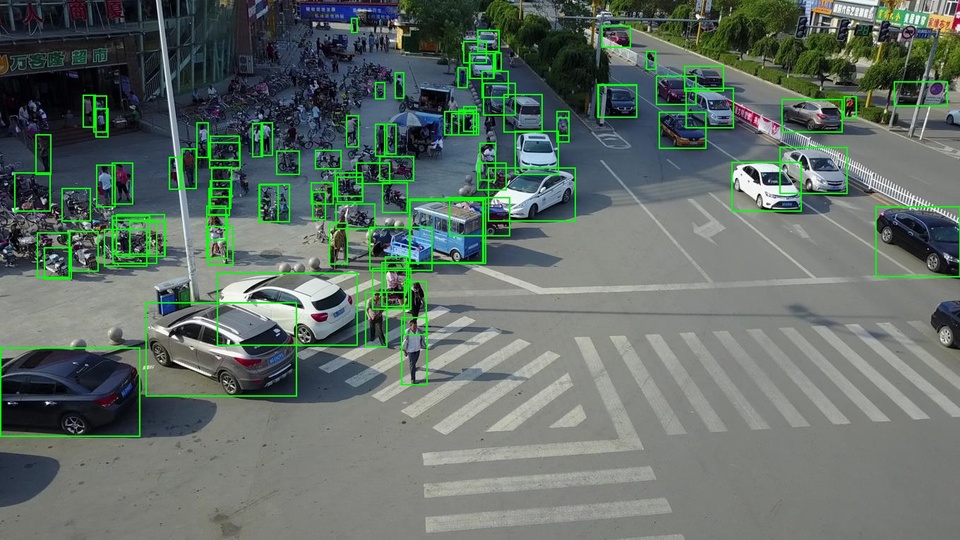} &
      \includegraphics[width=0.19\textwidth]{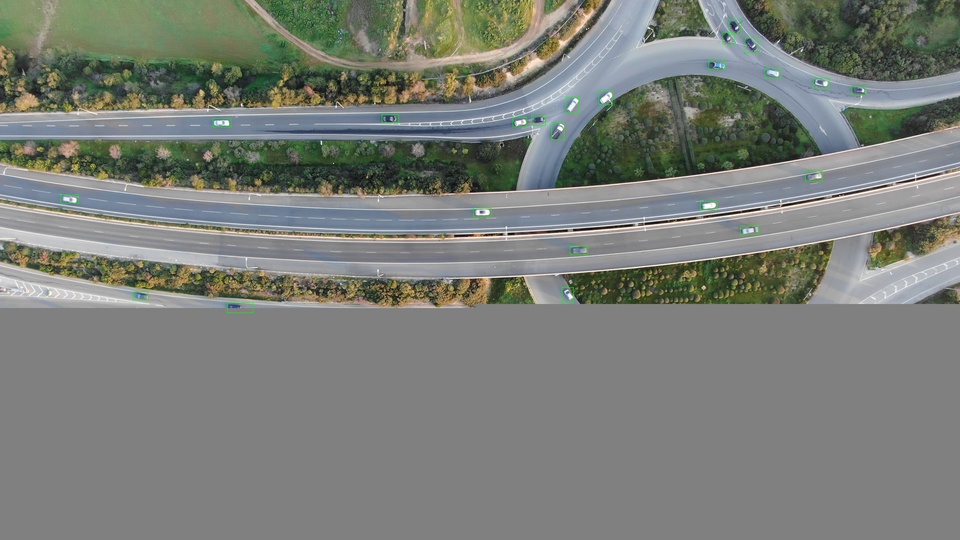} \\
      \includegraphics[width=0.19\textwidth]{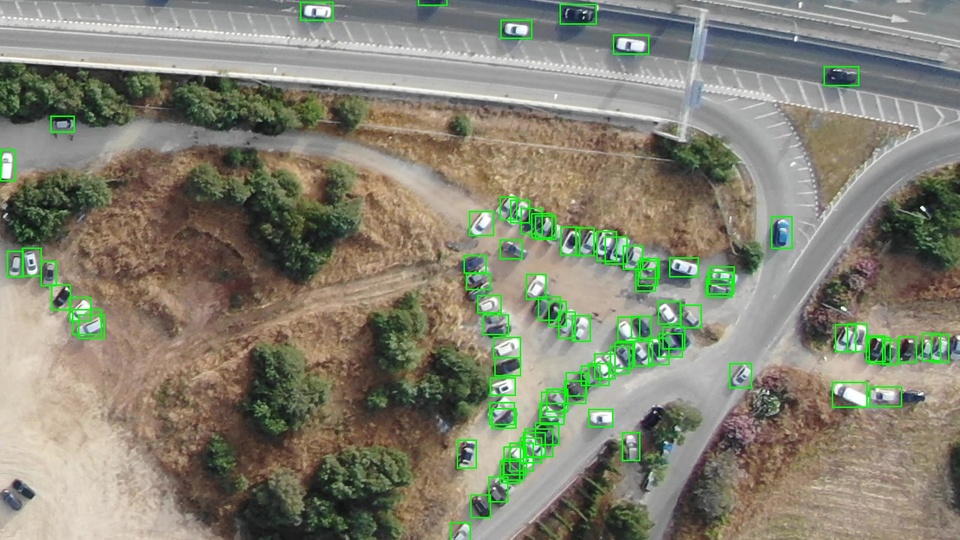} &
      \includegraphics[width=0.19\textwidth]{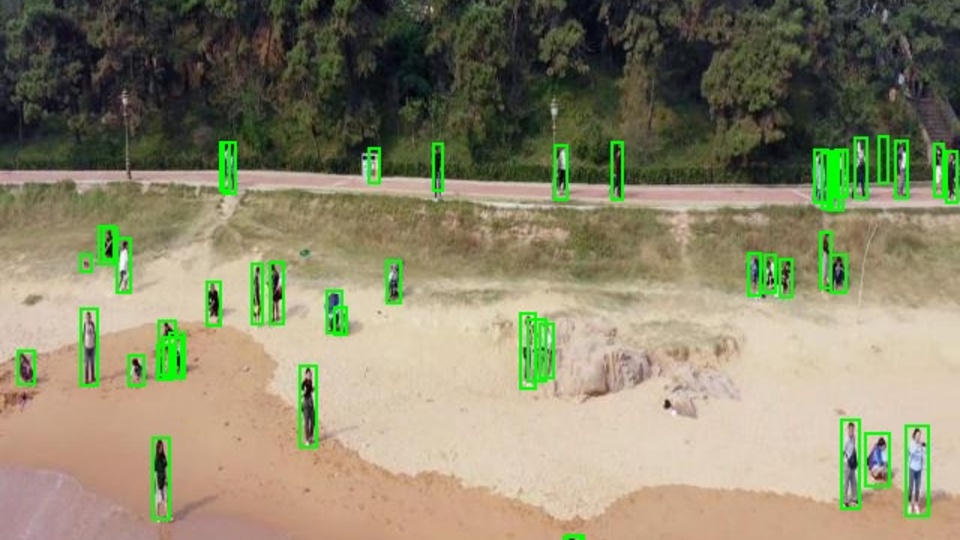} &
      \includegraphics[width=0.19\textwidth]{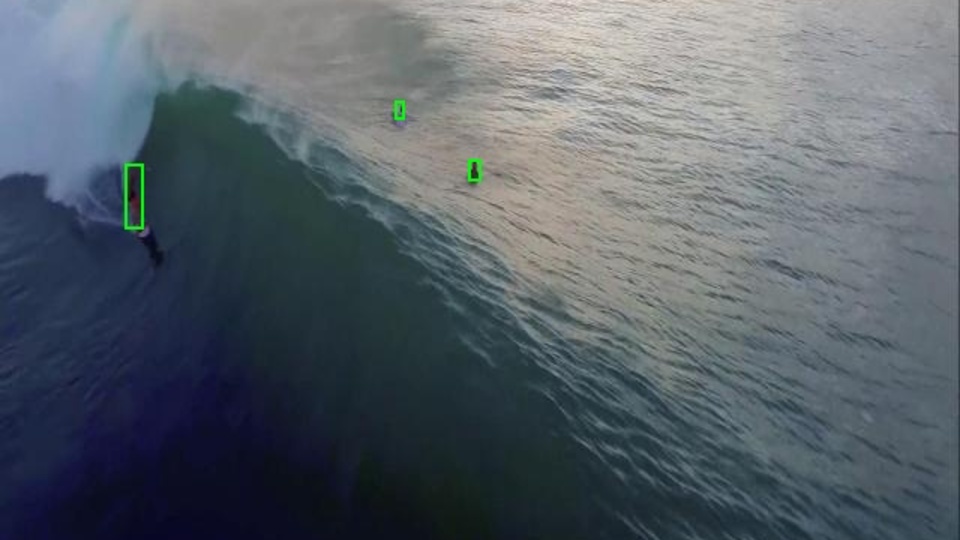} &
      \includegraphics[width=0.19\textwidth]{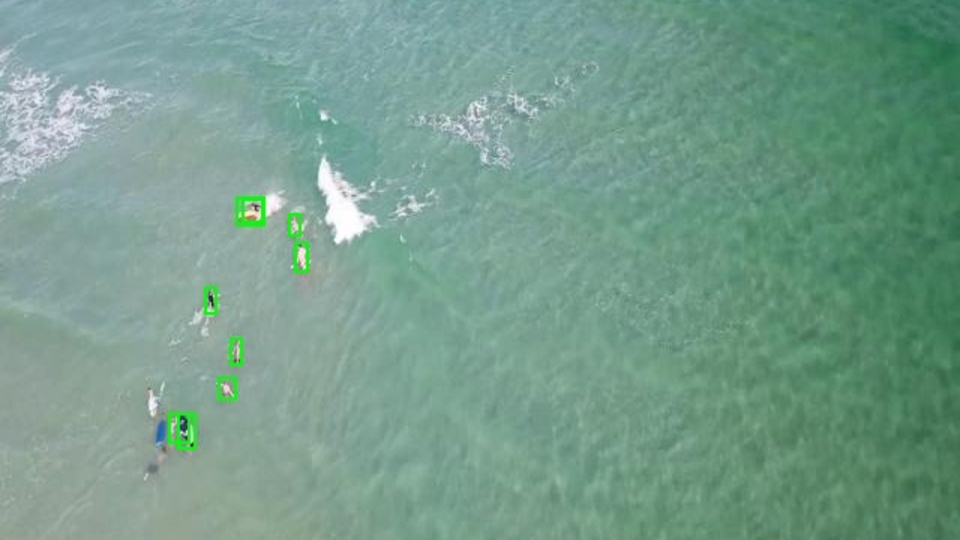} &
      \includegraphics[width=0.19\textwidth]{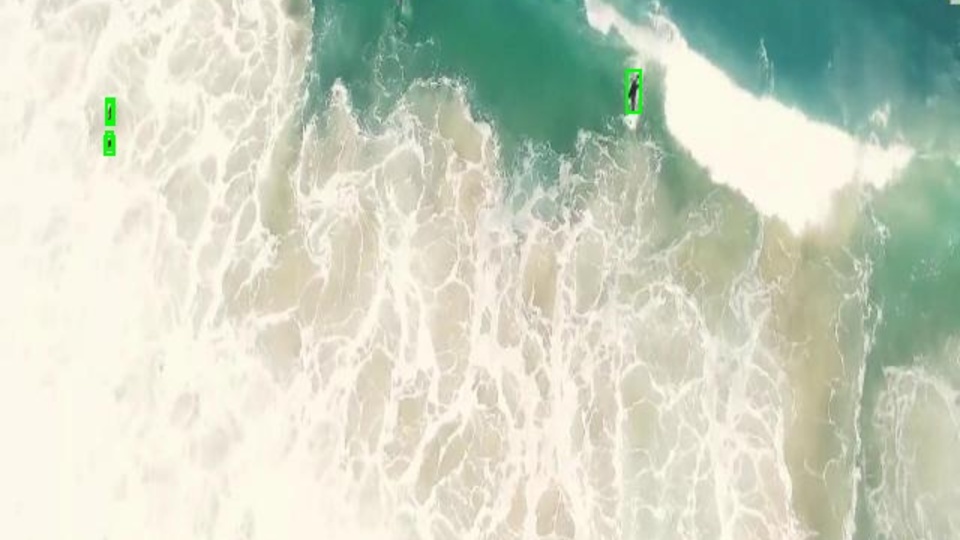} \\[0.7ex]
  \end{tabular}

  \caption{Successful detection cases where DERNet-S accurately detects and localizes small and distant objects.}
  \label{fig:successful_cases}
\end{figure*}

\begin{figure*}[t]
  \centering
  \setlength{\tabcolsep}{0pt}
  \renewcommand{\arraystretch}{0}
  \begin{tabular}{@{}c@{}c@{}c@{}c@{}c@{}}
      \includegraphics[width=0.19\textwidth]{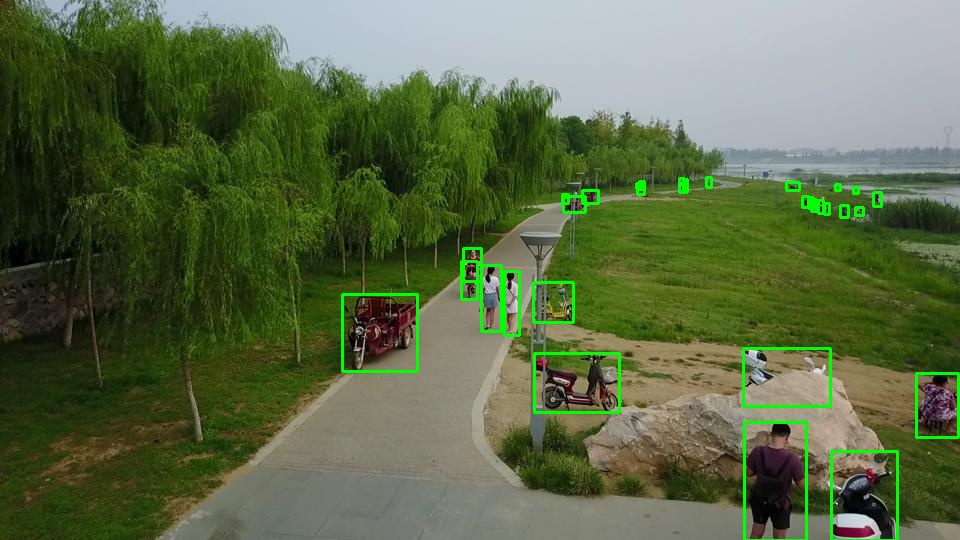} &
      \includegraphics[width=0.19\textwidth]{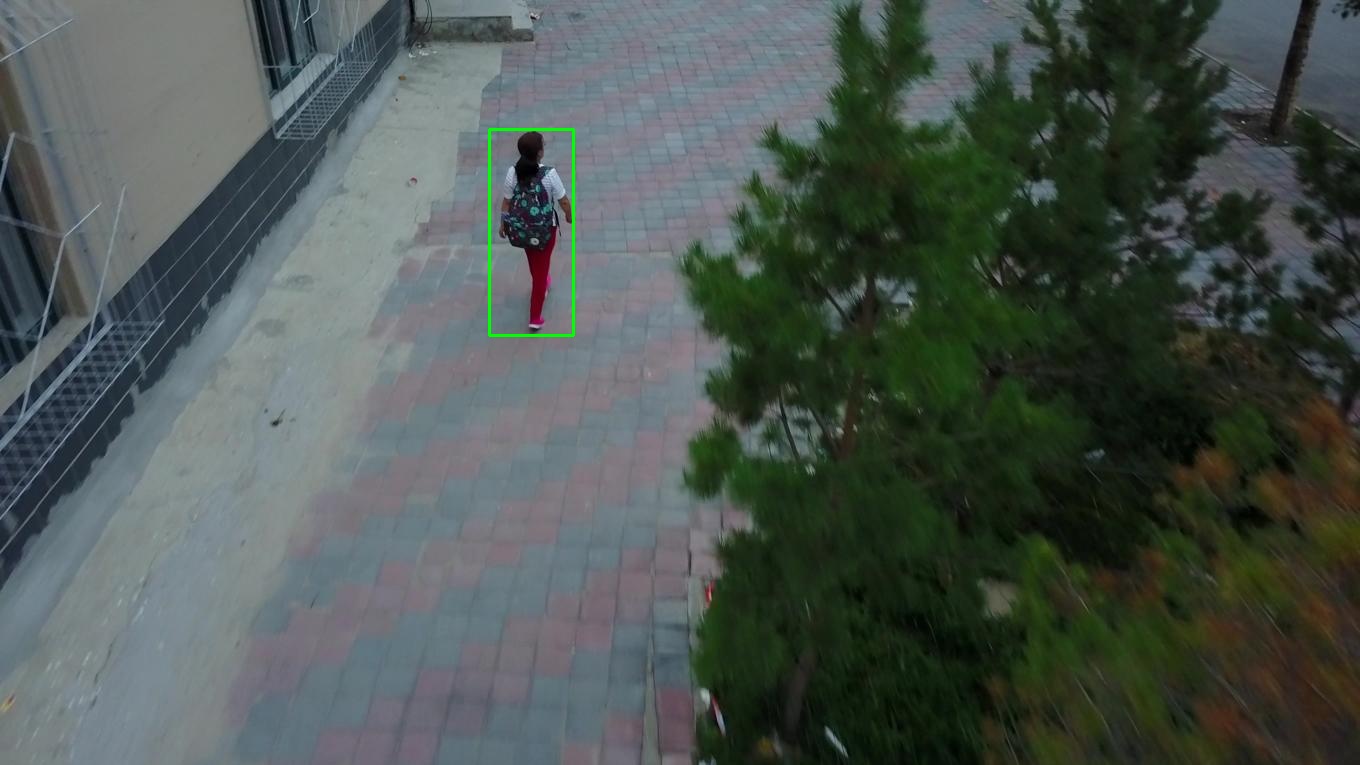} &
      \includegraphics[width=0.19\textwidth]{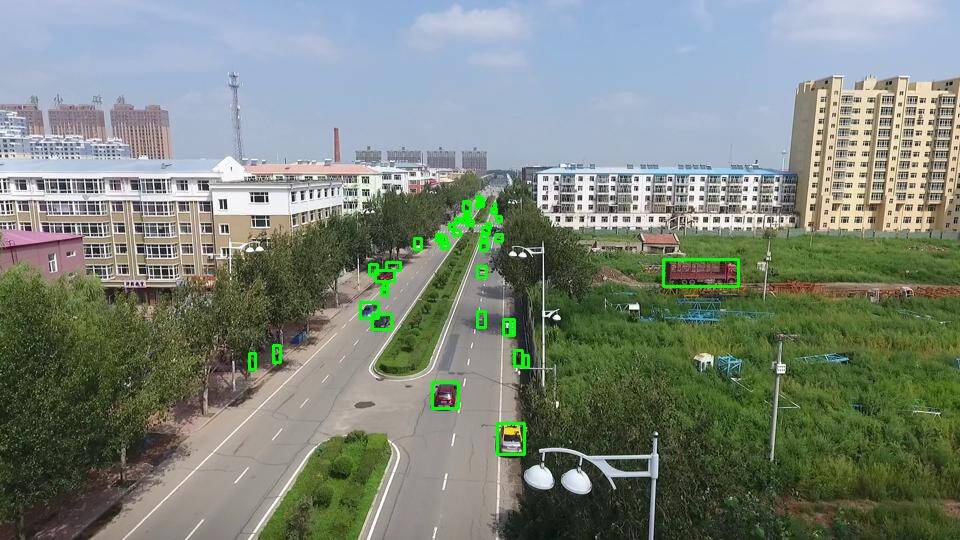} &
      \includegraphics[width=0.19\textwidth]{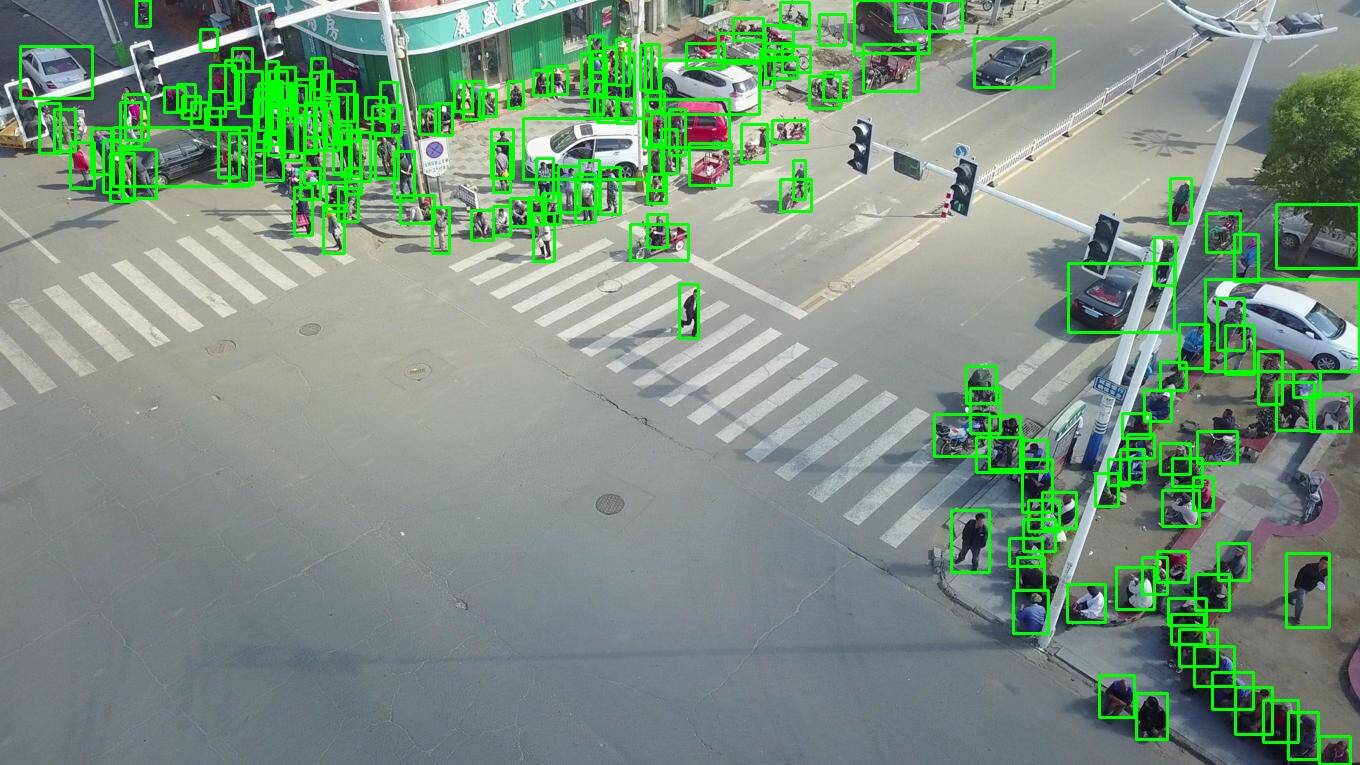} &
      \includegraphics[width=0.19\textwidth]{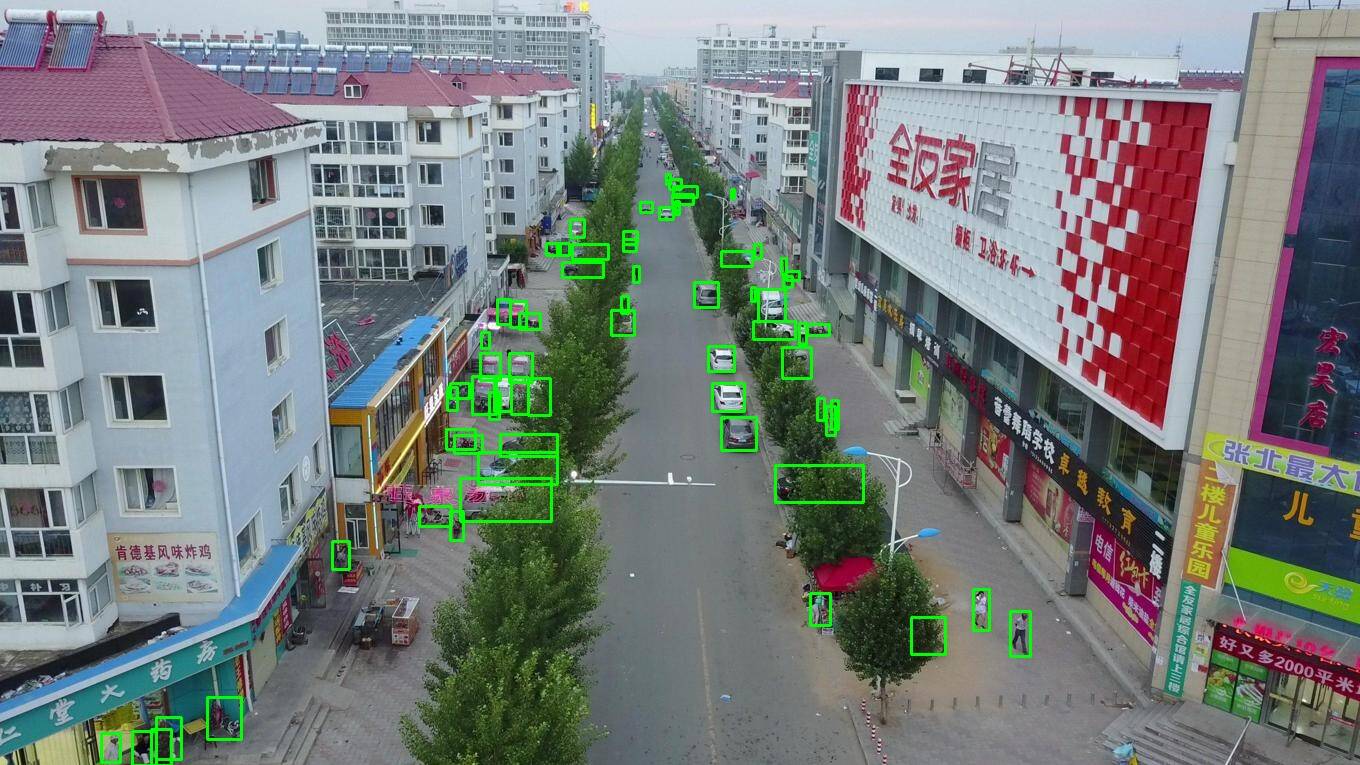} \\
      \includegraphics[width=0.19\textwidth]{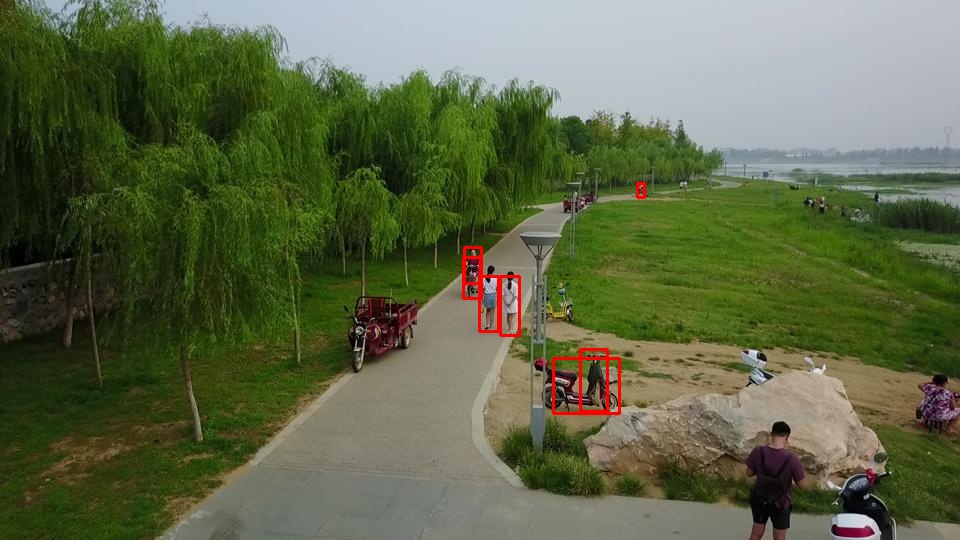} &
      \includegraphics[width=0.19\textwidth]{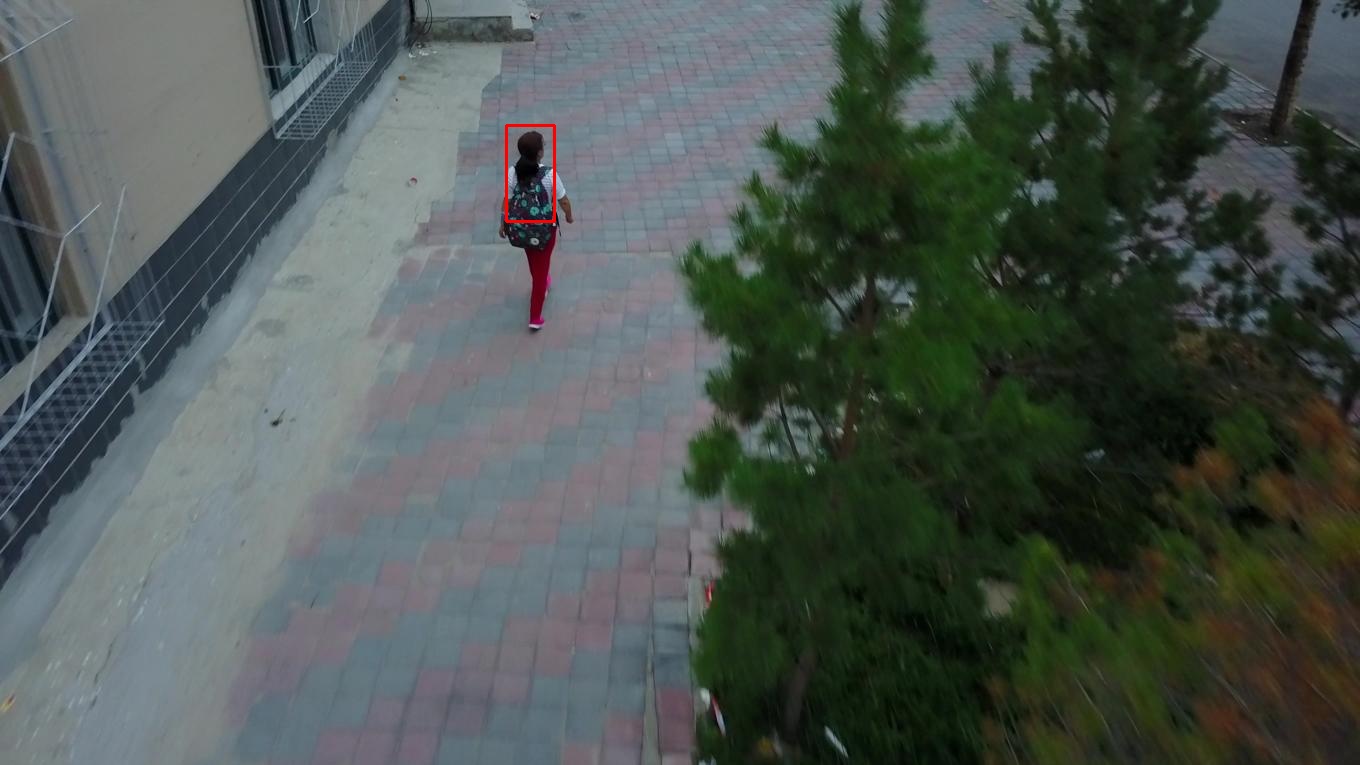} &
      \includegraphics[width=0.19\textwidth]{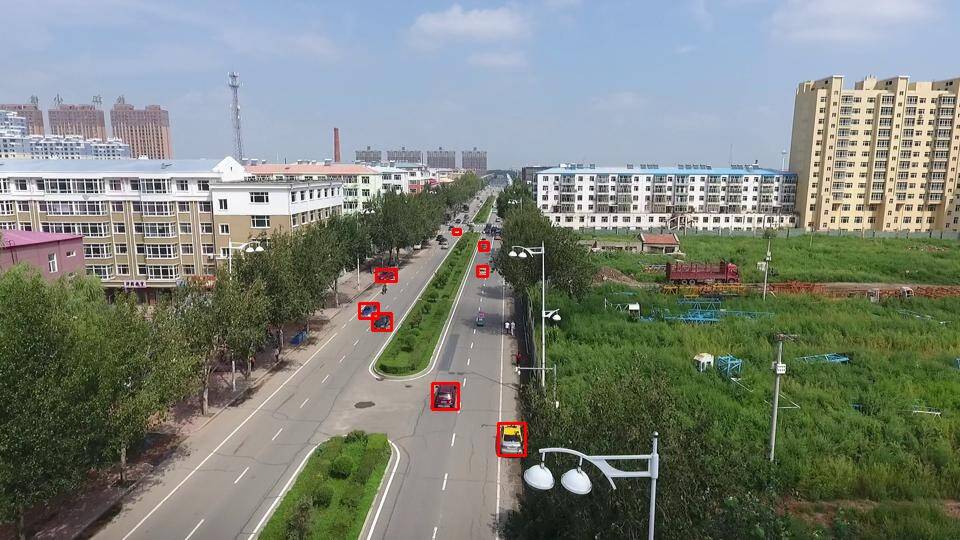} &
      \includegraphics[width=0.19\textwidth]{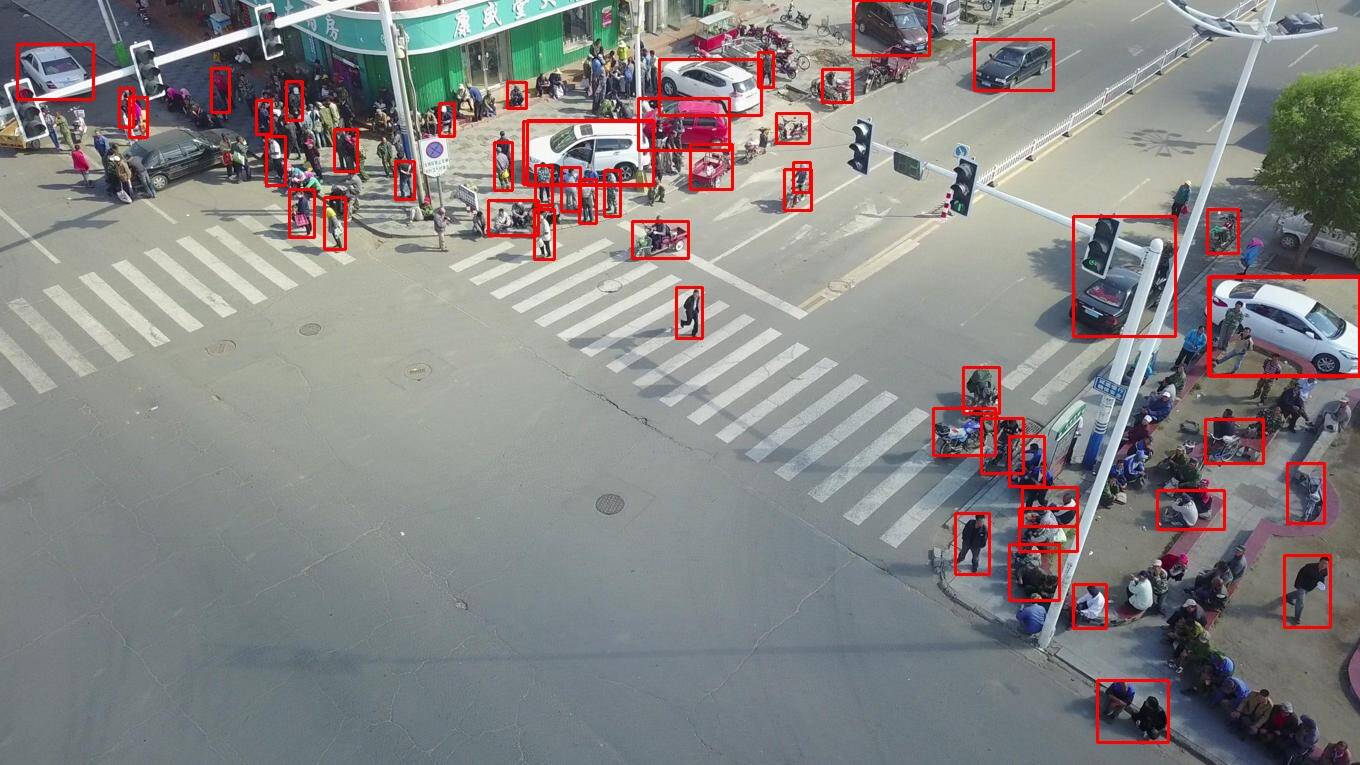} &
      \includegraphics[width=0.19\textwidth]{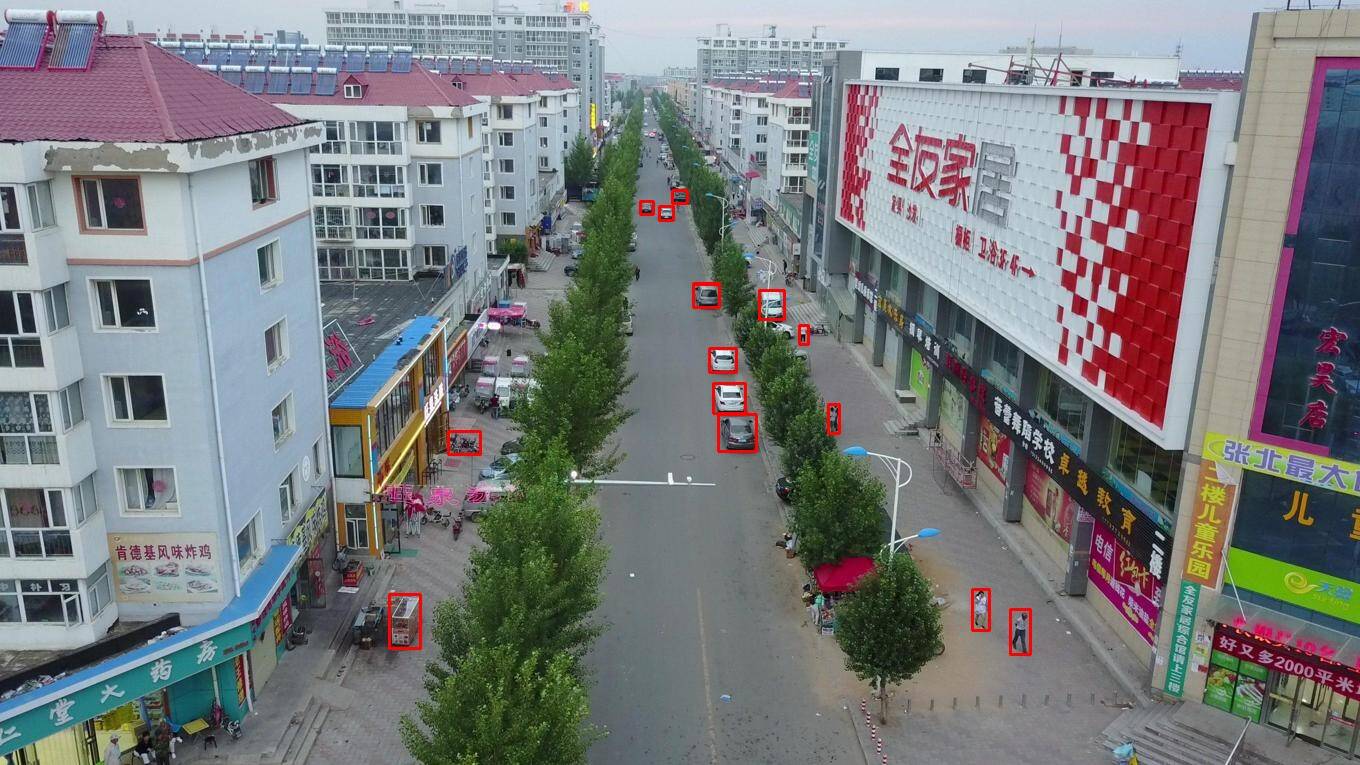} \\[0.7ex]
      \tiny \textbf{(a) Distant objects} & \tiny \textbf{(b) Incorrect segmentation} & \tiny \textbf{(c) Extremely small objects} & \tiny \textbf{(d) Tightly-clustered objects} & \tiny \textbf{(e) Occluded objects}
  \end{tabular}

  \caption{Qualitative comparison between ground truth and DERNet-S predictions on challenging small object detection scenarios. The first row shows ground truth annotations, while the second row shows DERNet-S model predictions.}
  \label{fig:qualitative_analysis}
\end{figure*}

\section{Appendix F: Successful Detection Cases}
\label{sec:appendix_f_successful_cases}

Figure~\ref{fig:successful_cases} showcases a series of inherently challenging small-object detection cases: some images contain distant and low-resolution targets that occupy only a few pixels; others exhibit strong background clutter or low contrast between objects and their surroundings; yet others involve partially occluded or densely packed instances that are hard to distinguish. Despite these difficulties, DERNet-S consistently produces accurate detections with tight bounding boxes around the objects of interest. This demonstrates that our frequency-guided representation learner is able to robustly capture and localize small and distant targets even under adverse imaging conditions.

\section{Appendix G: Error Analysis of Small Object Detection}
\label{sec:appendix_g_error_analysis}

Fig.~\ref{fig:qualitative_analysis} reveals several challenging scenarios in small object detection: (a) shows distant objects that are difficult to detect due to their small size and distance; (b) demonstrates incorrect segmentation where objects are improperly divided due to color variations; (c) illustrates extremely small objects that challenge detection algorithms; (d) depicts tightly-clustered objects where individual elements are hard to distinguish; (e) shows occluded objects that are partially hidden by other elements, making complete detection difficult. These persistent errors highlight that while our model advances in frequency-guided representation, it still faces challenges in robustly handling extreme scale variations and complex occlusion. Further efforts are needed to enhance its precision in these intricate scenarios.

\section{Appendix H: Training Configurations for Other Architectures}
\label{sec:appendix_h_training_configs}

Table~\ref{tab:config} summarizes the configuration of experiment environments. 
Table~\ref{tab:jetson_config} summarizes the edge hardware used for Jetson Nano FPS evaluation.
And Table~\ref{tab:training_configs} summarizes the training hyperparameters used for PP-PicoDet, RTMDet-R2, and RT-DETR models, ensuring a fair and comparable experimental setup across all models.

\begin{table}[H]
  \centering
  \caption{Configuration of Experiment Environments.}
  \label{tab:config}
  \setlength{\tabcolsep}{6pt}
  \renewcommand{\arraystretch}{1.05}
  \fontsize{9pt}{10.5pt}\selectfont
  \begin{tabular}{@{}l p{0.70\textwidth}@{}}
      \toprule
      \rowcolor{blue!5}\textbf{Environment} & \textbf{Parameter} \\
      \midrule
      CPU              & Intel(R) Xeon(R) Gold 5218R CPU @ 2.10GHz \\
      GPU              & NVIDIA A100-PCIE-40GB \\
      VRAM             & 40 GB \\
      RAM              & 46 GB \\
      Operating System & Rocky Linux 8.5 (Green Obsidian) \\
      Language         & Python 3.10.14 \\
      Frame            & PyTorch 2.1.0 \\
      CUDA Version     & 12.6 \\
      \bottomrule
  \end{tabular}
  \vspace{-1mm}
\end{table}

\begin{table}[H]
  \centering
  \caption{Configuration of Edge Inference Environment.}
  \label{tab:jetson_config}
  \setlength{\tabcolsep}{6pt}
  \renewcommand{\arraystretch}{1.05}
  \fontsize{9pt}{10.5pt}\selectfont
  \begin{tabular}{@{}l p{0.70\textwidth}@{}}
      \toprule
      \rowcolor{blue!5}\textbf{Environment} & \textbf{Parameter} \\
      \midrule
      Device           & NVIDIA Jetson Nano \\
      CPU              & Quad-core ARM Cortex-A57 MPCore processor \\
      GPU              & NVIDIA Maxwell architecture with 128 NVIDIA CUDA cores \\
      Memory           & 4 GB 64-bit LPDDR4, 1600 MHz \\
      Memory Bandwidth & 25.6 GB/s \\
      Storage          & microSD card \\
      Software Stack   & NVIDIA JetPack SDK \\
      Frame            & PyTorch \\
      \bottomrule
  \end{tabular}
\end{table}

\begin{table}[H]
  \caption{Training configurations for YOLOv11, PP-PicoDet, RTMDet-R2, and RT-DETR architectures.}
  \label{tab:training_configs}
  \centering
  \setlength{\tabcolsep}{2pt}
  \renewcommand{\arraystretch}{1.0}
  \fontsize{8pt}{9pt}\selectfont
  \begin{tabular}{@{}>{\raggedright\arraybackslash}p{0.19\textwidth}
  >{\centering\arraybackslash}p{0.055\textwidth}
  >{\centering\arraybackslash}p{0.085\textwidth}
  >{\centering\arraybackslash}p{0.085\textwidth}
  >{\centering\arraybackslash}p{0.120\textwidth}
  >{\centering\arraybackslash}p{0.170\textwidth}
  >{\centering\arraybackslash}p{0.225\textwidth}
  >{\centering\arraybackslash}p{0.060\textwidth}@{}}
    \toprule
    \cellcolor{blue!5}\textbf{Architecture}
    & \cellcolor{blue!5}\scriptsize\textbf{Epochs}
    & \cellcolor{blue!5}\scriptsize\textbf{Input Size}
    & \cellcolor{blue!5}\scriptsize\textbf{Batch Size}
    & \cellcolor{blue!5}\scriptsize\textbf{Optimizer}
    & \cellcolor{blue!5}\scriptsize\textbf{LR Schedule}
    & \cellcolor{blue!5}\scriptsize\textbf{Augmentation}
    & \cellcolor{blue!5}\scriptsize\textbf{Workers} \\
    \midrule

    \textbf{YOLOv11-S}
    & 300 & 640 & 16 & SGD & Cosine
    & Mosaic
    & 4 \\
    \textbf{YOLOv11-M}
    & 300 & 640 & 16 & SGD & Cosine
    & Mosaic
    & 4 \\

    \textbf{PP-PicoDet-S}
    & 300 & 416 & 12 & Momentum & Cosine
    & Crop+Flip+Color
    & 8 \\
    \textbf{PP-PicoDet-L}
    & 300 & 640 & 12 & Momentum & Cosine
    & Crop+Flip+Color
    & 8 \\

    \specialrule{0.45pt}{0pt}{0pt}
    \specialrule{0.15pt}{0.20ex}{0.4ex}

    \textbf{RTMDet-R2-T}
    & 200 & 1024 & 8 & AdamW & Cosine
    & Flip+Rotation
    & 8 \\
    \textbf{RTMDet-R2-S}
    & 200 & 1024 & 8 & AdamW & Cosine
    & Flip+Rotation
    & 8 \\

    \specialrule{0.45pt}{0pt}{0pt}
    \specialrule{0.15pt}{0.20ex}{0.4ex}

    \textbf{RT-DETR-R18}
    & 72 & 640 & 16 & AdamW & Warmup+Constant
    & Photo+Zoom+Crop
    & 4 \\
    \textbf{RT-DETR-R50}
    & 72 & 640 & 16 & AdamW & Warmup+Constant
    & Photo+Zoom+Crop
    & 4 \\

    \bottomrule
  \end{tabular}
  \vspace{-1mm}
\end{table}


\section{Appendix I: Detailed Formulations for WDG}
\label{sec:appendix_wdg_details}

This appendix provides the complete mathematical formulations for the Wavelet-Difference Gate (WDG) module, including the Haar wavelet transform and the Re-parameterized Central-Difference Convolution (RepCDC).

\subsection{Haar Discrete Wavelet Transform}

The 2D Haar DWT decomposes each spatial $2{\times}2$ block into four subbands using the Haar matrix $\mathbf{H}_2 = \begin{pmatrix} 1 & 1\\ 1 & -1 \end{pmatrix}$. For each channel $c$ and spatial location $(u,v)$, the forward and inverse transforms are:
\vspace{-0.2cm}
\begin{equation}
\label{eq:haar_dwt_idwt_appendix}
\begin{aligned}
\mathbf{S}^{(c)}_{u,v} &= \tfrac{1}{2}\,\mathbf{H}_2\,\mathbf{X}^{(c)}_{u,v}\,\mathbf{H}_2^{\top} \quad \text{(DWT)},\\
\mathbf{X}^{(c)}_{u,v} &= \tfrac{1}{2}\,\mathbf{H}_2^{\top}\,\mathbf{S}^{(c)}_{u,v}\,\mathbf{H}_2 \quad \text{(IDWT)},
\end{aligned}
\vspace{-0.2cm}
\end{equation}
where $\mathbf{X}^{(c)}_{u,v} \in \mathbb{R}^{2\times 2}$ is the local spatial block and $\mathbf{S}^{(c)}_{u,v} = \begin{pmatrix}\mathbf{x}^{(c)}_{LL,u,v} & \mathbf{x}^{(c)}_{LH,u,v}\\ \mathbf{x}^{(c)}_{HL,u,v} & \mathbf{x}^{(c)}_{HH,u,v}\end{pmatrix}$ collects the four subbands. Expanding the matrix multiplication yields the element-wise expressions:
\vspace{-0.2cm}
\begin{equation}
\begin{aligned}
\mathbf{x}_{LL} &= \tfrac{1}{2}(x_{00} + x_{01} + x_{10} + x_{11}), \\
\mathbf{x}_{LH} &= \tfrac{1}{2}(x_{00} - x_{01} + x_{10} - x_{11}), \\
\mathbf{x}_{HL} &= \tfrac{1}{2}(x_{00} + x_{01} - x_{10} - x_{11}), \\
\mathbf{x}_{HH} &= \tfrac{1}{2}(x_{00} - x_{01} - x_{10} + x_{11}),
\end{aligned}
\vspace{-0.2cm}
\end{equation}
where $x_{ij}$ denotes the pixel at row $i$, column $j$ within the $2{\times}2$ block. The LL subband is a local average (low-pass), while LH, HL, and HH capture horizontal, vertical, and diagonal gradients (high-pass) respectively.

\subsection{Re-parameterized Central-Difference Convolution (RepCDC)}

RepCDC enhances edge sensitivity by modifying the center coefficient of a standard $3{\times}3$ convolution kernel. Let $\mathbf{W} \in \mathbb{R}^{C_{\mathrm{out}} \times C_{\mathrm{in}} \times 3 \times 3}$ be the base kernel and $\boldsymbol{\theta} \in \mathbb{R}^{C_{\mathrm{out}} \times C_{\mathrm{in}}}$ be a learnable center-difference parameter. The output at spatial location $(p,q)$ for output channel $o$ is:
\vspace{-0.2cm}
\begin{equation}
\label{eq:wdg_repcdc_appendix}
\mathbf{y}^{(o)}_{p,q} = \sum_{c=1}^{C_{\mathrm{in}}}\sum_{i=-1}^{1}\sum_{j=-1}^{1} \mathbf{W}^{(o,c)}_{i,j}\,\mathbf{z}^{(c)}_{p+i,q+j} \;-\; \sum_{c=1}^{C_{\mathrm{in}}} \boldsymbol{\theta}^{(o,c)}\,\mathbf{z}^{(c)}_{p,q},
\vspace{-0.2cm}
\end{equation}
where $\mathbf{z}$ is the input feature map. This is equivalent to constructing an effective kernel $\widetilde{\mathbf{W}}$ where:
\vspace{-0.2cm}
\begin{equation}
\widetilde{\mathbf{W}}^{(o,c)}_{i,j} = 
\begin{cases}
\mathbf{W}^{(o,c)}_{0,0} - \boldsymbol{\theta}^{(o,c)}, & \text{if } i=j=0, \\
\mathbf{W}^{(o,c)}_{i,j}, & \text{otherwise}.
\end{cases}
\vspace{-0.2cm}
\end{equation}
By subtracting the center value, RepCDC emphasizes the difference between the center pixel and its neighbors, effectively embedding a gradient operator into the convolution. During deployment, $\widetilde{\mathbf{W}}$ is precomputed and used as a standard convolution kernel, incurring no additional inference cost.

\subsection{WDG Pseudo-code}

 \begin{algorithm}[H]
 \caption{WDG forward pass (Wavelet-Difference Gate).}
 \label{alg:wdg}
 \scriptsize
\begin{algorithmic}
   \STATE \textbf{Input:} $\mathbf{x}\in\mathbb{R}^{C\times H\times W}$, expansion ratio $e$
   \STATE \textbf{Output:} $\mathbf{y}\in\mathbb{R}^{C\times H\times W}$
   \STATE $\mathbf{x}' \leftarrow f_{1\times1}(\mathbf{x})$ \COMMENT{project to $C'=\lfloor eC\rfloor$}
   \STATE Align $H,W$ to even size by cropping/padding if needed
   \STATE $(\mathbf{x}_{LL},\mathbf{x}_{LH},\mathbf{x}_{HL},\mathbf{x}_{HH}) \leftarrow \mathrm{DWT}(\mathbf{x}')$
   \STATE $\mathbf{y}_{LL} \leftarrow \mathrm{RepCDC}(\mathbf{x}_{LL})$; apply norm+activation
   \STATE $\mathbf{g} \leftarrow \sigma\!\left(f_g\bigl(\operatorname{Concat}(\mathbf{x}_{LH},\mathbf{x}_{HL},\mathbf{x}_{HH})\bigr)\right)$
   \STATE $\widetilde{\mathbf{x}}_{LL} \leftarrow \mathbf{y}_{LL} \odot (\mathbf{1}+\mathbf{g})$
   \STATE $\widehat{\mathbf{x}}' \leftarrow \mathrm{IDWT}(\widetilde{\mathbf{x}}_{LL},\mathbf{x}_{LH},\mathbf{x}_{HL},\mathbf{x}_{HH})$; restore original size
   \STATE $\mathbf{y} \leftarrow f^{\mathrm{out}}_{1\times1}(\widehat{\mathbf{x}}')$
   \STATE \textbf{if} channels match \textbf{then} $\mathbf{y} \leftarrow \mathbf{x}+\mathbf{y}$
   \STATE \textbf{return} $\mathbf{y}$
 \end{algorithmic}
 \end{algorithm}

\section{Appendix J: Detailed Formulations for LGE}
\label{sec:appendix_lge_details}

This appendix provides the complete mathematical formulations for the Log-Gabor Enhancer (LGE) module, including the Log-Gabor filter construction and the WTConv variant.

\subsection{Log-Gabor Filter Construction}

The Log-Gabor kernel $\mathbf{g}_{s,k}$ is constructed in the spatial domain by rotating a centered coordinate grid and applying a log-normal radial envelope with a cosine angular term. For orientation index $k \in \{0,\ldots,K{-}1\}$ and scale index $s \in \{0,\ldots,S{-}1\}$:
\vspace{-0.2cm}
\begin{equation}
\label{eq:loggabor_appendix}
\begin{gathered}
(u',v') = (u\cos\phi_k + v\sin\phi_k, -u\sin\phi_k + v\cos\phi_k),\\
r = \sqrt{{u'}^{2}+{v'}^{2}}+\varepsilon,\quad \theta = \mathrm{atan2}(v',u'),\\
\mathbf{g}_{s,k}(u,v) = \exp\!\left(-\frac{[\log(r/\rho_s)]^{2}}{2\,[\log 2]^{2}}\right)\cos\theta,
\end{gathered}
\vspace{-0.2cm}
\end{equation}
where $\phi_k = k\pi/K$ is the orientation angle, $\rho_s = 0.5 + s \cdot 0.3$ is the scale parameter, and $\varepsilon$ is a small constant for numerical stability. The denominator $2[\log 2]^2 \approx 0.961$ arises from setting the bandwidth parameter $\sigma = 2\rho_s$, which yields one octave of frequency coverage per scale.

The Log-Gabor function has two key properties that distinguish it from hand-crafted edge filters:
\begin{itemize}
    \item \textbf{Zero DC response}: The log-normal envelope naturally excludes the DC component, preventing the filter from responding to uniform regions.
    \item \textbf{Optimal bandwidth}: The symmetric Gaussian profile in log-frequency space provides uniform coverage across scales, avoiding the bandwidth asymmetry of standard Gabor filters.
\end{itemize}

\subsection{Wavelet-Transform Convolution (WTConv)}

LGE-W replaces the local mixing operator $f_{\mathrm{mix}}$ with a Wavelet-Transform Convolution (WTConv), which performs multi-level subband mixing in the Haar domain:
\vspace{-0.2cm}
\begin{equation}
\label{eq:wtconv_appendix}
{\scriptsize
\mathrm{WTConv}(\mathbf{z}) = \mathcal{S}_0\,\mathcal{D}_0(\mathbf{z}) + \mathrm{IDWT}\!\left(\sum_{i=1}^{L} \mathcal{S}_i\,\mathcal{D}_{4,i}(\mathrm{DWT}_i(\mathbf{z}))\right),
}
\end{equation}
where the notation is summarized in Table~\ref{tab:wtconv_notation}.

\begin{table}[h]
\centering
\caption{Notation for WTConv.}
\label{tab:wtconv_notation}
\begin{tabular}{@{}l>{\raggedright\arraybackslash}p{0.70\linewidth}@{}}
 \toprule
 \cellcolor{blue!5}\textbf{Notation} & \cellcolor{blue!5}\textbf{Description} \\
 \midrule
 $\mathcal{D}_0$ & Depthwise conv in spatial domain ($5{\times}5$) \\
 $\mathcal{D}_{4,i}$ & Grouped depthwise conv over four subbands at level $i$ \\
 $L$ & Number of wavelet decomposition levels \\
 $\mathcal{S}_0$ & Learnable scale for base path (init $1.0$) \\
$\mathcal{S}_i$ & Learnable scales for wavelet levels (init $0.1$) \\
\bottomrule
\end{tabular}
\end{table}
The base path provides direct spatial-domain processing, while the wavelet path enables frequency-selective refinement with adaptive scaling. By operating in the wavelet domain, WTConv achieves a larger effective receptive field than a standard convolution of the same kernel size, making it particularly effective at the highest-resolution neck layer where context aggregation is most beneficial for small object detection.

\subsection{LGE Pseudo-code}

 \begin{algorithm}[H]
 \caption{LGE / LGE-W forward pass (Log-Gabor Enhancer and WTConv variant).}
 \label{alg:lge}
 \scriptsize
\begin{algorithmic}
   \STATE \textbf{Input:} $\mathbf{x}\in\mathbb{R}^{C\times H\times W}$, orientations $K$, scales $S$
   \STATE \textbf{Learnable:} logits $\boldsymbol{\alpha}\in\mathbb{R}^{S}$, $\boldsymbol{\beta}\in\mathbb{R}^{K}$, gate $\gamma$
   \STATE \textbf{Output:} $\mathbf{y}\in\mathbb{R}^{C\times H\times W}$
   \STATE $\mathbf{x}_{\mathrm{skip}} \leftarrow \mathbf{x}$ \COMMENT{or $1\times1$ projection if channels change}
   \STATE Compute Log-Gabor subbands $\{\mathbf{h}_{s,k}\}_{s=1..S,k=1..K}$ per channel (fixed filter bank; grouped/depthwise conv)
   \STATE $\mathbf{w}^{(S)}\leftarrow\operatorname{softmax}(\boldsymbol{\alpha})$, $\mathbf{w}^{(K)}\leftarrow\operatorname{softmax}(\boldsymbol{\beta})$
   \STATE $\mathbf{h} \leftarrow \sum_{s=1}^{S}\sum_{k=1}^{K} \mathbf{w}^{(S)}_s\,\mathbf{w}^{(K)}_k\,\mathbf{h}_{s,k}$
   \STATE $\mathbf{h} \leftarrow \sigma(\gamma)\,\mathbf{h}$ \COMMENT{global gate}
   \STATE \textbf{if} LGE \textbf{then} $\mathbf{u} \leftarrow \mathrm{DWConv}_{3\times3}(\mathbf{h})$
   \STATE \textbf{else} \COMMENT{LGE-W} $\mathbf{u} \leftarrow \mathrm{WTConv}(\mathbf{h})$ when $C$ preserved; otherwise use standard $3\times3$ conv
   \STATE $\mathbf{y} \leftarrow \mathbf{x}_{\mathrm{skip}} + \mathbf{u}$
   \STATE \textbf{return} $\mathbf{y}$
 \end{algorithmic}
 \end{algorithm}

 \begin{algorithm}[H]
 \caption{WTConv forward pass (used in LGE-W).}
 \label{alg:wtconv}
 \scriptsize
\begin{algorithmic}
   \STATE \textbf{Input:} $\mathbf{z}\in\mathbb{R}^{C\times H\times W}$, levels $L$ (default $1$)
   \STATE \textbf{Output:} $\mathrm{WTConv}(\mathbf{z})\in\mathbb{R}^{C\times H\times W}$
   \STATE $\mathbf{z}_{\mathrm{base}} \leftarrow \mathcal{S}_0\,\mathcal{D}_0(\mathbf{z})$ \COMMENT{spatial depthwise conv + learnable scale}
   \STATE Initialize $\mathbf{z}_{LL} \leftarrow \mathbf{z}$; stacks $\mathcal{L}\leftarrow[\ ]$, $\mathcal{H}\leftarrow[\ ]$, shapes $\mathcal{S}\leftarrow[\ ]$
   \FOR{$i=1$ to $L$}
     \STATE Record shape; pad to even size if needed; save shape to $\mathcal{S}$
     \STATE $(\mathbf{z}_{LL}^{\mathrm{raw}},\mathbf{z}_{LH},\mathbf{z}_{HL},\mathbf{z}_{HH}) \leftarrow \mathrm{DWT}(\mathbf{z}_{LL})$
     \STATE $\mathbf{z}_{LL} \leftarrow \mathbf{z}_{LL}^{\mathrm{raw}}$ \COMMENT{recurse on raw LL}
     \STATE Apply $\mathcal{S}_i\,\mathcal{D}_{4,i}$ over the 4-subband tensor (reshape to $4C$, grouped depthwise conv, reshape back)
     \STATE Push processed LL to $\mathcal{L}$ and processed $(LH,HL,HH)$ to $\mathcal{H}$
   \ENDFOR
   \STATE $\mathbf{z}_{LL}^{\mathrm{next}} \leftarrow 0$
   \FOR{$i=L$ down to $1$}
     \STATE Pop processed LL from $\mathcal{L}$ and processed $(LH,HL,HH)$ from $\mathcal{H}$
     \STATE $\mathbf{z}_{LL} \leftarrow \mathbf{z}_{LL} + \mathbf{z}_{LL}^{\mathrm{next}}$ \COMMENT{level-wise LL accumulation}
     \STATE $\mathbf{z}_{LL}^{\mathrm{next}} \leftarrow \mathrm{IDWT}(\mathbf{z}_{LL},\mathbf{z}_{LH},\mathbf{z}_{HL},\mathbf{z}_{HH})$; crop to saved shape
   \ENDFOR
   \STATE $\mathbf{z}_{\mathrm{wav}} \leftarrow \mathbf{z}_{LL}^{\mathrm{next}}$
   \STATE \textbf{return} $\mathbf{z}_{\mathrm{base}} + \mathbf{z}_{\mathrm{wav}}$
 \end{algorithmic}
 \end{algorithm}

\section{Appendix K: Detailed Formulations for FDHead}
\label{sec:appendix_fdhead_details}

This appendix provides the complete mathematical formulations for the Frequency-Driven Head (FDHead), including the DEConv re-parameterization and the P2 high-frequency gate.

\subsection{DEConv Re-parameterization}

The DEConv block aggregates five directional-difference kernels into a single convolution at inference:
\vspace{-0.2cm}
\begin{equation}
\label{eq:deconv_appendix}
\mathrm{DEConv}(\mathbf{u}) = \varphi\Bigl(\bigl(\sum_{m=1}^{5} \mathbf{K}_{m}\bigr) * \mathbf{u} + \sum_{m=1}^{5}\mathbf{b}_{m}\Bigr),
\vspace{-0.2cm}
\end{equation}
where $\varphi(\cdot)$ denotes normalization and activation. The five kernels $\mathbf{K}_{m}\in\mathbb{R}^{C\times C\times 3\times 3}$ are summarized in Table~\ref{tab:deconv_kernels}.

\begin{table}[h]
\centering
\caption{DEConv kernel descriptions.}
\label{tab:deconv_kernels}
\begin{tabular}{@{}l>{\raggedright\arraybackslash}p{0.70\linewidth}@{}}
 \toprule
 \cellcolor{blue!5}\textbf{Kernel} & \cellcolor{blue!5}\textbf{Description} \\
 \midrule
 $\mathbf{K}_1$ (cd) & Center-difference: center vs.\ surrounding context \\
 $\mathbf{K}_2$ (hd) & Horizontal-difference: horizontal edges \\
 $\mathbf{K}_3$ (vd) & Vertical-difference: vertical edges \\
 $\mathbf{K}_4$ (ad) & Adaptive-difference: rotation-invariant patterns \\
$\mathbf{K}_5$ (std) & Standard convolution: general features \\
\bottomrule
\end{tabular}
\end{table}
At inference, all five kernels are merged into a single $3{\times}3$ convolution, enabling efficient computation while preserving directional sensitivity.

\subsection{P2 High-Frequency Gate}

The P2 gate estimates boundary confidence from wavelet subband energy. Let $\mathbf{f}_1=[\mathbf{f}_a,\mathbf{f}_b]$ with $\mathbf{f}_a\in\mathbb{R}^{C_f\times H\times W}$ being the gated channel subset. The complete formulation is:
\vspace{-0.2cm}
\begin{equation}
\label{eq:hfgate_appendix}
\begin{gathered}
(\mathbf{f}_{LH},\mathbf{f}_{HL},\mathbf{f}_{HH}) = \mathrm{DWT}_{\mathrm{HF}}(\mathbf{f}_a),\\
\mathbf{h} = \sum_{k\in\{LH,HL,HH\}} \operatorname{softmax}(\boldsymbol{\omega})_k\,|\mathbf{f}_k|,\\
\mathbf{s} = \operatorname{GAP}(\mathbf{h}),\qquad \mathbf{g} = \sigma\bigl(\mathrm{Conv}_{1\times 1}^{(2)}(\mathrm{SiLU}(\mathrm{Conv}_{1\times 1}^{(1)}(\mathbf{s})))\bigr),\\
\widetilde{\mathbf{f}}_a = \mathbf{f}_a\odot(1+\alpha\,\mathbf{g}),
\end{gathered}
\vspace{-0.2cm}
\end{equation}
where $\boldsymbol{\omega}\in\mathbb{R}^{3}$ are learnable logits weighting the three high-frequency subbands, $\operatorname{GAP}(\cdot)$ averages over spatial dimensions, and the channel gate uses two $1\times 1$ convolutions with a SiLU nonlinearity and sigmoid output. The gated feature $\widetilde{\mathbf{f}}_1=[\widetilde{\mathbf{f}}_a,\mathbf{f}_b]$ is then fed only to the box branch.

\subsection{FDHead Pseudo-code}

 \begin{algorithm}[H]
 \caption{FDHead forward pass (Frequency-Driven Head).}
 \label{alg:fdhead}
 \scriptsize
\begin{algorithmic}
   \STATE \textbf{Input:} multi-level features $\{\mathbf{x}_i\}_{i=1}^{N}$ (with $i=1$ being $P2$)
   \STATE \textbf{Output:} box logits $\{\mathbf{b}_i\}$ and class logits $\{\mathbf{p}_i\}$
   \FOR{each level $i=1..N$}
     \STATE $\mathbf{u}_i \leftarrow \mathrm{Conv}_{1\times1}(\mathbf{x}_i)$; apply normalization
     \STATE $\mathbf{f}_i \leftarrow \mathrm{DW{-}PW}(\mathrm{DEConv}(\mathbf{u}_i))$ \COMMENT{shared tower}
     \IF{$i=1$ \COMMENT{P2, box branch only}}
       \STATE Split channels $\mathbf{f}_1=[\mathbf{f}_a,\mathbf{f}_b]$ with gated subset width $C_f$
       \STATE $(\mathbf{f}_{LH},\mathbf{f}_{HL},\mathbf{f}_{HH}) \leftarrow \mathrm{DWT}_{\mathrm{HF}}(\mathbf{f}_a)$
       \STATE $\mathbf{w} \leftarrow \operatorname{softmax}(\boldsymbol{\omega})$; $\mathbf{h}\leftarrow w_{LH}|\mathbf{f}_{LH}|+w_{HL}|\mathbf{f}_{HL}|+w_{HH}|\mathbf{f}_{HH}|$
       \STATE $\mathbf{s} \leftarrow \operatorname{GAP}(\mathbf{h})$; $\mathbf{g} \leftarrow \mathrm{Gate}(\mathbf{s})$ \COMMENT{$1\times1$ conv $\rightarrow$ SiLU $\rightarrow$ $1\times1$ conv $\rightarrow$ sigmoid}
       \STATE $\widetilde{\mathbf{f}}_a \leftarrow \mathbf{f}_a\odot(1+\alpha\,\mathbf{g})$; $\mathbf{f}^{\mathrm{box}}_1 \leftarrow [\widetilde{\mathbf{f}}_a,\mathbf{f}_b]$
       \STATE $\mathbf{b}_1 \leftarrow \mathrm{Scale}_1(\mathcal{H}_{\mathrm{box}}(\mathbf{f}^{\mathrm{box}}_1))$; $\mathbf{p}_1 \leftarrow \mathcal{H}_{\mathrm{cls}}(\mathbf{f}_1)$
     \ELSE
       \STATE $\mathbf{b}_i \leftarrow \mathrm{Scale}_i(\mathcal{H}_{\mathrm{box}}(\mathbf{f}_i))$; $\mathbf{p}_i \leftarrow \mathcal{H}_{\mathrm{cls}}(\mathbf{f}_i)$
     \ENDIF
   \ENDFOR
   \STATE \textbf{return} $\{\mathbf{b}_i\}$ and $\{\mathbf{p}_i\}$ \COMMENT{decode boxes as in main text}
 \end{algorithmic}
 \end{algorithm}

\clearpage
\section{Appendix M: Comparison of Frequency-Domain Methods in Object Detection}
\label{sec:appendix_l_comparison}

Table~\ref{tab:frequency_methods_comparison} positions our work relative to existing frequency-domain approaches in object detection.
We compare methods at the level of \emph{frequency-domain enhancement operators}, focusing on where and how frequency information is injected into detection pipelines.

\noindent\textbf{Clarification on comparison criteria.}
All attributes are evaluated at the level of \emph{frequency-domain enhancement operators}.
\emph{Invertible} refers to the reversibility of the employed frequency transform (e.g., FFT/IFFT, WT/IWT), rather than end-to-end network invertibility.
\emph{Extra Branch} indicates explicit \emph{parallel computational pathways} introduced by the frequency operator, which incur additional architectural and optimization overhead.

\begin{table}[H]
  \centering
  \caption{Comparison of frequency-domain methods in object detection.
  The comments emphasize the \emph{limitations} implied by each design choice.
  Our DERNet framework is highlighted in the last row.}
  \label{tab:frequency_methods_comparison}
  \vspace{0.1cm}
  \setlength{\tabcolsep}{1.2pt}
  \renewcommand{\arraystretch}{1.05}
  \fontsize{3.4pt}{4.0pt}\selectfont
  \resizebox{\textwidth}{!}{
  \begin{tabular}{@{}>{\raggedright\arraybackslash}p{1.05cm}
  >{\centering\arraybackslash}p{1.05cm}
  >{\centering\arraybackslash}p{0.8cm}
  >{\centering\arraybackslash}p{0.55cm}
  >{\centering\arraybackslash}p{0.6cm}
  >{\centering\arraybackslash}p{0.7cm}
  >{\centering\arraybackslash}p{0.8cm}
  >{\raggedright\arraybackslash}p{1.15cm}@{}}
  \toprule
  \cellcolor{blue!5}\textbf{Method}
  & \cellcolor{blue!5}\textbf{Domain Op.}
  & \cellcolor{blue!5}\textbf{Stage}
  & \cellcolor{blue!5}\textbf{Localized}
  & \cellcolor{blue!5}\textbf{Invertible}
  & \cellcolor{blue!5}\textbf{Learn Filters}
  & \cellcolor{blue!5}\textbf{Extra Branch}
  & \cellcolor{blue!5}\textbf{Comment} \\
  \midrule
  
  GFNet
  & FFT
  & B
  & \textcolor{red}{No}
  & \textcolor{green}{Yes}
  & \textcolor{green}{Yes}
  & \textcolor{orange}{Partial}
  & No localization \\
  
  FDConv
  & Frequency-aware
  & B/N
  & \textcolor{green}{Yes}
  & \textcolor{red}{No}
  & \textcolor{green}{Yes}
  & \textcolor{red}{No}
  & Implicit only \\
  
  FreqFusion
  & Adaptive Filter
  & N
  & \textcolor{green}{Yes}
  & \textcolor{red}{No}
  & \textcolor{green}{Yes}
  & \textcolor{green}{Yes}
  & Generator overhead \\
  
  WTConv
  & Wavelet
  & B/N
  & \textcolor{green}{Yes}
  & \textcolor{green}{Yes}
  & \textcolor{orange}{Partial}
  & \textcolor{red}{No}
  & No adaptivity \\
  
  HS-FPN
  & HF Response
  & N
  & \textcolor{green}{Yes}
  & \textcolor{red}{No}
  & \textcolor{green}{Yes}
  & \textcolor{green}{Yes}
  & Extra branch \\
  
  Wavelet Transformer
  & Complex Wavelet
  & B
  & \textcolor{green}{Yes}
  & \textcolor{green}{Yes}
  & \textcolor{orange}{Partial}
  & \textcolor{red}{No}
  & Single-stage only \\
  
  \textbf{DERNet (Ours)}
  & \textbf{Wavelet/Log-Gabor}
  & \textbf{B/N/H}
  & \textbf{\textcolor{green}{Yes}}
  & \textbf{\textcolor{green}{Yes/No}}
  & \textbf{\textcolor{green}{Yes}}
  & \textbf{\textcolor{red}{No}}
  & \textbf{Full-stage adaptive} \\
  
  \bottomrule
  \end{tabular}}
  \vspace{-2mm}
  \end{table}
  
  \noindent\textbf{Positioning of Our Work:}
  Unlike prior approaches that introduce frequency-domain enhancement operators at isolated stages,
  DERNet provides a \emph{stage-adaptive frequency modeling framework} spanning backbone, neck, and head.
  
  \begin{itemize}
  \item \textbf{Spatial Localization:}
  We employ wavelet transforms in backbone and head stages to preserve spatial-frequency correspondence,
  while using directional Log-Gabor filters in the neck to enhance edge and texture sensitivity.
  This contrasts with global FFT-based methods (e.g., GFNet) that discard spatial locality.
  
  \item \textbf{Invertibility:}
  DERNet preserves transform-level invertibility where structural fidelity is critical (B/H),
  and deliberately relaxes invertibility in the neck where directional discrimination is more beneficial,
  reflecting a stage-aware trade-off rather than a uniform design constraint.
  
  \item \textbf{Learnable Filters:}
  By combining fixed analytical filter banks with learnable aggregation weights,
  DERNet achieves adaptability with reduced parameter redundancy,
  avoiding the heavy reliance on fully learnable spectral filters.
  
  \item \textbf{Architectural Overhead (Extra Branches):}
  Many frequency-based methods (e.g., FreqFusion, HS-FPN) introduce explicit parallel branches to inject spectral cues.
  While effective, such designs increase architectural complexity, parameter coupling,
  and optimization difficulty, particularly in multi-stage detectors.
  In contrast, DERNet integrates frequency-domain operators \emph{inline} within existing B/N/H pathways,
  achieving frequency awareness \emph{without introducing extra branches},
  thereby improving scalability, stability, and plug-and-play compatibility.
  \end{itemize}
  
  This design highlights that effective frequency modeling in detection is not only a matter of \emph{what} operator is used,
  but also \emph{where and how} it is integrated across stages.

  \clearpage
\section{Appendix N: Per-Class Performance Visualization}
\label{sec:appendix_j_perclass}

\begin{figure}[H]
\centering
\includegraphics[width=0.85\columnwidth]{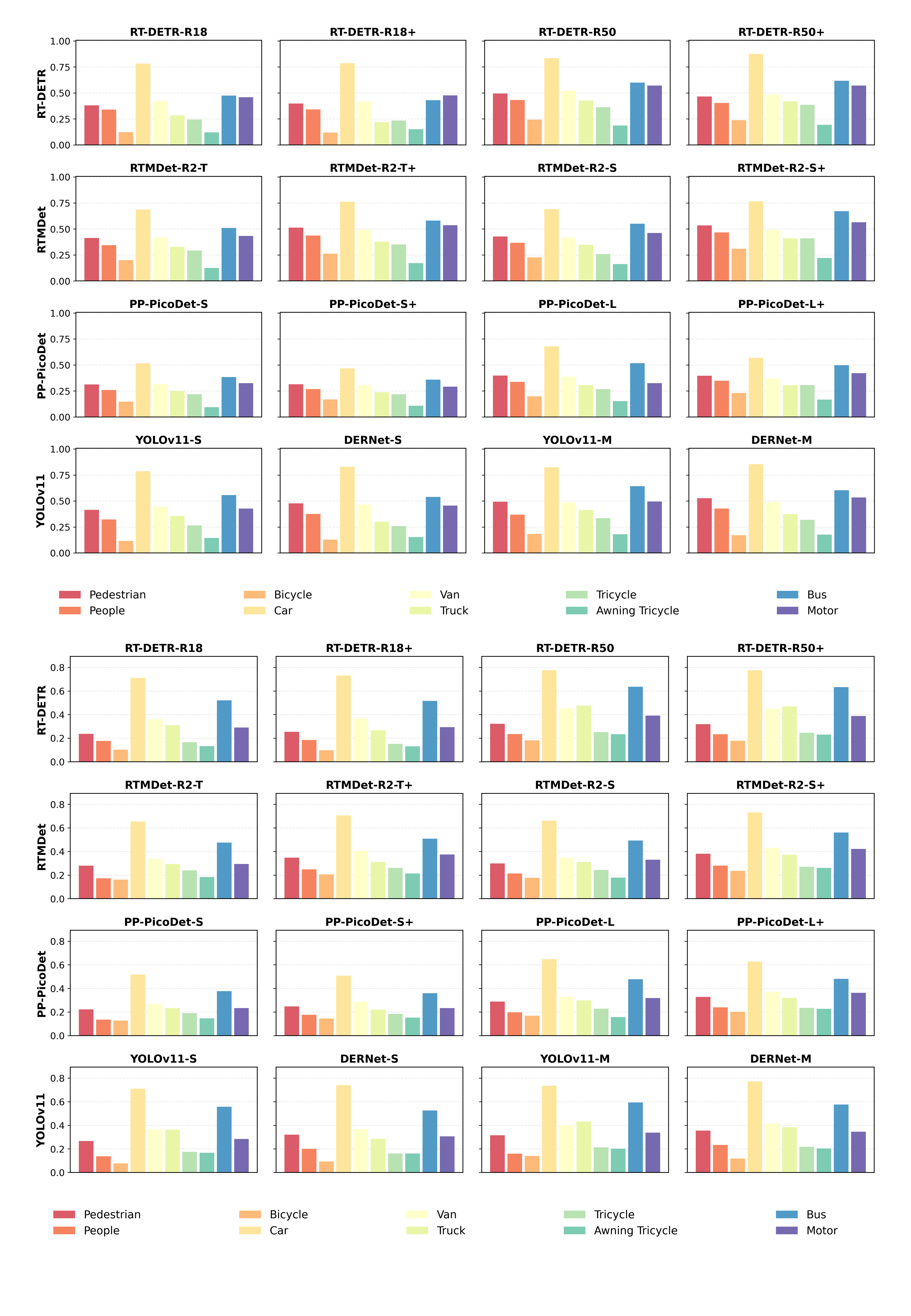}
\caption{Per-class performance comparison on VisDrone2019 validation set and test set.}
\label{fig:perclass_val}
\end{figure}

\clearpage
\begin{figure}[H]
\centering
\includegraphics[width=0.85\columnwidth]{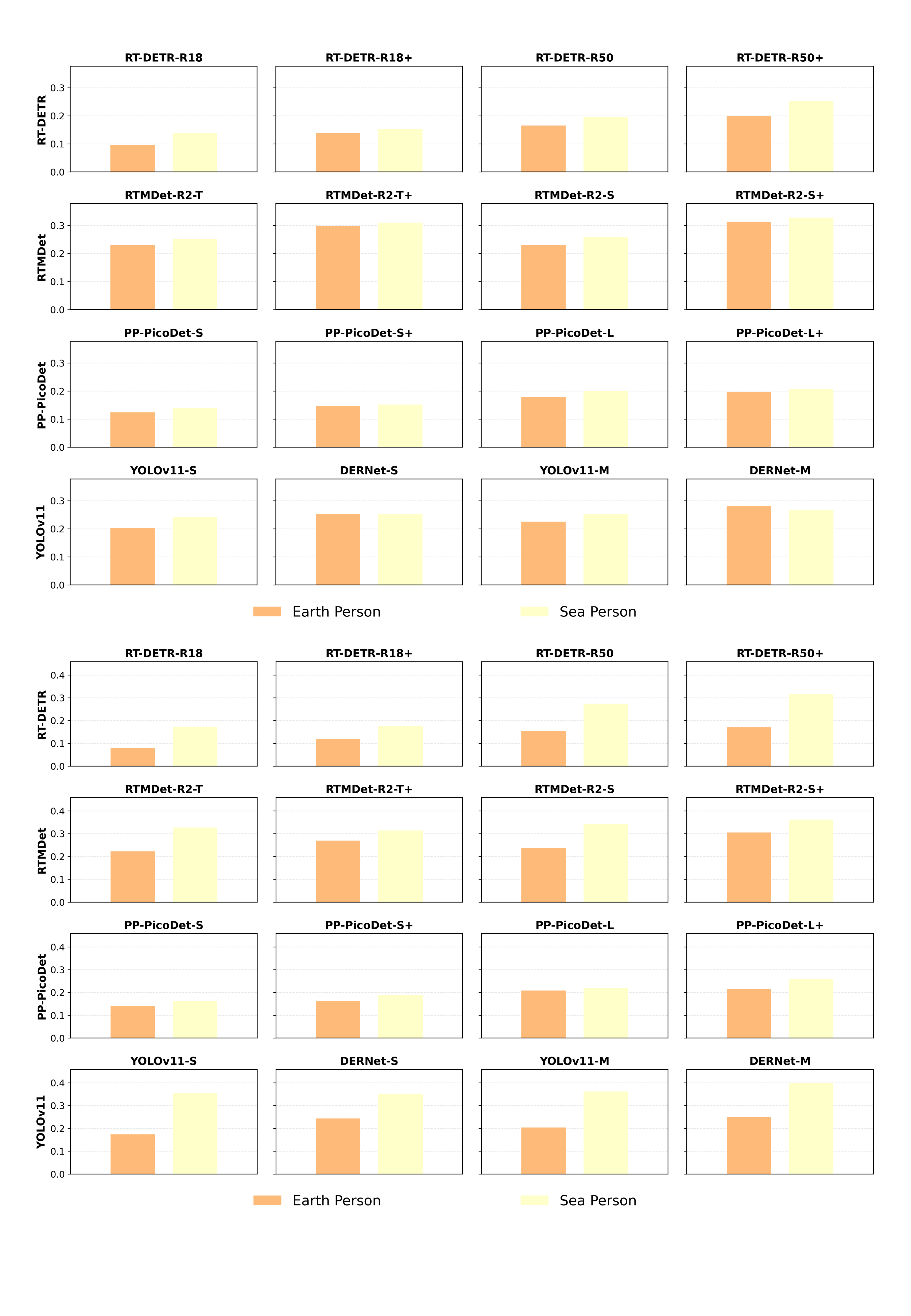}
\caption{Per-class performance comparison on TinyPerson validation set and test set.}
\label{fig:perclass_tp}
\end{figure}

\clearpage
\begin{figure}[H]
\centering
\includegraphics[width=0.85\columnwidth]{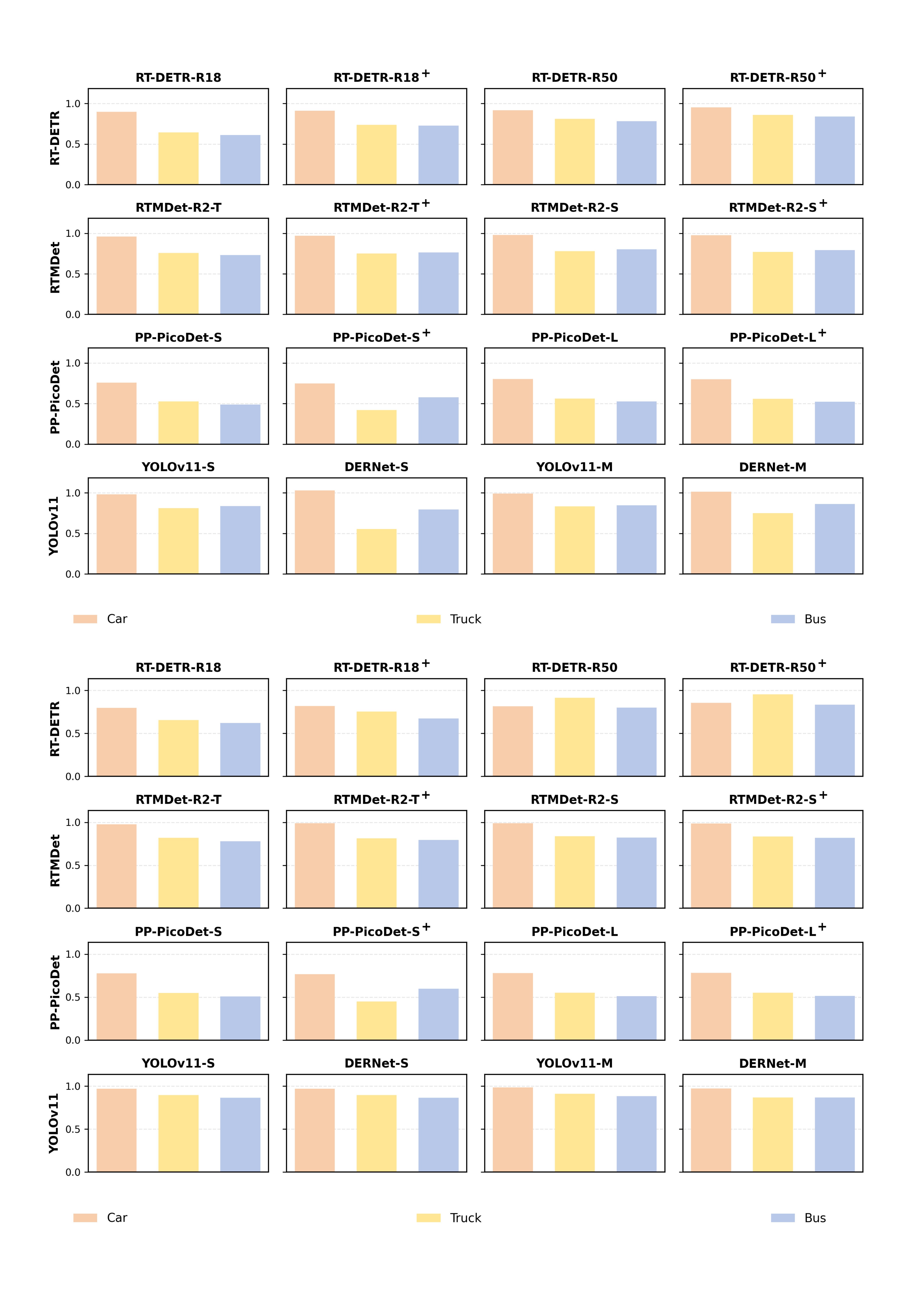}
\caption{Per-class performance comparison on UAVDT validation set and test set.}
\label{fig:perclass_uavdt}
\end{figure}

\begin{figure}[H]
\centering
\includegraphics[width=0.85\columnwidth]{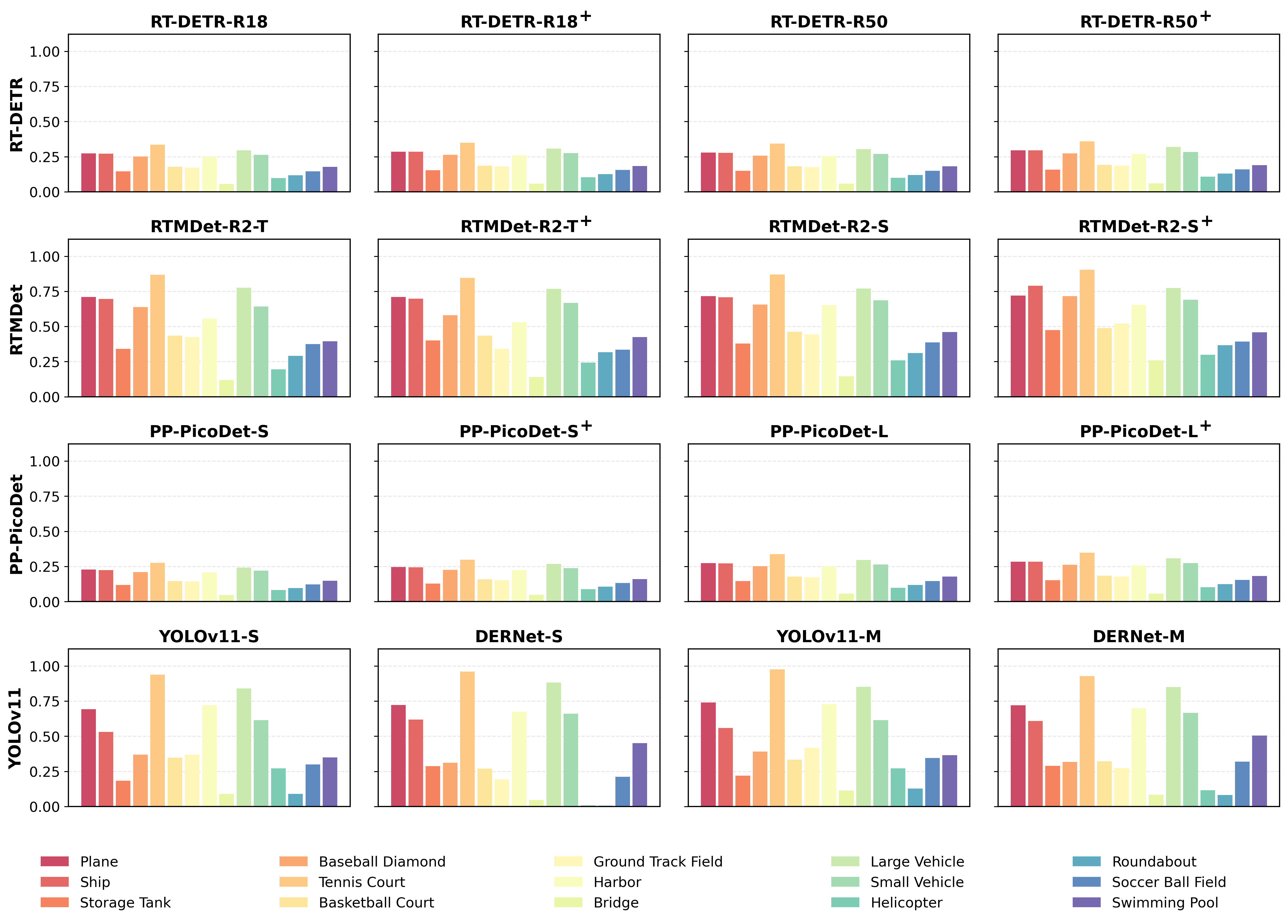}
\caption{Per-class performance comparison on Dotav1 validation set.}
\label{fig:perclass_dotav1}
\end{figure}

\section{Dataset-dependent Benefit Analysis}
\label{sec:appendix_dataset_benefit}

\begin{figure}[H]
  \centering
  \setlength{\tabcolsep}{1.2pt}
  \renewcommand{\arraystretch}{0.95}
  \begin{tabular}{c c}
  \includegraphics[width=0.42\textwidth]{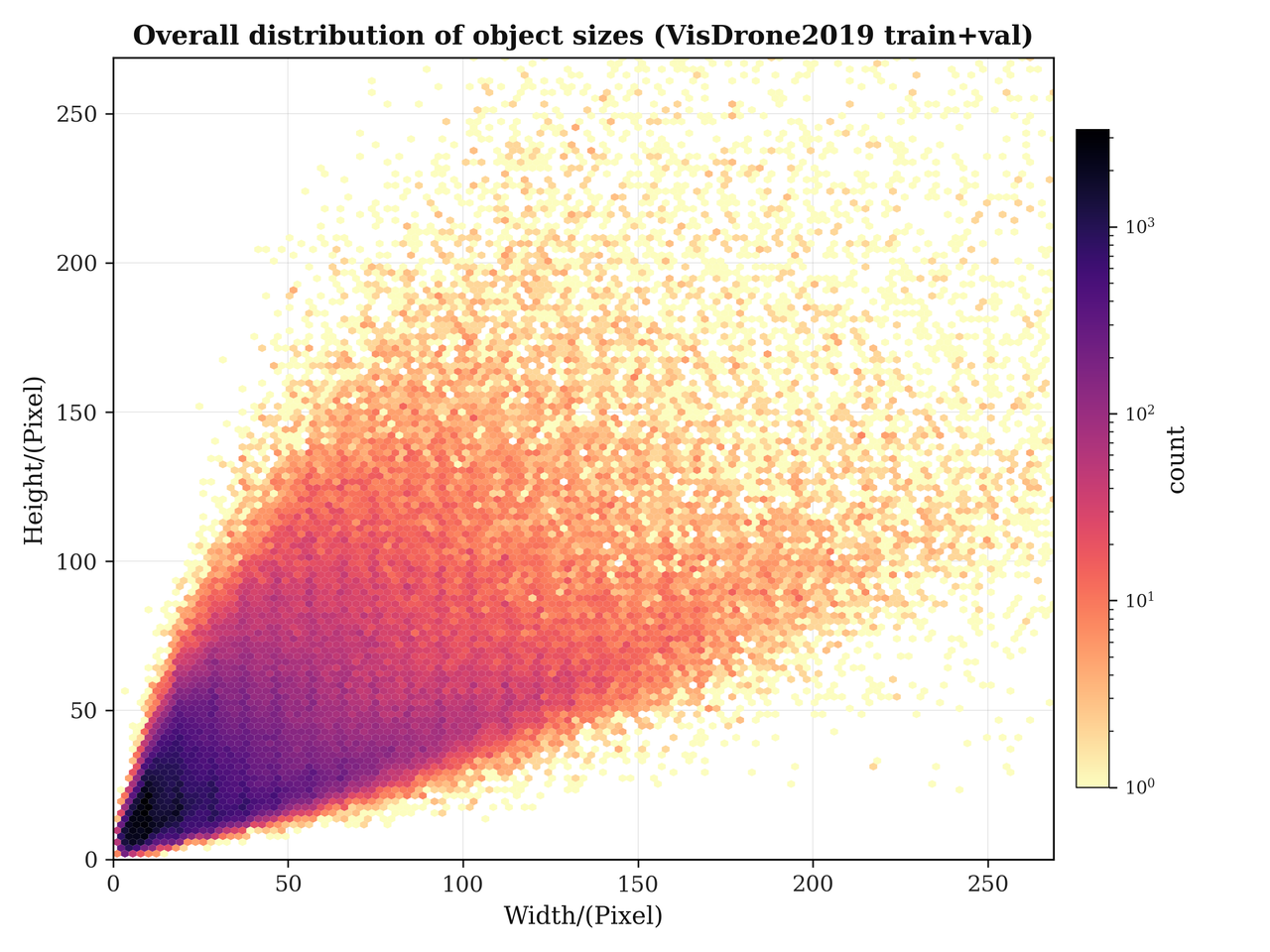} &
  \includegraphics[width=0.42\textwidth]{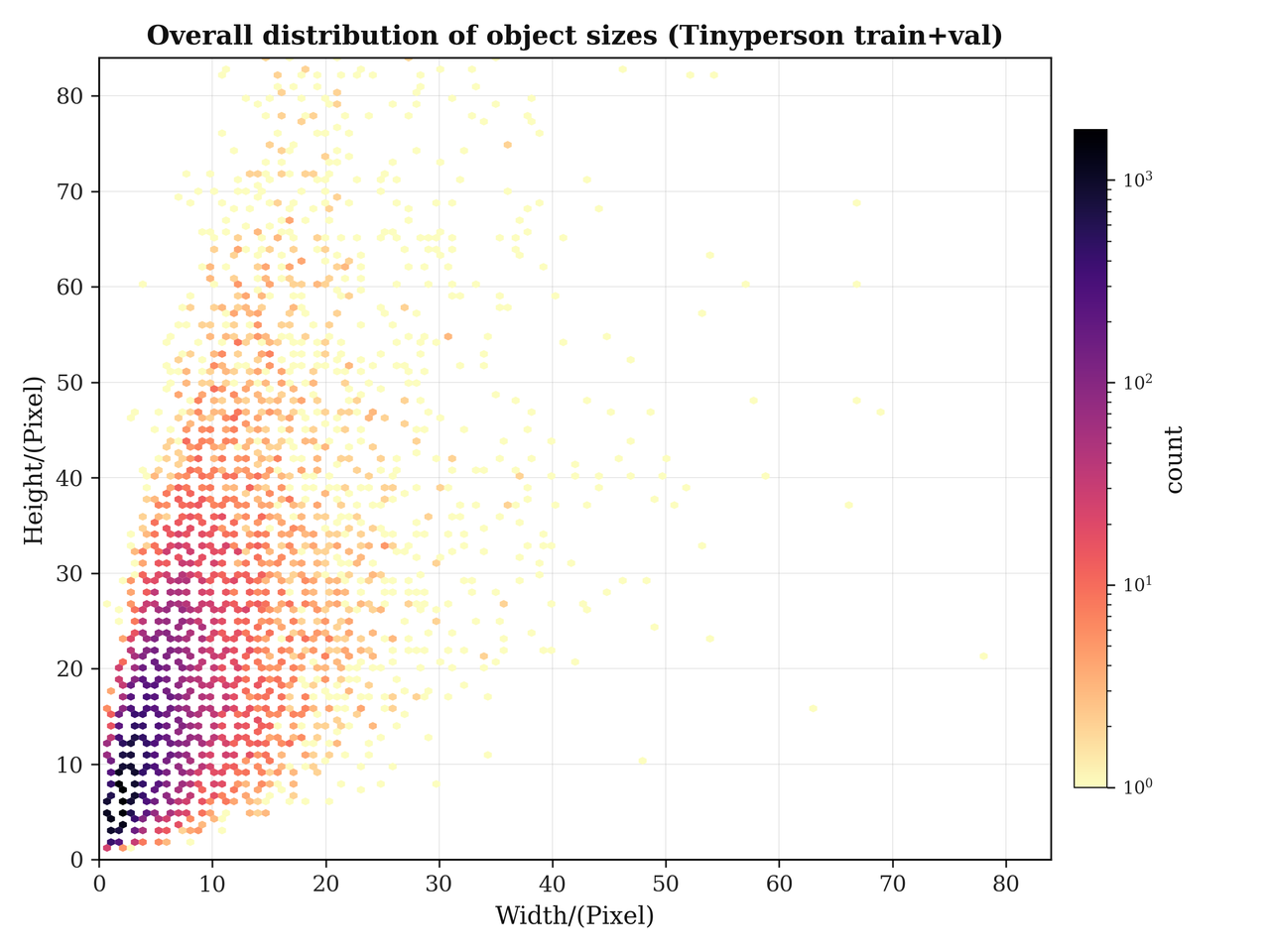} \\
  \includegraphics[width=0.42\textwidth]{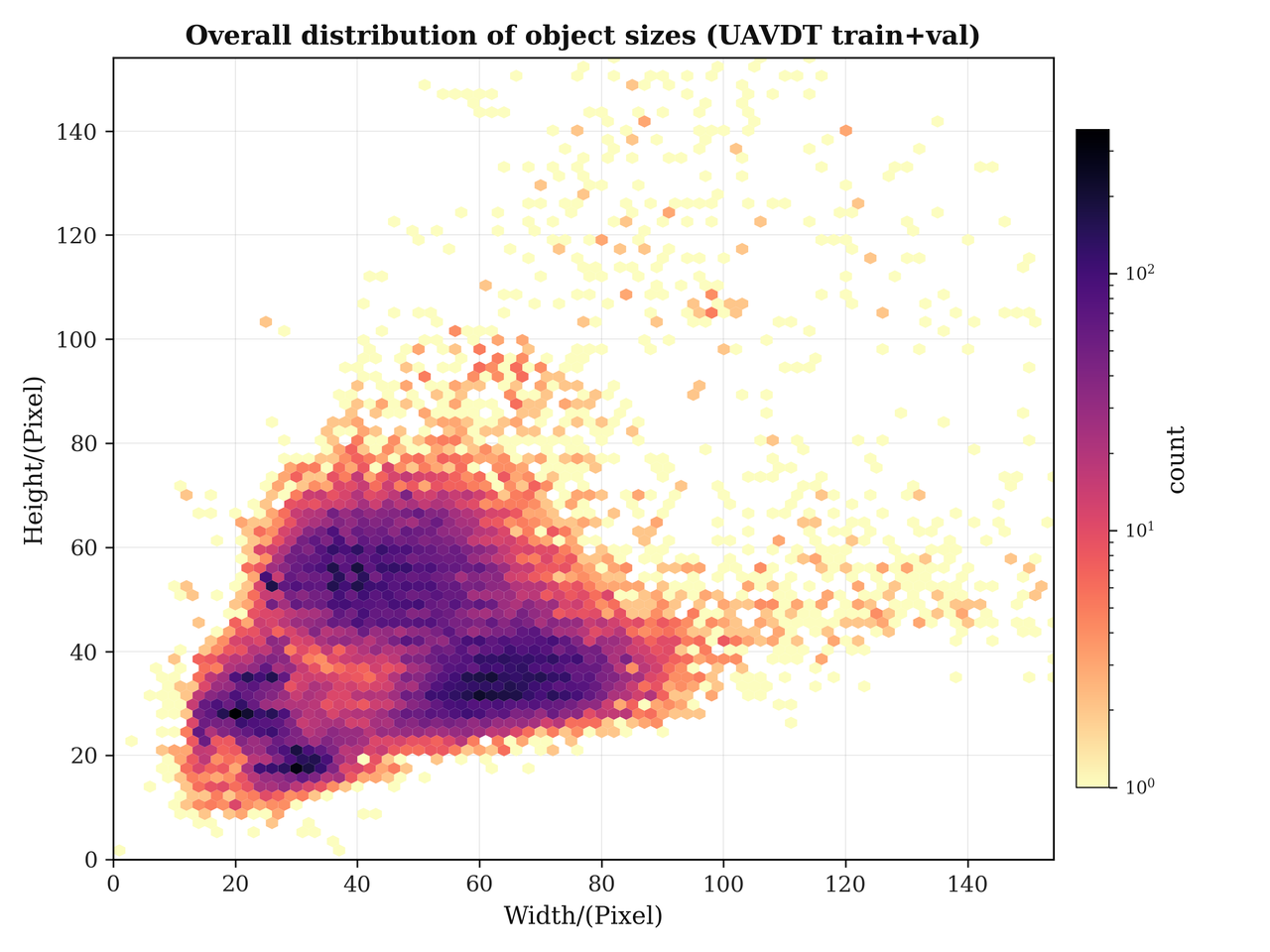} &
  \includegraphics[width=0.42\textwidth]{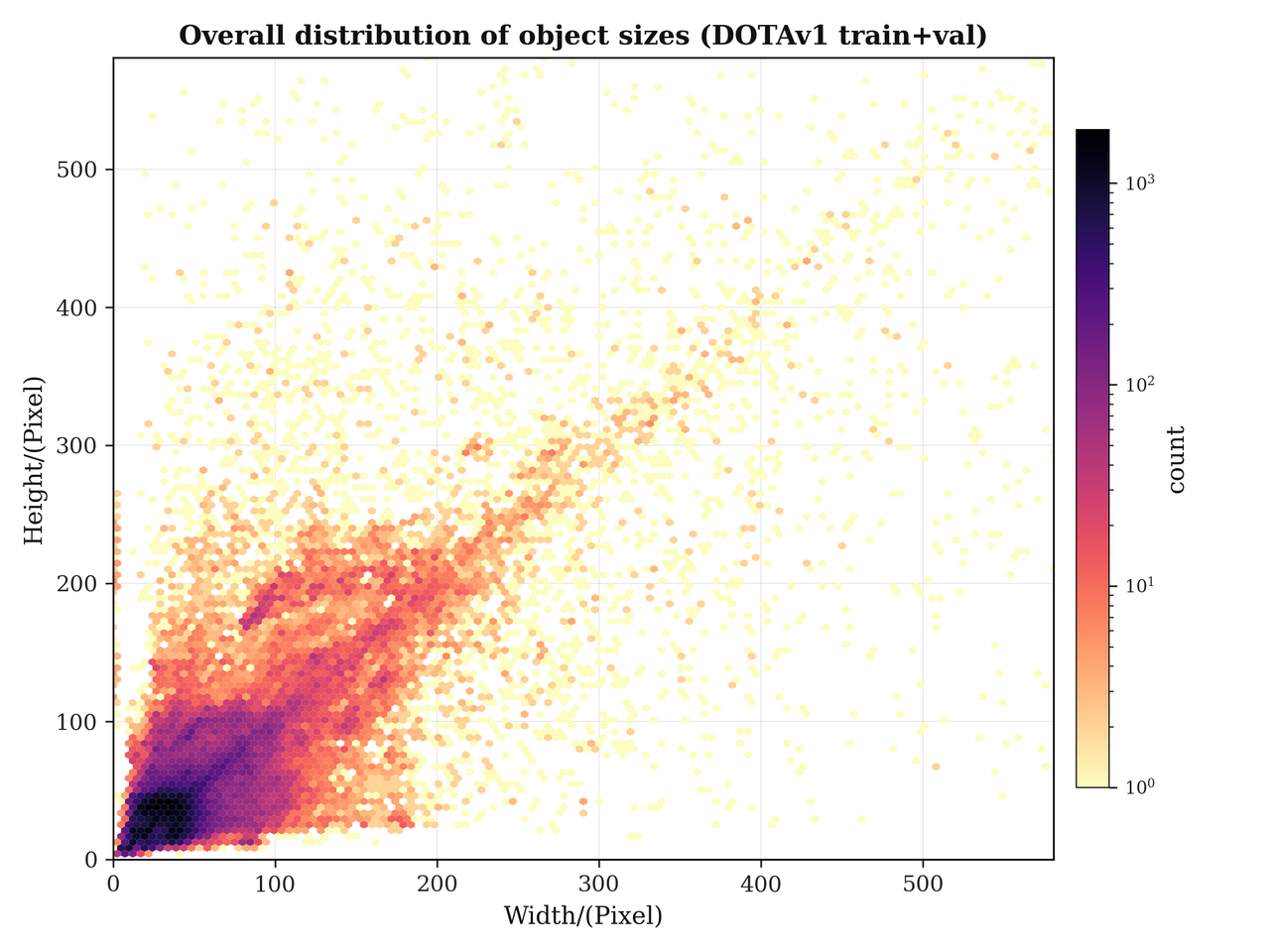} \\
  \end{tabular}
  \caption{Dataset-dependent overall distribution comparison across four benchmarks.}
  \label{fig:dataset_distribution_overall}
  \end{figure}

The dataset-dependent trend in Table~\ref{tab:across_arch} supports DER's
spectral motivation. As shown in Fig.~\ref{fig:dataset_distribution_overall}, DER yields
the largest gains in information-starved scenes with extremely small, dense, or
low-contrast targets, such as TinyPerson and distant crowded regions in
VisDrone2019, where fragile high-frequency cues are easily lost. For semantically
richer objects, such as UAVDT vehicles category and DOTAv1 planes/ships category, baselines already
capture reliable low-frequency semantics, so DER mainly provides boundary
refinement and yields moderate but consistent gains. Thus, DER is most beneficial
when spectral degradation is the dominant bottleneck, while remaining robust
across easier regimes.

\section{Fixed versus Learnable Log-Gabor Filters}
\label{sec:appendix_fixed_learnable_loggabor}

Fixed Log-Gabor filters preserve the intended zero-DC, frequency-selective
bandpass prior for directional high-frequency recovery. If these filters are
made fully learnable without explicit spectral constraints, this prior is no
longer guaranteed: the filters may drift away from the Log-Gabor form and
effectively become Log-Gabor-initialized unconstrained depthwise convolutions,
potentially fitting dataset-specific spatial texture patterns. As shown in
Table~\ref{tab:learnable_loggabor}, learnable filters bring no measurable GFLOPs
increase, but increase parameters, cause a large module-level parameter surge,
and consistently reduce mAP$_{50}$. These results empirically support the
fixed-filter design, which preserves the intended spectral prior while avoiding
unnecessary parameters.

\begin{table}[H]
\centering
\caption{
Fixed versus learnable Log-Gabor filters on VisDrone2019 val. $K$ and $S$ denote
orientations and scales. Learnable filters increase module parameters but
consistently degrade mAP$_{50}$, supporting the fixed LGE design.
}
\label{tab:learnable_loggabor}
\setlength{\tabcolsep}{5pt}
\renewcommand{\arraystretch}{1.08}
\fontsize{9pt}{10.5pt}\selectfont
\begin{tabular}{@{}lcccc@{}}
\toprule
\rowcolor{blue!5}
\textbf{Configuration} 
& \textbf{Total Params} 
& \textbf{LGE Module Params} 
& \textbf{Inference GFLOPs} 
& \textbf{mAP$_{50}$} \\
\rowcolor{blue!5}
\textbf{(Learnable vs. Fixed)}
& \textbf{Increase}
& \textbf{Surge}
& \textbf{Increase}
& \textbf{Change} \\
\midrule
$K=1, S=2$ & $+0.045$M & $+115\%$ to $+276\%$ & $+0.0$ & $-0.004$ \\
$K=2, S=1$ & $+0.045$M & $+115\%$ to $+276\%$ & $+0.0$ & $-0.006$ \\
$K=2, S=2$ & $+0.090$M & $+142\%$ to $+448\%$ & $+0.0$ & $-0.007$ \\
\bottomrule
\end{tabular}
\end{table}
 

\end{document}